\documentclass{article} 
\usepackage{iclr2025_conference,times}

\usepackage{fancyhdr}

\usepackage{amsmath,amsfonts,bm}

\def\eqref#1{equation~\ref{#1}}

\def\1{\bm{1}}

\DeclareMathAlphabet{\mathsfit}{\encodingdefault}{\sfdefault}{m}{sl}
\SetMathAlphabet{\mathsfit}{bold}{\encodingdefault}{\sfdefault}{bx}{n}

\usepackage{hyperref}
\usepackage{lipsum}
\usepackage{url}
\usepackage{hyperref}       
\usepackage{url}            
\usepackage{booktabs}       
\usepackage{amsfonts}       
\usepackage{nicefrac}       
\usepackage{microtype}      
\usepackage{graphicx}
\usepackage{subcaption}
\usepackage{causality}

\usepackage[accept]{editing}
\usepackage{wrapfig}
\usepackage{stfloats}
\usetikzlibrary{decorations.pathreplacing,calligraphy}
\usetikzlibrary{patterns}
\usepackage{hyperref}
\usepackage{amsthm}

\usepackage{amsmath}
\usepackage{amssymb}
\usepackage{bbm}
\usepackage{multicol}
\usepackage{csquotes}
\usepackage{lipsum}
\usetikzlibrary{positioning, shadows}
\usetikzlibrary {arrows.meta,bending}

\usepackage{cutwin}
\usepackage{caption}
\usepackage{enumitem}
\usepackage{titletoc}
\usetikzlibrary {shapes.symbols}

\theoremstyle{plain}
\newtheorem{theorem}{Theorem}[section]

\newtheorem{lemma}[theorem]{Lemma}
\newtheorem{corollary}[theorem]{Corollary}
\theoremstyle{definition}
\newtheorem{definition}[theorem]{Definition}
\newtheorem{assumption}[theorem]{Assumption}
\newtheorem{example}[theorem]{Example}
\theoremstyle{remark}
\newtheorem{remark}[theorem]{Remark}

\usepackage{algorithm}
\usepackage{algorithmic}
\usepackage{xcolor}  
\definecolor{ForestGreen}{RGB}{34,139,34}

\title{Counterfactual Realizability}

\author{Arvind Raghavan \;{\normalfont and}\; Elias Bareinboim \vspace{+0.06in}\\
  Causal Artificial Intelligence Lab\\
  Columbia University\\
  \texttt{\{ar, eb\}@cs.columbia.edu} \\}

\iclrfinalcopy 
\begin{document}

\maketitle

\begin{abstract}
\xst{A commonly accepted belief suggests that}\xadd{It is commonly believed that}, in a real-world environment, samples can only be drawn from observational and interventional distributions, corresponding to Layers 1 and 2 of the \textit{Pearl Causal Hierarchy}. \xst{According to this belief, }Layer 3, representing counterfactual distributions, is believed to be inaccessible by definition.
However, Bareinboim, Forney, and Pearl (2015) introduced a procedure that allows an agent to sample directly from a counterfactual distribution, leaving open the \xst{possibility that }\xadd{question of what} other counterfactual quantities can be estimated directly via physical experimentation. \xst{In this paper, we}
We resolve this by introducing a formal definition of \textit{realizability}, the ability to draw samples from a distribution, and then developing a complete algorithm to determine whether an arbitrary counterfactual distribution is realizable given fundamental physical constraints, such as the inability to go back in time and subject the same unit to a different experimental condition. 
\xst{Building on this characterization, we further develop an algorithm for an agent to construct an optimal \textit{realizable strategy} in multi-arm bandit settings.}\xadd{We illustrate the implications of this new framework for counterfactual data collection using motivating examples from causal fairness and causal reinforcement learning.} While the baseline approach in these motivating settings typically follows an interventional or observational strategy, we show that a counterfactual strategy provably dominates \xst{(i.e. it is as good as or better than) }both.
\end{abstract}

\section{Introduction}
\label{sec:intro}

The \textit{Pearl Causal Hierarchy}, or PCH, is an important recent milestone in our understanding of causality \citep{pearl:mackenzie2018,Bareinboim2022OnPH}. The three layers of the PCH represent the distinct regimes of \textit{seeing}, \textit{doing}, and \textit{imagining}, with regard to an environment\xst{where an agent is deployed}. Consider an environment involving a decision variable $X$ and an outcome $Y$. Layer 1 ($\mathcal{L}_1$) represents \textit{observational} distributions, such as $P(Y\mid x)$. Layer 2 ($\mathcal{L}_2$) represents \xadd{\textit{interventional} }distributions\xst{from \textit{interventional} regimes}, such as $P(Y; \doo{x})$, using the $\doo{}$ operator. Layer 3 ($\mathcal{L}_3$) represents \textit{counterfactual} distributions {dealing with conflicting realities}, such as $P(Y_x \mid x', y')$: the distribution of $Y$ had $X$ been fixed as $x$, given that $X,Y$ were in fact naturally observed to be $x', y'$. Higher layers subsume lower ones, but are underdetermined by them \citep{ibeling2020probabilistic,Bareinboim2022OnPH}.

\xadd{Reasoning about $\mathcal{L}_3$-quantities plays a vital role in personalized decision-making \citep{MuellerPearl_2023}, analysing a causal effect into direct and indirect pathways \citep{pearl:01, rubin:04}, and constructing explanations for decisions, among other topics, in applications such as healthcare \citep{mueller:pearl23}, economics \citep{li:pearl19}, epidemiology \citep{robins:gre92} etc.} Suppose an economist were interested in estimating $P(y_x \mid x')$, an important $\mathcal{L}_3$-quantity called the {effect of the treatment on the treated}, or ETT \citep{heckman:rob85,heckman:rob86}. One approach to computing such quantities is through \textit{identification} \citep[~\S 3.2.4]{pearl:2k}: leveraging causal knowledge about the environment, typically a causal graph or parametric assumptions, to infer the higher-layer quantity using lower-layer data. This approach fails when the quantity is nonidentifiable, e.g. ETT in the general setting \citep{shpitser:pea09,correaetal:21}.

However, another approach uses physical experimentation to attempt to directly draw samples from the relevant distribution, $P(Y_x, X)$ in the case of ETT, and then uses statistical methods to estimate $P(Y_x=y, X=x')$. This approach is only possible if there is some sequence of physical actions by which an agent can measure these random variables simultaneously for a single unit.

\begin{wrapfigure}{r}{0.3\textwidth}
    \vspace{-0.in}
    \begin{tikzpicture}
        \node[align=center] (t1) at (0.625, -.3) {\footnotesize PCH};
        
        \node[align=center] (t1) at (-0.4, .4) {\small $\mathcal{L}_1$};
        \filldraw [fill=yellow!10,line width=0.5mm] (0, 0) rectangle (1.25,0.8);
        \draw [fill=black] (0.75, .5) circle (0.05);
        \draw [fill=black] (0.3, .6) circle (0.05);
        \node (dot1) at (0.75, .5) {\ };
        \node (py) at (1.75, 0) {\tiny $P(Y)$};
        \draw [fill=black] (0.25, .25) circle (0.05);
        \node (dot2) at (0.25, .25) {\ };
        \node (pyx) at (1.75, 0.5) {\tiny $P(Y| x)$};
        \path[-Latex] (dot1) edge[bend left=10] (pyx);
        \path[-Latex] (dot2) edge[bend left=10] (py);

        \node[align=center] (t1) at (-0.4, 1.2) {\small $\mathcal{L}_2$};
        \filldraw [fill=green!10,line width=0.5mm] (0, 0.8) rectangle (1.25,1.6);
        \draw [fill=black] (0.5, 1.4) circle (0.05);
        \draw [fill=black] (0.6, 1.05) circle (0.05);
        \node (dot3) at (0.6, 1.05) {\ };
        \node (pydox) at (2, 1.3) {\tiny $P(Y;do(x))$};
        \path[-Latex] (dot3) edge[bend left=10] (pydox);

        \draw [decorate, decoration = {calligraphic brace, raise=5pt, amplitude=5pt, aspect=0.5}](-0.5,0.3) --  (-0.5,1.3);
        \node[align=center,rotate=90] (t1) at (-1.2, 0.8) {\small realizable};

        \node[align=center] (t1) at (-0.4, 2) {\small $\mathcal{L}_3$};
        \filldraw [fill=blue!10,line width=0.5mm] (0, 1.6) rectangle (1.25,2.4);
        \draw [fill=black] (0.75, 2.2) circle (0.05);
        \draw [fill=black] (0.3, 2.2) circle (0.05);
        \node (dot4) at (0.75, 2.2) {\ };
        \node (ett) at (2, 1.80) {\tiny $P(Y_x| x')$};
        \draw [fill=black] (0.7, 1.85) circle (0.05);
        \node (dot5) at (0.7, 1.85) {\ };
        \node (pnps) at (2, 2.2) {\tiny $P(Y_x| x',y')$};
        \path[-Latex] (dot5) edge[bend left=10] (ett);
        \path[-Latex] (dot4) edge[bend left=5] (pnps);

        \draw [decorate, decoration = {calligraphic brace, raise=5pt, amplitude=5pt, aspect=0.5}](-0.5,1.8) --  (-0.5,2.2);
        \node[align=center] (t1) at (-1.2, 2) {\small ?};

        \filldraw [fill=gray!20,line width=0.5mm,pattern=north west lines] (0, 2.6) rectangle (1.25,3.1);
        \node (ett) at (1.65, 2.85) {\tiny $\mathcal{M^\star}$};

        \node[align=center] (t1) at (0.625, 3.4) {\footnotesize SCM (unknown)};

    \end{tikzpicture}
    \vspace{-0.2in}
    \caption{It is commonly assumed an agent can sample only from $\mathcal{L}_1$ and $\mathcal{L}_2$ distributions in the real world.}
    \label{fig:pch_intro}
    \vspace{-0.1in}
\end{wrapfigure}
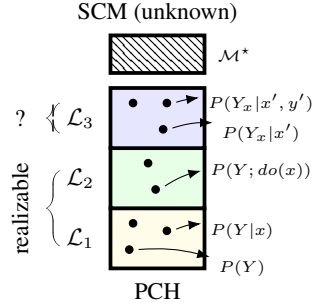

\newpage
It is generally believed to be feasible to draw samples only from $\mathcal{L}_1$- and $\mathcal{L}_2$-distributions, the latter by interventions like randomized controlled trials (RCT), à la Fisher \citep{fisher:35}, and the former by simply observing the natural behaviour of the system. $\mathcal{L}_3$-distributions like $P(Y_x, X)$ are deemed non-realizable in general, {as the potential response $Y_x$ and natural decision $X$ belong to different "worlds". Once a unit naturally adopts decision $X=x'$, $Y_x$ cannot be evaluated in the $\doo{x}$ regime for the same unit.}\footnote{{E.g., "The problem with counterfactuals like [$P(Y_x \mid x’)$] is [that] … we simply cannot perform an experiment where the same person is both given and not given treatment" \citep{shpitser:pea07}. Also, "By definition, one can never observe [counterfactuals], nor assess empirically the validity of any modeling assumptions made about them..." \citep{dawid:00}.}} However, Bareinboim, Forney \& Pearl have shown it is feasible to draw samples from the ETT distribution $P(Y_x, X)$ through a \textit{counterfactual randomization} procedure \citep{bareinboim:etal15,forney:etal17}. This leaves open the possibility that other $\mathcal{L}_3$-distributions, say perhaps $P(Y_x, X,Y)$, are also realizable through clever experimental setups, allowing one to estimate important quantities like the {probability of sufficiency}, $P(y_x\mid y',x')$ \citep{pearl:99c}.

This brings us to the central question motivating this work: \textit{from which $\mathcal{L}_3$-distributions is it possible to draw samples given fundamental physical constraints like the inability to travel back in time and subject the original unit to a different experimental condition?}
We resolve this open question with a rigorous formal treatment of the \textit{realizability of an $\mathcal{L}_3$-distribution} (Def. \ref{def:realizability}).

Our main contributions in this work are as follows:
\begin{itemize}[leftmargin=0.5cm]
    \item In Sec. \ref{sec:data_collection_procedures} we introduce a physical procedure called \textit{counterfactual randomization} (Def. \ref{def:ctf-rand}) by which an agent can gather counterfactual data, subsuming previous similar notions. 
    \item In Sec. \ref{sec:realizability_main_sec} we develop the \textbf{CTF-REALIZE} algorithm (Algo. \ref{alg:realize}) to determine whether an $\mathcal{L}_3$-distribution is physically realizable. We prove the algorithm is complete (Thm. \ref{thm:completeness}), and derive important corollaries characterizing realizable distributions (Cors. \ref{cor:maximal_amwn},\ref{cor:fpci}). For instance, we show that our main result generalizes an influential notion in the causal inference literature, known as the \textit{fundamental problem of causal inference} \citep{holland:86}.
    \item In Sec. \ref{sec:applications} we discuss important practical implications of counterfactual realizability. The traditional route of computing $\mathcal{L}_3$-quantities through identification or bounding often fails. Our work suggests opportunities for novel experiment-design ideas to directly estimate these quantities, as illustrated through Examples \hyperlink{example1}{1},\hyperlink{example2}{2} and \hyperlink{example3}{3}. More concretely,
    \begin{itemize}
        \item In Sec. \ref{sec:fairness_example}, we describe an application in causal fairness, where the naive approach of constraining a classifier using an interventional ($\mathcal{L}_2$) fairness metric fails to prevent disparities in outcomes across groups, but where a counterfactual ($\mathcal{L}_3$) approach works.
        \item In Sec. \ref{sec:bandit_example}, we show how counterfactual randomization can be used to improve RL algorithms. The baseline approach in a multi-arm bandit setting is to use allocation procedures (e.g., UCB, EXP3, Thompson Sampling) to discover which arm $x$ optimizes the expected outcome $\mathbb{E}[Y ; \doo{x}]$, which is an interventional ($\mathcal{L}_2$) strategy \citep{sutton1998reinforcement,Lattimore_Szepesvári_2020}. It turns out there are provably superior strategies (w.r.t expected outcome) based on directly optimizing counterfactual ($\mathcal{L}_3$) objectives, as we demonstrate in Example \hyperlink{example3}{3}. We prove optimality of our proposed strategy in a bandit setting with a generic causal template (Thm. \ref{thm:optimality}, Cor. \ref{cor:l3_domination}).
    \end{itemize}
\end{itemize}


Proofs and details of simulations are included in Appendices.

\vspace{-0.1in}
\paragraph{Preliminaries.} We denote variables by capital letters, $X$, and values by small letters, $x$. Bold letters, $\*X$, are sets of variables and $\*x$ sets of values. $P(\*x)$ is shorthand for $P(\*X = \*x)$. $\mathbbm{1}[.]$ is the indicator function. 

We use \textit{Structural Causal Models} (SCM) to describe the generative process for a system of interest \citep[~Def. 1]{Bareinboim2022OnPH}\citep{pearl:2k}. An SCM $\mathcal{M}$ is a tuple $\langle \*V, \*U, \mathcal{F}, P(\*u) \rangle$. $\*V$ is the set of observable variables. $\*U$ is the set of unobservable variables exogenous to the system, distributed according to $P^{\mathcal{M}}(\*U)$. $\mathcal{F} = \{f_V\}$ is a set of functions s.t. each $f_V$ causally generates the value of $V \in \*V$ as $V \gets f_V(\*U_V, \*{Pa}_V)$, where $\*U_V \subseteq \*U$ and $\*{Pa}_V \in \*V \setminus V$. 

Each $\mathcal{M}$ induces a \textit{causal diagram} $\mathcal{G}$ \citep[~Def. 13]{Bareinboim2022OnPH}, which is a graph containing a vertex for each $V \in \*V$, a directed edge from each node in $\*{Pa}_V$ to $V$, and a bidirected edge between $V, V'$ if $\*U_V, \*U_{V'}$ are not independent. Given a graph $\mathcal{G}$, $\mathcal{G}_{\overline{\*X}\underline{\*W}}$ is the result of removing edges coming
into variables in $\*X$, and edges coming out of $\*W$. We use standard terminology like parents, descendants of a node (see App. \ref{app:scm}). Our treatment is limited to \textit{recursive} SCMs, which implies acyclic diagrams, with finite discrete domains over $\*V$. 

The $\doo{\*x}$ operator indexes a sub-model $\mathcal{M}_{\*x}$ where the functions generating variables $\*X$ are replaced with constant values $\*x$. In other words, this is an intervention in the model $\mathcal{M}$ which overrides the natural mechanisms that generates $\*X$ and assign fixed values $\*x$ to these variables. A variable $Y \not \in \*X$ evaluated in this regime is called a \textit{potential response}, denoted $Y_{\*x}$.

($\*W_\star=\*w$) denotes an arbitrary counterfactual event, e.g. ($Y_x = y \land Y_{x'} = y' \land X = x''$). The probability of such an event is given by the $\mathcal{L}_3$-valuation \citep[~Def. 7]{Bareinboim2022OnPH}: \xadd{$P^{\mathcal{M}}(\*W_\star = \*w) = \sum_{\*u}\bigg(\prod_{W_\*t \in \*W_\star} \mathbbm{1}[W_\*t(\*u) = w] \bigg)P^{\mathcal{M}}(\*u),$ with $w$ taken from $\*w$.}



\section{Data-collection procedures}
\label{sec:data_collection_procedures}

In this section, we define a procedure, \textit{counterfactual randomization}, that extends the scope of traditional \textit{Fisherian} experimentation (discussed below). Consider a system of interest modeled by unknown SCM $\mathcal{M}$. Interventions and counterfactual events are typically defined in terms of \textit{symbolic} operations on $\mathcal{M}$. To conceptually separate this from the \textit{physical} constraints experienced by an agent (natural or artificial), we define the following physical actions that an agent can perform in the system. These are simply the physical counterparts to symbolic procedures.


We call each discrete episode of the system’s
behaviour a \textit{unit}. Examples of units are patients in a clinical trial, neighbourhoods in a social science experiment, rounds played on a slot machine etc. We index units w.l.o.g. by $i=1,2,3...$, which constitute a {target population} in the system.\xst{ Each unit $i$ has latent attributes $\*U^{(i)} = \*u$. Given $\*U^{(i)}$, the natural behaviour of unit $i$ is fully determined by the mechanisms in the system. An agent can interact with unit $i$ by the following physical actions.}

\medskip
\begin{definition}[Physical actions]\label{def:physical_actions}
\end{definition}
\begin{itemize}[leftmargin=0.5cm]
    \item [(1)] $\textsc{Select}^{(i)}$: randomly choosing, without replacement, a unit $i$ from the target population, to observe in the system.
    \item [(2)] {$\textsc{Read}(V)^{(i)}$: measuring the realized feature $V^{(i)}$ of unit $i$, produced by a causal mechanism $f_V \in \mathcal{F}$ operating on $i$}.
    \item [(3)] $\textsc{Rand}(X)^{(i)}$: erasing and replacing $i$'s natural mechanism $f_X$ for a decision variable $X$ with an enforced value drawn from a randomizing device having support over $\text{Domain}(X)$. $\hfill \blacksquare$
\end{itemize}


$\textsc{Read}(V)^{(i)} = v$ and $\textsc{Rand}(X)^{(i)} = x$ are also overloaded to refer to the {values} read and enforced,  respectively. $\textsc{Rand}(X)^{(i)}$ is the standard {Fisherian} randomization of a decision variable $X$, corresponding to the symbolic procedure of a \textit{stochastic} intervention \xst{$\sigma_X$ }on $X$ \citep{correa2020calculus}.\footnote{If the device used for enforcing the value of $X$ is a constant function, this action simply becomes $\textsc{Write}(X:x)^{(i)}$, corresponding to the {atomic} intervention $\doo{x}$. See Preliminaries in Sec. \ref{sec:intro}.} As $\textsc{Rand}(X)^{(i)}$ erases the unit $i$'s natural decision, $\textsc{Read}(X)^{(i)}$ will yield the value randomly assigned to unit $i$. The discovery of this procedure marked an important achievement in the history of science and experiment-design \citep{fisher:25,fisher:35}. Since the use of a randomizing device eliminates by design any confounding between the assigned decision and the unit's latent attributes $\*U^{(i)}$, it allows researchers to estimate causal effects.

It is evident that the actions in Def. \ref{def:physical_actions} are sufficient for an agent to physically draw samples from any $\mathcal{L}_1$- or $\mathcal{L}_2$-distribution\xst{. The former by simply performing $\textsc{Read}(\*V)^{(i)}$ on selected units, and the latter by a combination of $\textsc{Read}$ and $\textsc{Rand}$}, as discussed in App. \ref{app:l1l2_realizability}. Until recently, it was generally presumed these were the only physical actions possible on units in a system. However, we discuss some important extensions of experimental capabilities next.

\begin{wrapfigure}{r}{0.28\textwidth}
    \begin{center}
    \begin{tikzpicture}

        \node (X) at (-1,0) {\small $X$};
        \node[circle,draw=black,fill=black, inner sep=1.5] (Xnode) at (-0.65,0) {\ };
        \node (Y) at (2-0.5,0) {\small $Y$};
        \node[circle,draw=black,fill=black, inner sep=1.5] (Ynode) at (2-0.35-0.5,0) {\ };
        \path [-Latex] (Xnode) edge (Ynode);
        \path [Latex-Latex,dashed] (Xnode) edge[bend left=45] (Ynode);


        \node (X) at (-1,-1.75+0.5) {\small $X$};
        \node[circle,draw=black,fill=black, inner sep=1.5] (Xnode) at (-0.65,-1.75+0.5) {\ };
        \node[orange] (x1) at (-0.4,-0.625) {\small $x$};
        \node (Y) at (2-0.5,-1.75+0.5) {\small $Y$};
        \node[circle,draw=black,fill=black, inner sep=1.5] (Ynode) at (2-0.35-0.5,-1.75+0.5) {\ };
        \node[fill=orange,draw,inner sep=0.2em, minimum width=0.2em] (intervention) at (-0.65,-0.625) {\ };
        \path [-Latex] (intervention) edge (Xnode);
        \path [-Latex] (Xnode) edge (Ynode);

        \node (X) at (-1,-1.75+0.5-1.25) {\small $X$};
        \node[circle,draw=black,fill=black, inner sep=1.5] (Xnode) at (-0.65,-1.75+0.5-1.25) {\ };
        \node[orange] (x1) at (0,-1.45+0.5-1.25) {\small $x$};
        \node (Y) at (2-0.5,-1.75+0.5-1.25) {\small $Y$};
        \node[circle,draw=black,fill=black, inner sep=1.5] (Ynode) at (2-0.35-0.5,-1.75+0.5-1.25) {\ };
        \node[fill=orange,draw,inner sep=0.2em, minimum width=0.2em] (intervention) at (0,-1.75+0.5-1.25) {\ };
        \path [-Latex] (intervention) edge (Ynode);
        \path [Latex-Latex,dashed] (Xnode) edge[bend left=55] (Ynode);
        
    \end{tikzpicture}
    \end{center}
    \vskip -0.1in
    \caption{(Top) Causal diagram with decision variable $X$; (Middle) Fisherian randomization of overriding the unit's natural decision $X$ and assigning a fixed value; (Bottom) Randomizing the actual decision affecting $Y$ without erasing the unit's natural decision $X$.}
    \label{fig:natural_choice}
    \vskip -0.2in
    \end{wrapfigure}
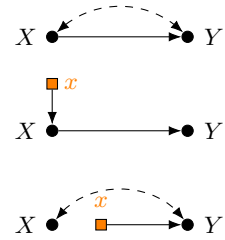


\paragraph{Counterfactual data-collection procedures.} In an \xst{important}\xadd{early} work from the causal reinforcement learning \xst{(CRL) }literature, Bareinboim, Forney \& Pearl describe an experimental setting in which it is possible to both randomize a unit's actual decision, and also record the natural decision the unit \textit{would have normally} taken \citep{bareinboim:etal15, forney:etal17}. Subsequently, this procedure has been used to establish benchmarks in counterfactual decision making \citep{zhang22a}. 
These settings involve an agent introspecting to gauge their natural choice, or otherwise revealing their natural choice by some indication, e.g. physical gestures prior to decision-time. Importantly, this form of randomization does not erase the unit's natural choice of decision variable $X$, as schematically illustrated in Fig. \ref{fig:natural_choice}.

Building on the idea, we \xadd{formalize this into a more general} \xst{note a natural}extension of the agent's capabilities: the ability to intervene on a variable $X$'s value \textit{as perceived} by its causal children. To illustrate this, consider the $\mathcal{L}_3$-quantity known as {natural direct effect}, or NDE, which is used in mediation analysis to measure the effect of $X$ on $Y$ via a "direct" path, as opposed to an "indirect" path via a mediator $Z$ \citep{pearl:01} -- highly relevant in several fields, as discussed in Sec. \ref{sec:intro}. The NDE is generally considered as identifiable from experimental data only under certain conditions \citep{pearl:01,correaetal:21}. The following example details an experiment design where it is possible to compute the NDE even when these identification conditions are not met, by randomizing the \textit{perception} of $X$.

\vspace{-0.1in}

\paragraph{Example 1 (\xst{Traffic Camera}\xadd{Mediation analysis}).} \hypertarget{example1}{ }A computer vision company's tool is being evaluated for an automated speeding ticket system that uses footage from traffic cameras. But the government's audit team has a concern: it is possible the model is trained on footage with a strong correlation between the color of the car and speeding (perhaps due to color preference of different socioeconomic neighbourhoods), and unfairly penalizes certain car colors. 

\begin{wrapfigure}{r}{0.3\textwidth}
    \vspace{-0.3in}
    \begin{center}
    \begin{tikzpicture}

        \node (X) at (-1,0) {\small $X$};
        \node[circle,draw=black,fill=black, inner sep=2-0.5] (Xnode) at (-0.65,0) {\ };
        \node[orange] (W) at (0.2,0.3) {\small $W$};
        \node[circle,draw=black,fill=orange, inner sep=2-0.5] (Wnode) at (0.2,0) {\ };
        \node (Z) at (0.,-0.85) {\small $Z$};
        \node[circle,draw=black,fill=black, inner sep=2-0.5] (Znode) at (0.5,-0.75) {\ };
        \node (Y) at (2,0) {\small $Y$};
        \node[circle,draw=black,fill=black, inner sep=2-0.5] (Ynode) at (2-0.35,0) {\ };
        \path [-Latex] (Xnode) edge (Wnode);
        \path [-Latex] (Wnode) edge (Ynode);
        \path [-Latex] (Znode) edge (Ynode);
        \path [-Latex] (Xnode) edge (Znode);
        \path [Latex-Latex,dashed] (Znode) edge[bend right=40] (Ynode);

        \node (X) at (-1,-2+0.3) {\small $X$};
        \node[circle,draw=black,fill=black, inner sep=2-0.5] (Xnode) at (-0.65,0-2+0.3) {\ };
        \node[fill=orange,draw,inner sep=0.2em, minimum width=0.2em] (intervention) at (0.2,-2+0.3) {\ };
        \node (Z) at (0.,-2.85+0.3) {\small $Z$};
        \node[circle,draw=black,fill=black, inner sep=2-0.5] (Znode) at (0.5,-2.75+0.3) {\ };
        \node (Y) at (2,0-2+0.3) {\small $Y$};
        \node[circle,draw=black,fill=black, inner sep=2-0.5] (Ynode) at (2-0.35,0-2+0.3) {\ };
        \path [-Latex] (intervention) edge (Ynode);
        \path [-Latex] (Znode) edge (Ynode);
        \path [-Latex] (Xnode) edge (Znode);
        \node[orange] (x1) at (0.2,-1.7+0.3) {\small $x$};
        \path [Latex-Latex,dashed] (Znode) edge[bend right=40] (Ynode);
      
    \end{tikzpicture}
    \end{center}
    \vskip -0.1in
    \caption{(a) "Expanded" diagram for Example 1, where $W$ is \textit{counterfactual mediator} for $X$; (b) Randomizing the value of $X$ as perceived by $Y$.}
    \label{fig:trafficcamera_example}
    \vskip -0.7in
    \end{wrapfigure}
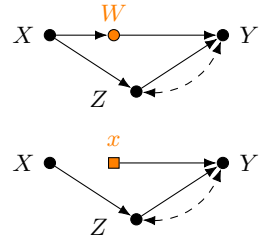

This amounts to a hypothesis that $X$ (car's color) affects $Y$ (AI decision to issue a ticket) via a direct path as opposed to the indirect path via $Z$ (speeding). \xadd{The indirect path describes
the causal effects of, say, how pedestrians and other drivers react to
a red car and affect its speeding. }This hypothesis is true iff NDE is measured to be non-0, where NDE is defined as the following expression: $\text{NDE}_{x,x'}(y) = P(y_{x'Z_{x}}) - P(y_{x})$ \citep{pearl:01}. The second term, $P(y_x)$, can be estimated from a Fisherian randomization of $X$ (say, an experiment recruiting drivers and assigning them random cars). Inconveniently, the first term, $P(y_{x'Z_x})$, is nonidentifiable for Fig. \ref{fig:trafficcamera_example}(a), even using RCT data. So it is unclear how to make progress with this hypothesis test.

However, the audit team recognizes there exists a special mediator, viz. the features $W$ in the video\xst{ footage} which reveal the car's color to the model (say, \xst{the }RGB values of \xst{the }pixels in the video frames). They use standard video-editing tools to randomly swap the color of the car in the footage. By randomly assigning a particular car $W \gets \text{red}$, they are able to affect the mechanism $f_Y$'s \textit{perception} of $X$:
    \begin{align}
        & \!\!\!P(Y_{W=\text{red}}\mid X = \text{blue}) &\text{est. from $\mathcal{L}_2$ data}\\
        =&P(Y_{W=\text{red},Z}\mid X = \text{blue}) &\text{$Z:$ natural value}\\
        =&P(Y_{W=\text{red},Z_{X=\text{blue}}}\mid X = \text{blue}) &\text{consistency property}\\
        =&P(Y_{X=\text{red},Z_{X=\text{blue}}}\mid X = \text{blue}) & \text{Def.}~\ref{def:ctf_mediator_informal}, X \equiv W \label{eq:l2_proxy_traffic}  \\ 
        =& P(Y_{X=\text{red},Z_{X=\text{blue}}}) & \text{d-separation} 
    \end{align}
Eq. \ref{eq:l2_proxy_traffic} is justified because $W$ controls $Y$'s perception of $X$ given a fixed $z$ (formalized in Lemma \ref{lem:l2_proxy}). Thus, they are able to directly sample from the $\mathcal{L}_3$-distribution $P(Y_{x'Z_x},X)$ via a physical procedure, and use identification rules to obtain $P(y_{x'Z_x})$. Using the formula for NDE, they can evaluate whether a car's color has a {direct} effect on the odds of getting a speeding ticket.\xst{ A detailed version of this example, including a discussion of assumptions involved, is in App. \ref{app:traffic_camera}.} $\hfill \blacksquare$

Here, one is able to randomize $X$ as {perceived} by one of its children, by leveraging the variable $W$ (RGB values) that fully encodes information about $X$ (color) and mediates its effect on $Y$. \xadd{We capture this intuition with the following (informal) definition.

\begin{definition}[Counterfactual mediator (informal)] \label{def:ctf_mediator_informal}

We call $W$ a \textit{counterfactual mediator} of $X$ w.r.t $Y \in Ch(X)$ if the value of $X$ can be retrieved from $W$ by the mechanism generating $Y$.   $\hfill \blacksquare$
\end{definition}}

Other examples of interventions on perceived attributes via counterfactual mediators include changing details on a job application (name, pronouns, keywords) to simulate a perceived alternate demographic identity \citep{bertrand:03}, or editing specific portions of text input to a language model \citep{feder-etal-2022-causal}. Randomizing perception has been discussed in \citet[~\S 4.4.4]{pearl:etal16-primer}. For a detailed discussion of the causal semantics of intervening on perceptions, and the related literature, see \citep[~App. D.1]{plecko:bareinboim24}. For the interested reader, we provide a rigorous treatment in App. \ref{app:l3_conditions}\xst{ of the structural assumptions involved}, including a formal Def. \ref{def:l2_proxy} of a {counterfactual mediator}.


This important extension to experimental capabilities is captured in the following definition of a new physical action that an agent might be able to perform in an environment.

\begin{definition}[Counterfactual (ctf-) randomization]\label{def:ctf-rand}
    
    $\textsc{ctf-Rand}(X \rightarrow \*C)^{(i)}$: fixing the value of $X$ \textit{as an input to} the mechanisms generating $\*C \subseteq Ch(X)_{\mathcal{G}}$ using a randomizing device having support over Domain($X$), for unit $i$, given causal diagram $\mathcal{G}$. $\hfill \blacksquare$
\end{definition}

The key differences between the Fisherian $\textsc{Rand}(X)^{(i)}$ and $\textsc{ctf-Rand}(X \rightarrow \*C)^{(i)}$ are (1) $\textsc{ctf-Rand}$ does not erase the unit $i$'s natural decision $X^{(i)}$\xst{\footnote{Another way of understanding this difference is that the unit's natural inclination is taken into account. }}; and (2) while $\textsc{Rand}$ affects all children of $X$, $\textsc{ctf-Rand}$ does not affect $Ch(X) \setminus \*C$. $\textsc{ctf-Rand}$ can only be enacted under certain structural conditions, viz., either in environments which permit the measurement of a unit's natural decision while simultaneously randomizing the actual decision \citep{bareinboim:etal15}, or where counterfactual mediators can be used to alter $X$ as perceived by a subset of children. \xadd{Whether the agent is indeed able to perform this action thus depends on the specific experimental setting. }\xst{We provide a systematic way to translate such structural conditions into a list of the $\textsc{ctf-Rand}$ procedures possible in the given environment, in Algo. \ref{alg:ctf_procedures}. }

\xadd{Note: Def. \ref{def:ctf-rand} implies that it is possible to physically perform multiple randomizations involving the same variable $X$ on a single unit $i$, with each intervention affecting a different subset of children. }\xst{Ctf-randomization makes it possible to physically perform multiple randomizations involving the same variable $X$ on a single unit $i$ (explained with example in App. \ref{app:multiple_rands}).} Further, $\textsc{ctf-Rand}$ may only be performed w.r.t a graphical child variable; it is not possible to bypass a child and directly affect a descendant's perception of $X$\xst{ (justified in App. \ref{app:ctf_rand_constraints})}.


\section{Counterfactual realizability}
\label{sec:realizability_main_sec}

Given the possibility of performing ctf-randomization (Def. \ref{def:ctf-rand}), we are interested in knowing which $\mathcal{L}_3$-distributions can be accessed directly by experimentation. In this section, we discuss the constraints imposed by nature on an agent\xst{, notably the \textit{fundamental constraint of experimentation} (FCE)}. We then formally define \textit{realizability} and develop a complete algorithm to determine whether an $\mathcal{L}_3$-distribution is realizable.


The most \xst{important }\xadd{basic }constraints experienced by the agent (natural or artificial) are physical. Each mechanism $f_V \in \mathcal{F}$ represents some physical process that transforms a unit $i$ according to the laws of nature. For instance, taking a drug, $X$, produces a side effect in the patient, $Y$, by a biochemical reaction $f_Y(X, U_Y)$, which depends on the drug and the patient's latent health condition, $U_Y$. Once patient $i$ has been subjected to mechanism $f_Y$ under $X=x$, there appears to be no way to go back in time and subject the same patient to mechanism $f_Y$ under $X=x'$. Even if technologically feasible to reverse the process (e.g., by taking an antidote to the drug\xst{, or simply by waiting for the drug's active ingredient to leave the body}), the latent factors $\*U=\*u$ might have changed after the experiment (e.g., the patient could have developed tolerance to the drug). Repeating the experiment on this patient is tantamount to testing a \textit{new} unit with unknown latent features $\*U = \*u'$.\footnote{In the philosophy of science literature, similar ideas have been discussed under the topic of the temporal asymmetry of causation \citep[~\S III-IV]{reichenbach:56}.} This observation is made more formal through the following assumption. 

\begin{assumption}[Fundamental constraint of experimentation (FCE)]\label{assn:fce}
    A unit $i$ in the target population can physically undergo a causal mechanism $f_V \in \mathcal{F}$ at most once. $\hfill \blacksquare$
\end{assumption}

\begin{remark}
    The FCE assumption entails that a unit $i$ can only be submitted to a particular mechanism $f_V(\*{Pa}_V,\*U_V)$ under a single set of experimental conditions, received as input to $f_V$. By implication, the physical actions in Defs. \ref{def:physical_actions}, \ref{def:ctf-rand} can only be performed at most once per unit $i$. $\hfill \blacksquare$
\end{remark}

Once unit $i$ has been subjected to $f_V$, it is not possible to re-run $f_V$ with differently fixed inputs. $\textsc{Read}(V)^{(i)}$ thus only yields one value for $i$. Although ctf-randomization permits multiple interventions involving the same variable $X$\xst{ (App. \ref{app:multiple_rands})}, each such intervention can only be performed once, since it impacts different child mechanisms that can each only occur once for unit $i$. We also assume that the agent can only perform the physical actions in Defs. \ref{def:physical_actions}, \ref{def:ctf-rand}, up to isomorphism. 




\begin{definition}[I.i.d sample]\label{def:iid}
    Given an $\mathcal{L}_3$-distribution $Q = P(\*W_\star)$ and a sequence of physical actions $\mathcal{A}^{(i)}$ performed on unit $i$ in an environment modeled by SCM $\mathcal{M}$, producing a vector of realized values $\*W_\star^{(i)} = \*w$
    for the variables in $\*W_\star$, the vector is said to be an \textit{i.i.d sample} from $Q$ if \xadd{$P^{\mathbb{C}}(\*W_\star^{(i)} = \*w \mid \mathcal{A}^{(i)}) = P^{\mathcal{M}}(\*W_\star = \*w), \forall \*w,$} where $P^{\mathbb{C}}$ is the probability measure over the beliefs of the acting agent $\mathbb{C}$, and the l.h.s is the probability of physical actions $\mathcal{A}^{(i)}$ producing the vector $\*w$ when performed on some unit $i$.\hfill  $\blacksquare$
\end{definition}

\begin{definition}[Realizability] \label{def:realizability}
    Given a causal diagram $\mathcal{G}$ and the set of physical actions $\mathbb{A}$, an $\mathcal{L}_3$-distribution $P(\*W_\star)$ is \textit{realizable given $\mathbb{A}$ and $\mathcal{G}$} iff there exists a sequence of actions $\mathcal{A}$ from $\mathbb{A}$ by which an agent can draw an i.i.d sample (Def. \ref{def:iid}) from $P^{\mathcal{M}}(\*W_\star)$, for any $\mathcal{M} \in M(\mathcal{G})$, the class of SCMs compatible with $\mathcal{G}$. $\hfill \blacksquare$
\end{definition}

We emphasize the \textit{distinction between {realizability} and {identifiability}}. {Identifiability} \citep[~Def. 3.2.3]{pearl:2k} from $\mathcal{G}$ states that a distribution (say, $P(\*v; \doo{x})$) can be uniquely computed from the available data (say, $P(\*v)$) for any SCM compatible with the assumptions in $\mathcal{G}$. {Realizability} of a distribution states that it is physically possible for an agent to actually gather data samples according to this distribution.


\begin{wrapfigure}{r}{0.40\textwidth}
\vspace{-0.25in}
\begin{center}
\begin{tikzpicture}

        \node (T) at (0,0) {\small $T$};
        \node[red] (t) at (-0.5,-0.5) {\small $t$};
        \node[teal] (Nat) at (0.75,-0.5) {\small nat.};
        \node (A) at (0,-1) {\small $A$};
        \node (X) at (2-0.2-0.2,0) {\small $X$};
        \node (W) at (0,-2) {\small $W$};
        \node (Z) at (2-0.2-0.2,-2) {\small $Z$};
        \node (dot1) at (0.25,0) { };
        \node (dot2) at (0.25,-1) { };
        \node (dot3) at (-0.25,0) { };
        \node (dot4) at (-0.25,-1) { };
        \node (dot5) at (-0.25,-2) { };
        \node (dot6) at (1.75-0.2-0.2,-1.7) { };
        \path [->] (T) edge (A);
        \path [->] (A) edge (W);
        \path [->] (A) edge (Z);
        \path [->] (X) edge (Z);
        \path [<->,dashed] (W) edge[bend right=20] (Z);
        \path [-Latex, teal] (dot1) edge (dot2);
        \path [-Latex, teal] (dot2) edge (dot6);
        \path [-Latex, red] (dot3) edge (dot4);
        \path [-Latex, red] (dot4) edge (dot5);

        \node (T) at (4-0.75-0.2,0) {\small $T$};
        \node[red] (t) at (3.5-0.75-0.2,-0.5) {\small $t$};
        \node[teal] (Nat) at (5.25-0.75-0.1-0.4,-0.5) {\small nat.};
        \node (X) at (6-0.75-0.2-0.4,0) {\small $X$};
        \node (W) at (4-0.75-0.2,-2) {\small $W$};
        \node (Z) at (6-0.75-0.2-0.4,-2) {\small $Z$};
        \node (dot1) at (4.25-0.75-0.3,0) { };
        \node (dot3) at (3.75-0.75-0.2,0) { };
        \node (dot5) at (3.75-0.75-0.2,-2) { };
        \node (dot6) at (5.75-0.72-0.2-0.4,-1.52) { };
        \path [->] (T) edge (W);
        \path [->] (T) edge (Z);
        \path [->] (X) edge (Z);
        \path [<->,dashed] (W) edge[bend right=20] (Z);
        \path [-Latex, teal] (dot1) edge (dot6);
        \path [-Latex, red] (dot3) edge (dot5);

    \end{tikzpicture}
\end{center}
\vskip -0.1in
\caption{Testing realizability of $P(Z_x, W_t)$ for $\mathcal{G}_1$ (left) and $\mathcal{G}_2$ (right). $\mathcal{G}_1$ yields conflicting requirements.}
\label{fig:realize_examples}
\vskip -0.in
\end{wrapfigure}
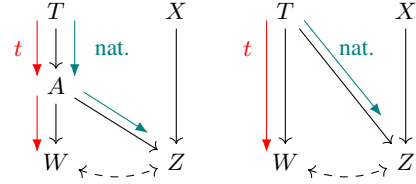

We next develop an algorithm to decide whether a distribution is realizable. As an intuition pump, suppose that an agent is able to perform $\textsc{ctf-Rand}(V \rightarrow C), \forall V, C \in Ch(V)$, w.r.t an input causal diagram, and wants to obtain samples from\xst{the distribution} $P(Z_x, W_t)$. Consider the diagram $\mathcal{G}_2$ in Fig. \ref{fig:realize_examples}. By performing $\textsc{ctf-Rand}(T \rightarrow W)$ and $\textsc{ctf-Rand}(X \rightarrow Z)$, the distribution is realizable. However, suppose the input diagram \xst{according to which the agent can perform $\textsc{ctf-Rand}$ }is $\mathcal{G}_1$. A necessary condition to measure $Z_x$ for a unit is for mechanism $f_A$ to receive the natural value of $T$, illustrated in green. \xadd{While a}\xst{This conflicts with the} necessary condition to {simultaneously }measure $W_t$\xst{, requiring}\xadd{ is for} $f_W$ to receive $A_t$, which in turn requires $f_A$ to receive a fixed $t$, shown in red. This conflict in necessary conditions renders the query non-realizable.\footnote{To be clear, the input to the algorithm is a graph and an accurate set of actions the agent can perform in the environment. If the graph is per $\mathcal{G}_1$ in Fig. \ref{fig:realize_examples}, then $\textsc{ctf-Rand}(T \rightarrow Z)$ is not possible in this environment. Marginalizing out $A$ and providing graph $\mathcal{G}_2$ as input does not help.\xst{See Remark \ref{rem:nobypassing} and App. \ref{app:ctf_rand_constraints}.}}


\begin{algorithm}[t]
\caption{CTF-REALIZE}
\label{alg:realize}
\begin{multicols}{2}
\begin{algorithmic}[1]
   \STATE {\bfseries Input:} $\mathcal{L}_3$-distribution $Q = P(\*W_\star)$; causal diagram $\mathcal{G}$; action set $\mathbb{A}$
   \STATE {\bfseries Output:} I.i.d sample $\*W_\star^{(i)}$ from $Q$; FAIL if $Q$ is not realizable given $\mathcal{G}, \mathbb{A}$
   
   \smallskip
   \STATE Fix a topological ordering Top($\mathcal{G}$)
   \STATE $\textsc{Select}^{(i)}$ for a new unit $i$
   
   \smallskip
   \FOR{$V$ in order Top($\mathcal{G})$} 

   \smallskip
   \STATE $\text{INT}_V \gets \emptyset$ \COMMENT{Interventions for $V$}
   \STATE $\text{OUTPUT}_V \gets \emptyset$ \COMMENT{Index in output vector}

   \smallskip
   \FOR{each term $W_{\*t}$ in expression $\*W_\star$}

   \IF{$V \in An(W)_{\mathcal{G}_{\overline{\*T}}}$ \AND $V \neq W$}
   \STATE Call \textbf{COMPATIBLE}($V, W_{\*t}$) Alg. \ref{alg:subroutine}
   \ENDIF
   \IF{$V = W$}
   \STATE Add $\{W_\*t\}$ to $\text{OUTPUT}_V$
   \ENDIF
   \ENDFOR

   \FOR{each $\{$action : tag$\}$ $\in \text{INT}_V$}
   \STATE Perform the randomization on unit $i$
   \STATE If the random-generated value $\neq$ tag, discard the unit and return to Line 4
   \ENDFOR
   
    \FOR{each $W_t \in \text{OUTPUT}_V$}
        \IF{$\{\textsc{Rand}(V): .\} \in \text{INT}_V$ }
            \STATE Return FAIL
        \ELSE
            \STATE Perform $\textsc{Read}(V)^{(i)} = v'$
            \STATE Assign $v'$ to each index $W_t^{(i)}$ in output vector $\*W_\star^{(i)}=\*w$
        \ENDIF
    \ENDFOR
   
   \smallskip
   \ENDFOR

   \smallskip
   \STATE Return i.i.d sample $\*W_\star^{(i)}=\*w$
\end{algorithmic}
\end{multicols}
\end{algorithm}

This "edge-coloring" intuition is formalized in Algo. \ref{alg:realize}. The algorithm \textbf{CTF-REALIZE} takes as input an $\mathcal{L}_3$-distribution $P(\*W_\star)$, a graph $\mathcal{G}$, and a set of physical actions $\mathbb{A}$ the agent is able to perform in the environment (viz., the $\textsc{Rand}$ and $\textsc{ctf-Rand}$ actions which are possible in the environment). It returns an i.i.d sample if the distribution is realizable, and FAIL otherwise. 

The algorithm works as follows (a more detailed walk-through is presented in App. \ref{app:algo_walkthrough}). The necessary and sufficient conditions to measure each potential response $W_t \in \*W_\star$ are {[i]} $T$ is fixed as $t$ (by intervention) as an input to all children $C \in Ch(T) \cap An(W)$; {[ii]} each $A \in An(W)_{\mathcal{G}_{\overline{T}}}$, $A \not \in \{T, W\}$ is received "naturally" (i.e., without intervention) by its children $C \in Ch(A) \cap An(W)$; and [iii] $f_W$ is not erased and overwritten (by Fisherian intervention). If these conditions can be met for all the terms in $\*W_\star$, the distribution is realizable. If there is a conflict in the necessary conditions for evaluating two terms (as we saw for $P(Z_x, W_t)$ in Fig. \ref{fig:realize_examples}, $\mathcal{G}_1$), the query is non-realizable.

The algorithm is fully general and does not make assumptions about the ability to perform any particular interventions. If the agent cannot perform any counterfactual randomization, the algorithm returns FAIL for non-$\mathcal{L}_2$ queries. If the agent cannot perform any interventions at all, the algorithm returns FAIL for non-$\mathcal{L}_1$ queries (we assume the ability to $\textsc{Read}$ all variables).\xadd{ Details about the time and space complexity of Algo. $\ref{alg:realize}$ are provided in App. \ref{app:realize_complexity}, for the interested reader.}


\begin{theorem}[Correctness and Completeness] \label{thm:completeness}
    An $\mathcal{L}_3$-distribution $Q = P(\*W_\star)$ is realizable given action set $\mathbb{A}$ and causal diagram $\mathcal{G}$ iff the algorithm \textbf{CTF-REALIZE}($Q, \mathcal{G}, \mathbb{A}$) returns a sample. $\hfill \blacksquare$
\end{theorem}

A further question we may ask is which $\mathcal{L}_3$-distributions are realizable if we assume maximum experimental capabilities, notably, the ability to perform separate ctf-randomization for each child of each variable. Given a causal diagram $\mathcal{G}$, we define the \textit{maximal feasible action set} $\mathbb{A}^\dag(\mathcal{G})$ as the set containing all of the following actions: $\textsc{Select}^{(i)}$, $\textsc{Read}(V)^{(i)}$  $, \forall V$, and $\textsc{ctf-Rand}(X \rightarrow C)^{(i)}$ $, \forall X$ and $C \in Ch(X)$. $\mathbb{A}^\dag(\mathcal{G})$ thus gives the agent the most granular interventional capabilities.


\begin{definition}[Ancestors of a counterfactual \citep{correaetal:21}] \label{def:ctf_ancestor} Given a causal diagram $\mathcal{G}$ and a potential response $Y_\*x$, the set of (counterfactual) ancestors of $Y_\*x$, denoted $An(Y_\*x)$, consists of each $W_\*z$ s.t. $W \in An(Y)_{\mathcal{G}_{\underline{\*X}}}$, and $\*z = \*x \cap An(W)_{\mathcal{G}_{\overline{\*X}}}$. For a set $\*W_\star$, $An(\*W_\star)$ is defined to be the union of the ancestors of each potential response in the set. $\hfill \blacksquare$
\end{definition}


\begin{corollary} \label{cor:maximal_amwn}
    An $\mathcal{L}_3$-distribution $Q = P(\*W_\star)$ is realizable given causal diagram $\mathcal{G}$ and action set $\mathbb{A}^\dag(\mathcal{G})$ iff the ancestor set $An(\*W_\star)$ does not contain a pair of potential responses $W_{\*t}, W_{\*s}$ of the same variable $W$ under different regimes. $\hfill \blacksquare$
\end{corollary}

For instance, if $\*W_\star = \{Z_x, W_t\}$ w.r.t graph $\mathcal{G}_1$ in Fig. \ref{fig:realize_examples}, then $An(\*W_\star) = \{Z_x, A, T, W_t, A_t\}$, which contains both $A, A_t$. Thus, $P(\*W_\star)$ is not realizable even with maximal experimentation capabilities. In App. \ref{app:realize_examples}, we provide further examples of using the \textbf{CTF-REALIZE} algorithm, and the \xst{ctf-ancestor }graphical criterion, to demonstrate the realizability of the ETT distribution $P(Y_x,X)$, the non-realizability of the probability of sufficiency distribution $P(Y_x,X,Y)$.

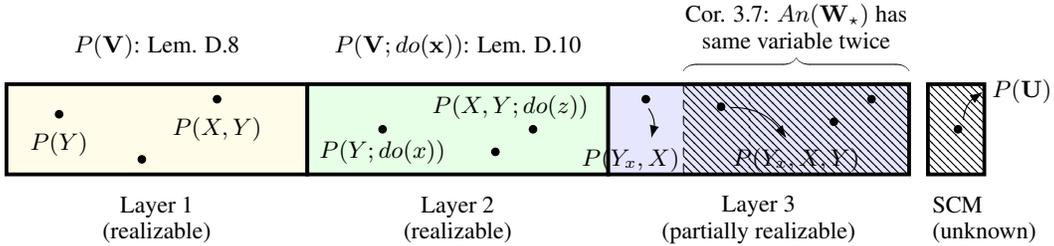
\begin{figure}[t]
\centering
\begin{tikzpicture}
    
    \filldraw [fill=yellow!10,line width=0.5mm] (-5, 0) rectangle (-1,-1.5+0.3);
    \draw [fill=black] (-4.3, -.4) circle (0.05);
    \draw [fill=black] (-3.2, -1) circle (0.05);
    \node[font=\small] (t1) at (-4.3, -.8) {$P(Y)$};
    \draw [fill=black] (-2.2, -0.2) circle (0.05);
    \node[font=\small] (t1) at (-2.2, -0.6) {$P(X,Y)$};

    \filldraw [fill=green!10,line width=0.5mm] (-1, 0) rectangle (3,-1.5+0.3);
    \draw [fill=black] (0, -.6) circle (0.05);
    \draw [fill=black] (2, -.6) circle (0.05);
    \node[font=\small] (t1) at (0, -0.9) {$P(Y;do(x))$};
    \draw [fill=black] (1.5, -0.9) circle (0.05);
    \node[font=\small] (t1) at (1.7, -0.3) {$P(X,Y;do(z))$};

    \filldraw [fill=blue!10,line width=0.5mm] (3, 0) rectangle (7,-1.5+0.3);
    \draw [fill=black] (3.5, -0.2) circle (0.05);
    \node (dot5) at (3.5, -0.2) {\ };
    \node[font=\small] (ett) at (3.3, -1.) {$P(Y_x,X)$};
    \path[-Latex] (dot5) edge[bend left=30] (ett);
    
     \draw [decorate, decoration = {calligraphic brace, raise=5pt, amplitude=5pt, aspect=0.5}](4,0) --  (7,0)
    node[align = center,pos=0.5,above=10pt,black,font=\small]{Cor. \ref{cor:maximal_amwn}: $An(\*W_\star)$ has 
    \\ same variable twice};
    \filldraw [fill=blue!10,line width=0.005mm,dashed,pattern=north west lines] (4, 0) rectangle (7,-1.5+0.3);
    \draw [fill=black] (4.5, -0.3) circle (0.05);
    \draw [fill=black] (6, -.5) circle (0.05);
    \draw [fill=black] (6.5, -0.2) circle (0.05);
    \node (dot7) at (4.5,-0.3) {\ };
    \node[font=\small] (pnps) at (5.5, -1.) {$P(Y_x,X,Y)$};
    \path[-Latex] (dot7) edge[bend left=30] (pnps);

    \filldraw [fill=gray!20,line width=0.5mm,pattern=north west lines] (7.25, 0) rectangle (8,-1.5+0.3);
    \draw [fill=black] (7.65, -0.6) circle (0.05);
    \node (dot10) at (7.65, -0.6) {\ };
    \node[font=\small] (unit) at (8.5, -0.1) {$P(\*U)$};
    \path[-Latex] (dot10) edge[bend left=30] (unit);
    
    \node[align=center,font=\small] (t1) at (-3, -1.8) {Layer 1\\(realizable)};
    \node[align=center,font=\small] (t1) at (-3, .5) {$P(\*V)$: Lem. \ref{lem:l1_pv}};
    \node[align=center,font=\small] (t1) at (1, -1.8) {Layer 2\\(realizable)};
    \node[align=center,font=\small] (t1) at (1, .5) {$P(\*V;do(\*x))$: Lem. \ref{lem:l2_pv}};
    \node[align=center,font=\small] (t1) at (5, -1.8) {Layer 3\\(partially realizable)};
    \node[align=left,font=\small] (t1) at (8, -1.8) {SCM\\(unknown)};

\end{tikzpicture}
\vskip -0.in
\caption{Pearl Causal Hierarchy (PCH) induced by an unknown SCM $\mathcal{M}$. An $\mathcal{L}_3$-distribution is realizable given a graph $\mathcal{G}$ and the maximal feasible action set $\mathbb{A}^\dag(\mathcal{G})$ iff the ancestor set $An(\*W_\star)$ does not contain the same variable under different regimes.}
\vskip -0.in
\label{fig:pch_realizability}
\end{figure}

\xadd{We believe this is an important contribution to  causal inference. Cor. \ref{cor:maximal_amwn} provides a graphical criterion to delineate how far up the PCH an agent can go via experimental methods, {in principle}. Often, counterfactuals have been criticized as being hypothetical, untestable, or unscientific assumptions.
Our analysis counters this claim, as summarized in Fig. \ref{fig:pch_realizability}.}

\begin{corollary} [Fundamental problem of causal inference (FPCI) \citep{holland:86}] \label{cor:fpci}
    The distribution $Q = P(Y_x, Y_{x'})$ is not realizable given maximal feasible action set $\mathbb{A}^\dag(\mathcal{G})$, for any causal diagram $\mathcal{G}$, and any variables $X, Y \in Desc(X)$. $\hfill \blacksquare$
\end{corollary}

The FPCI is an influential notion in the literature, and is often taken as a primitive, or in an axiomatic fashion. We show that it is rather a {specific} consequence of the more general FCE assumption \ref{assn:fce}, and follows from Thm. \ref{thm:completeness} and Cor. \ref{cor:maximal_amwn}. By itself, the FPCI does not translate to an operational criterion for determining which $\mathcal{L}_3$-distributions are realizable (Def. \ref{def:realizability}). For instance, it does not clarify that a distribution with potential responses under conflicting regimes like $P(Y_x, Z_{x'})$ may indeed be realizable via counterfactual randomization, as we show in Example \hyperlink{example2}{2}. It also does not tell us that $P(Z_x,W_t)$ may be realizable given causal diagram $\mathcal{G}_2$ in Fig. \ref{fig:realize_examples}, but not realizable given $\mathcal{G}_1$.

\section{Applications: Counterfactual Decision-Making and Fairness}
\label{sec:applications}

Next, we highlight the practical relevance of our results with some concrete use-cases. We already discussed in Example \hyperlink{example1}{1} how realizability can be used to design experiments for performing \textbf{mediation analysis} of direct and indirect effects, an important task in several fields. We now discuss applications in \textbf{causal fairness analysis} and \textbf{causal reinforcement learning (RL)}. Our goal is to underscore that the standard/baseline approaches in these areas, even among approaches that incorporate counterfactual reasoning, typically use observational ($\mathcal{L}_1$) or interventional ($\mathcal{L}_2$) data only, whereas a counterfactual ($\mathcal{L}_3$) data-collection approach can lead to demonstrably better results. Due to space constraints, we include in App. \ref{app:examples} the full specification of SCMs used and algorithms implemented.

\subsection{Causal fairness - using counterfactual data for fairer decisions}
\label{sec:fairness_example}

Causal fairness analysis is a burgeoning field and a full survey is beyond the scope of this paper (see, e.g., \citet{plecko:bareinboim24} for a review of related works). We limit our discussion to an example where counterfactual realizability is directly relevant. 

\begin{wrapfigure}{r}{0.4\textwidth}
\vspace{-0.2in}
\begin{center}
\includegraphics[width=0.35\textwidth]{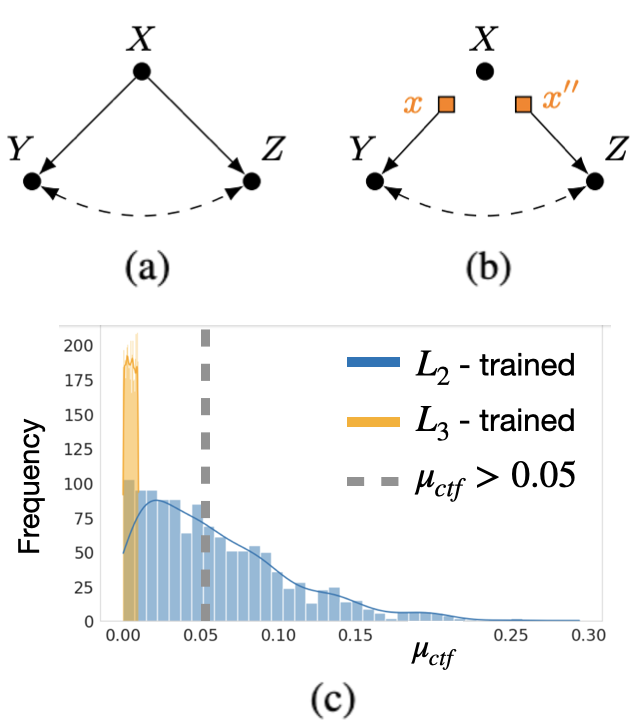}
\end{center}
\vskip -0.2in
\caption{(a) Causal diagram for Example 2; (b) $P(Y_x, Z_{x'})$ is realizable using the interventions $\textsc{ctf-Rand}(X \rightarrow Y)$ and $\textsc{ctf-Rand}(X \rightarrow Z)$; (c) Histogram of 1000 classifiers trained on $\mathcal{L}_2$ (blue) and $\mathcal{L}_3$ (orange) fairness measures. $\mathcal{L}_2$ classifiers show statisically significant discrimination ($\mu_{ctf}> 0.05$).}
\label{fig:example2}
\vskip -0.2in
\end{wrapfigure}

A common concern is that models trained to make {automated} decisions often reveal problematic biases \citep[~e.g.]{angwin2016machine,kodiyan19}. The causal approach to address this is typically to constrain a classifier to obey some causally-sensitive \textit{fairness measure}, $\mu$ \citep[~Def. 3.3]{plecko:bareinboim24}. Some measures in the literature involve $\mathcal{L}_3$-quantities, and thus face the familiar issue of nonidentifiability \citep{kusneretal17,imaietal_2023}. Other approaches acknowledge this limitation and try to construct interventional fairness measures that solely use $\mathcal{L}_2$-quantities \citep{salimietal_19}. We present next an example where relying only on $\mathcal{L}_2$-data can misleadingly approve a classifier as fair, but where a realizable $\mathcal{L}_3$ fairness measure actually ensures fairness. This scenario is inspired by a classic experiment in labor economics \citep{bertrand:03}.

\paragraph{Example 2 (Causal fairness).}\hypertarget{example2}{ } A college is developing an automated system to screen candidates in the first round of college applications, receiving as input a standardized CV per candidate. The system contains two models: model 1 outputs $Y$ and model 2 outputs $Z$, which are binary decisions of whether the applicant cleared the first review stage for admission and for financial scholarship, respectively. The two models are respectively trained using data from previous years where an admissions team and a separate scholarship team reviewed applications manually. The college wants to ensure fairness w.r.t $X$, a candidate's race (a binary variable, for simplicity). In particular, they want to ensure equitable financial access to education for all qualified candidates: a candidate of race $X=1$ who cleared the admissions screening $(Y=1)$ but was rejected for financial aid $(Z=0)$ should still receive $Z=0$ had they been of race $X=0$. The causal diagram is in Fig. \ref{fig:example2}(a), where the models' decisions $Y, Z$ might reflect the unconscious race bias of the two committees in previous years (including possibly shared biases, represented by the latent confounder).

The $\mathcal{L}_3$ fairness measure they ought to minimize is thus
\begin{align}
    \mu_{ctf} = |P(Y_{x_1} = 1, Z_{x_1} = 0) - P(Y_{x_1} = 1, Z_{x_0} = 0)|
\end{align}
But the second term $P(y_x, z_{x'}')$ is nonidentifiable from the causal diagram in \ref{fig:example2}(a). So the college instead uses the following $\mathcal{L}_2$ measures, as an approximation for the fairness condition:

\begin{align}
    \mu_{int1} = &|P(Y = 1; \doo{x_1}).P(Z = 0; \doo{x_1}) - P(Y = 1; \doo{x_1}).P(Z = 0; \doo{x_0})|\\
    \mu_{int2} = &|P(Y = 1, Z=0; \doo{x_1}) - P(Y = 1, Z=0; \doo{x_0})|
\end{align}
They train the models, adding $\mu_{int1} + \mu_{int2}$ as a penalty in the objective. $\mu_{int1}, \mu_{int2}$ are estimated using a holdout set of fake CVs, with the intervention $\doo{x}$ being enacted by randomly choosing an applicant name from an equivalence class which stereotypically indicates one unique race group $X=x$, e.g. names like Lakisha and Jamal for Blacks, or last names like Nguyen or Xi for Asians (cf. \citet{bertrand:03}). Since the holdout set's CV body is independent of $X$, any effect of $X$ on $Y$ and $Z$ is solely via the perception of race from the candidate name. We show in \ref{fig:example2}(c) simulations of such an optimization. In blue is the distribution of the true score $\mu_{ctf}$, when the models are trained using $\mu_{int1}, \mu_{int2}$. Out of 1000 simulated classifiers, we see $\mu_{ctf}> 5\%$ for nearly half the $\mathcal{L}_2$ simulations, indicating statistically significant discrimination roughly 50\% the time.

However, the distribution $P(Y_x, Z_{x'})$ is indeed realizable (Def. \ref{def:realizability}) via the interventions $\textsc{ctf-Rand}(X \rightarrow Y), \textsc{ctf-Rand}(X \rightarrow Z)$. The data science team notices that they can \textit{separately and simultaneously} randomize the candidate name as an input to the respective models, and enact these interventions, as shown in \ref{fig:example2}(b). Thus, they are able to directly use the counterfactual measure $\mu_{ctf}$ as a fairness constraint in training. Results from 1000 simulations show that the classifiers trained directly using $\mu_{ctf}$ (shown in orange) nearly always meet the fairness requirement.

Details of the implementation are in App. \ref{app:college_admissions}. Note: as in the original experiment, this example requires the structural assumption of race being revealed at the screening stage only by candidate name, which may be more defensible in highly standardized and controlled application processes. $\hfill \blacksquare$

\subsection{Causal RL - counterfactual policies for optimal decision-making}
\label{sec:bandit_example}

\begin{wraptable}{r}{0.5\textwidth}
\vspace{-0.15in}
\caption{Performance of different strategies in {Example 3}.}
        \begin{tabular}{p{2.4cm} p{2.2cm} p{1.2cm}}
            \toprule
            \textbf{Strategy}     & \textbf{Objective} & \textbf{Avg. Reward }\\
            \midrule
            \midrule
            Behavioral policy ($\mathcal{L}_1$)  & - &  0.65 \\
            \midrule
            Naive randomization ($\mathcal{L}_2$) & $\mathbb{E}[Y; \doo{x}]$ & 0.7 \\
            \midrule
            ETT baseline strategy ($\mathcal{L}_3$) & $\mathbb{E}[Y_x \mid x']$ & 0.75 \\
            \midrule
            Optimal $\mathcal{L}_3$ strategy (this work) & $\mathbb{E}[Y_x \mid x', d_{x''}]$ &  0.80 \\
            \bottomrule
          \end{tabular}
\label{table:alice_table}
\vspace{-0.in}
\end{wraptable}

Consider a multi-arm bandit problem in which $X$ represents the choice of bandit arm and $Y$ the outcome. The default online learning approach is for the agent to adopt an algorithm like EXP3, UCB or Thompson Sampling to converge to some arm $x^\star := \arg \max_x \mathbb{E}[Y; \doo{x}]$ \citep{Lattimore_Szepesvári_2020, sutton1998reinforcement}. Even in methods that explicitly incorporate causal knowledge, the common approach is to use a combination of offline ($\mathcal{L}_1$) and online ($\mathcal{L}_2$) data to converge more efficiently to the $\mathcal{L}_2$ optimization target $\arg \max_x \mathbb{E}[Y; \doo{x}]$  \citep[~e.g.]{zhang:bar17}. See \citet{crl_r65} for an overview of the emerging field of causal RL.

It was already shown in \citep{bareinboim:etal15,forney:etal17} that it is possible to perform better by deploying a counterfactual strategy based on sampling each unit's natural choice $X=x'$ and randomizing actual choice in the \textit{same} round, thus seeking to converge to $\arg \max_x \mathbb{E}[Y_x\mid x'], \forall x'$, as we discussed in Sec. \ref{sec:data_collection_procedures}. We call this the ETT baseline strategy, as it relies on drawing samples from the $\mathcal{L}_3$ ETT distribution, $P(Y_x, X)$, mentioned in Sec. \ref{sec:intro}.


We improve on this baseline by showing how an agent can leverage the realizability (Def. \ref{def:realizability}) of more nuanced counterfactuals like $P(Y_x, X, D_{x''})$ to construct superior counterfactual strategies. The following scenario involves an agent faced with adversarial latent confounding.

\begin{wrapfigure}{r}{0.45\textwidth}
    \vspace{-0.15in}
    \centering
    \includegraphics[width=0.4\textwidth]{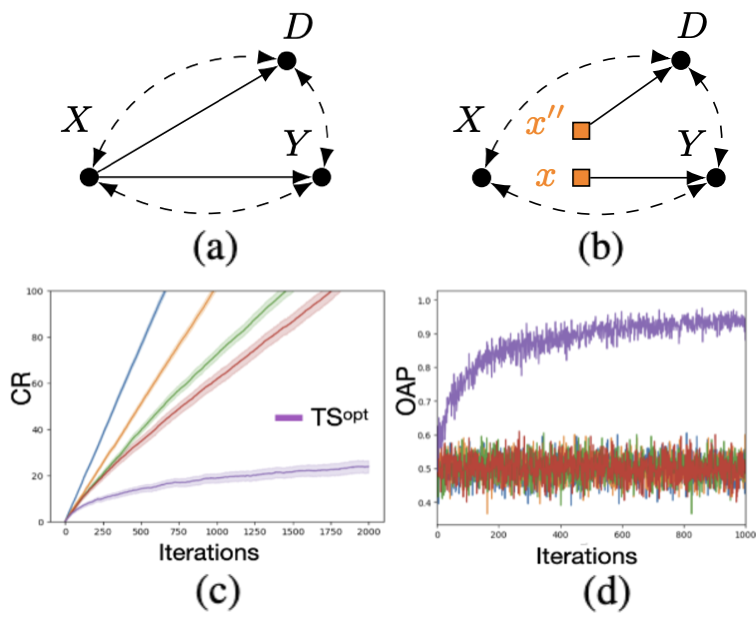}
    \vspace{-0.1in}
    \caption{(a) Causal diagram for Example 3; (b) $P(Y_x, X, D_{x''})$ is realizable using the interventions $\textsc{ctf-Rand}(X \rightarrow Y)$ and $\textsc{ctf-Rand}(X \rightarrow D)$; (c) Cumulative Regret (CR) for $\mathcal{L}_1$ strategy (blue) and Thompson Sampling algorithms implementing naive $\mathcal{L}_2$ (yellow, green), ETT baseline (red), and optimal $\mathcal{L}_3$ strategy (purple); (d) Optimal Arm Probability (OAP) for all algorithms.}
    \vspace{-0.in}
    \label{fig:example3}
\end{wrapfigure}

\vspace{-0.in}
\paragraph{Example 3 (Counterfactual bandit policies).}\hypertarget{example3}{ }  Consider a user of a social media platform which uses surveillance and predictions to increase user engagement through addictive notifications and recommendations \citep{zuboff}. The user chooses every evening whether to use the platform via desktop ($X=0$) or mobile ($X=1$). $Y$ is a binary indicator of whether she stays within her self-determined social media usage limit per day. She also notices that she receives ads when she logs in each evening as $D$ (0: streaming service, 1: food delivery ads). The usage type $X$ affects $D,Y$, as shown in Fig. \ref{fig:example3}(a). 

On average, the user experiences $\mathbb{E}[Y]=0.65$ from the observational ($\mathcal{L}_1$) policy of following her natural inclination each day. She suspects that the company could be tracking and exploiting her latent preferences, so she decides to randomize her daily choice and pick the best "arm". Sure enough, this naive $\mathcal{L}_2$ strategy breaks the adversarial confounding, and incurs a better avg. performance of $\mathbb{E}[Y ; \doo{x}]=0.7, \forall x$. 
She then decides to test the ETT-based strategy ($\mathcal{L}_3$) described earlier, by recording what she naturally feels like doing each day ($X=x'$), and subsequently randomizing her actual choice on the same day to optimize $\mathbb{E}[Y_x \mid x']$, getting an avg. performance of 0.75. However, at this point, she notices that she can do even better. The $\mathcal{L}_3$-distribution $P(Y_x, X, D_{x''})$ is realizable (Def. \ref{def:realizability}), since she can perform \textit{another} counterfactual randomization, by sampling her natural choice $(X=x')$, randomly logging in to just see what ads she gets $(D_{x''}=d)$, and again randomizing how she actually uses the platform that day to get $Y_x$. This strategy seeks an optimal $x^\star = \arg \max_x \mathbb{E}[Y_x \mid x', d_{x''}]$, which performs best as shown in Table \ref{table:alice_table}. Details of the SCM, latent confounders, and the optimal $\mathcal{L}_3$-strategy are in App. \ref{app:bigtechsurveillance}.

Simulations in the online setting corroborate this finding. Fig. \ref{fig:example3}(c,d) shows the cumulative regret (CR) and optimal arm probability (OAP) over 2000 iterations averaged over 200 epochs (CI=95\%). We adapt Thompson Sampling to implement the strategies in Table \ref{table:alice_table}. Details of implementation are in App. \ref{app:example3_simulations}. The optimal $\mathcal{L}_3$ strategy (purple) performs best, improving on the performance of the baseline ETT-based strategy (red). Naive randomizations, the standard $\mathcal{L}_2$ bandit strategy, are shown in yellow and green. All other algorithms fail to improve in OAP after 2000 iterations.  $\hfill \blacksquare$

We make two remarks. First, the optimal counterfactual strategy is not simply a contextual Thompson Sampling, where $X, D_{x''}$ are used "merely" as extra context variables per round; indeed, treating this merely as a contextual bandit problem is one of the naive $\mathcal{L}_2$-strategies that we test (green plot in Fig. \ref{fig:example3}(c-d)), which ignores the counterfactual relationship between these variables and incurs dramatically higher regret, as we discuss in App. \ref{app:example3_simulations}.

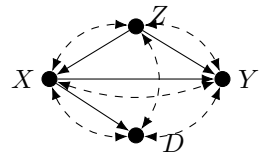
\begin{wrapfigure}{r}{0.28\textwidth}
    \vspace{-0.25in}
    \begin{center}
    \begin{tikzpicture}

        \node (X) at (-1,0) {$X$};
        \node[circle,draw=black,fill=black, inner sep=2] (Xnode) at (-0.65,0) {\ };
        \node (Z) at (0.8,0.85) {$Z$};
        \node[circle,draw=black,fill=black, inner sep=2] (Znode) at (0.5,0.7) {\ };
        \node (D) at (1,-0.85) {$D$};
        \node[circle,draw=black,fill=black, inner sep=2] (Dnode) at (0.5,-0.75) {\ };
        \node (Y) at (2,0) {$Y$};
        \node[circle,draw=black,fill=black, inner sep=2] (Ynode) at (2-0.35,0) {\ };
        \path [-Latex] (Znode) edge (Xnode);
        \path [-Latex] (Znode) edge (Ynode);
        \path [-Latex] (Xnode) edge (Ynode);
        \path [-Latex] (Xnode) edge (Dnode);
        \path [Latex-Latex,dashed] (Xnode) edge[bend left=40] (Znode);
        \path [Latex-Latex,dashed] (Xnode) edge[bend right=20] (Ynode);
        \path [Latex-Latex,dashed] (Xnode) edge[bend right=40] (Dnode);
        \path [Latex-Latex,dashed] (Znode) edge[bend left=40] (Ynode);
        \path [Latex-Latex,dashed] (Znode) edge[bend left=40] (Dnode);
        \path [Latex-Latex,dashed] (Ynode) edge[bend left=40] (Dnode);

    \end{tikzpicture}
    \end{center}
    \vskip -0.15in
    \caption{MAB template.}
    \label{fig:mab_template}
    \vskip -0.15in
    \end{wrapfigure}

Second, an interesting follow-up is whether we can guarantee that our strategy based on maximizing $\mathbb{E}[Y_x \mid x', d_{x''}]$ is optimal in this problem. Perhaps there are more refined $\mathcal{L}_3$-distributions like $P(Y_x, X, D_x, D_{x''})$ etc. that could yield better algorithms? It turns that this is indeed optimal, since most other $\mathcal{L}_3$-distributions are not realizable (Def. \ref{def:realizability}). As a bonus, we prove this claim next, for all bandit problems that fit a specific causal template defined below. Thereby, we avoid having to conduct an intractable search over the space of all possible $\mathcal{L}_3$-strategies, trying to assess their realizability.

To focus the discussion, we define a generic \textit{MAB template} (Fig. \ref{fig:mab_template}) that is generally representative of a broad class of bandit problems in the literature (the discussion can also be extended to other settings such as sequential or Markov decision processes in future work). $X$ is the decision variable\xst{ in the MAB problem}, $Z$ is a context variable, $Y$ is the reward, and $D$ is a descendant of $X$ confounded with $Y$.

\begin{definition}[Decision strategy]
    Given a decision problem following the MAB template (Fig. \ref{fig:mab_template}), a \textit{decision strategy} $\pi$ is a mapping from a set of variables $\*W_\star$ (possibly counterfactual) to a set of actions $\mathcal{A}$ involving decision variable $X$. The expected reward of following this strategy is notated $\mu_\pi := \mathbb{E}[Y_{\mathcal{A}} \mid \*W_\star]$, where $Y_{\mathcal{A}}$ is the potential response of $Y$ under the actions $\mathcal{A}$.\footnote{We use \textit{strategy} interchangeably with \textit{policy} when the context is clear. However, it should be noted that a \textit{ policy} is usually defined w.r.t. a certain policy space, mapping from a fixed domain to actions on $X$. Here, we consider different domains to map from.} $\hfill \blacksquare$
\end{definition}

\textit{Example.} The $\mathcal{L}_1$ strategy of simply observing the natural behavior of some behavioral agent is $\pi^{\text{obs}}: \{\} \mapsto \{\}$, which incurs the observational reward of $\mu_{\pi^{\text{obs}}} = \mathbb{E}[Y]$.\xst{ This is typically the performance of attained by a "behavioral" agent.} 

\textit{Example.} As discussed in Sec. \ref{sec:bandit_example}, the typical approach in the RL literature is the $\mathcal{L}_2$ strategy $\pi^{\text{int}}: \{z\} \mapsto \{\textsc{Write}(X:x^\star)\}$, where $x^\star := \arg \max_{x} \text{ } \mathbb{E}[Y_{x} \mid z]$. In words, this strategy involves observing context $Z=z$ for a each round, and then performing the intervention $\doo{x}$ that maximizes the $\mathcal{L}_2$ quantity $\mathbb{E}[Y_x \mid z]$, also known as the \textit{conditional average treatment effect}, or CATE. 

\textit{Example. } As discussed in Secs. \ref{sec:data_collection_procedures}, \ref{sec:bandit_example}, a valid counterfactual strategy would be $\pi^{\text{ett}}: \{x', z\} \mapsto \{\textsc{ctf-Write}(x^\star) \rightarrow Y\}$, where $x^\star:= \arg \max_x \mathbb{E}[Y_x \mid x', z]$.\footnote{$\textsc{ctf-Write}$ is simply the deterministic equivalent of $\textsc{ctf-Rand}$ (Def. \ref{def:ctf-rand}).} In words, this strategy involves observing context $Z=z$ and the unit's natural inclination $X=x'$ for each each round, and then performing the intervention $\doo{x}$ that maximizes the $\mathcal{L}_3$ quantity $\mathbb{E}[Y_x \mid x', z]$, related to the \textit{effect of the treatment on the treated}, or ETT.


In Example \hyperlink{example3}{3}, we introduced a superior counterfactual strategy
\begin{align}
    \pi^{\text{opt}}: \{X,Z,D_{x''}\} \mapsto \{\textsc{ctf-Write}(x \rightarrow Y),\textsc{ctf-Write}(x'' \rightarrow D)\},\label{eq:pi_opt}
\end{align}
where  $x,x'' := \arg \max_{x,x''} \mathbb{E}[Y_x \mid Z,X,D_{x''}]$.

With minor abuse of notation, this is the strategy that (1) observes $Z=z, X=x'$ for a round; (2) maps from $\{z,x'\} \mapsto x''$, to perform the counterfactual intervention $\textsc{ctf-Write}(x'' \rightarrow D)$ to observe $D_{x''}=d$; and (3) maps from $\{z,x',d_{x''}\} \mapsto x$, to perform the counterfactual intervention $\textsc{ctf-Write}(x \rightarrow Y)$ that maximizes $\mathbb{E}[Y_x\mid z, x', d_{x''}]$. 

For each mapping in (2), (3) yields a local optimum, in expectation over $X,Z,D_{x''}$. Optimizing over all choices of $x''$ in (2) yields a global optimum. Translating this to practice, we provide a general algorithm (Algorithm \ref{alg:l3_opt}) that adapts any standard MAB solver to implement the optimal $\mathcal{L}_3$-strategy $\pi^{\text{opt}}$. We provide examples using Thompson Sampling in the Appendix \ref{app:example3_simulations} (Algorithms \ref{alg:ts_opt},\ref{alg:ts_ett}).

\begin{wrapfigure}{r}{0.4\textwidth}
  \centering
  \vspace{-0.in}
  \includegraphics[width=0.375\textwidth]{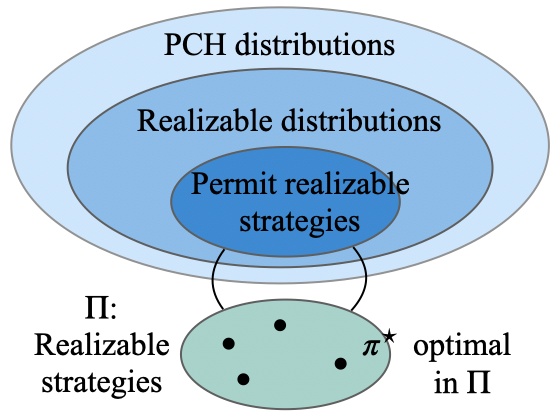}
  \caption{$\pi^{opt}$ is provably optimal in the space of realizable decision strategies an RL agent can adopt in a bandit problem.}
  \vspace{-0.15in}
  \label{fig:roadmap}
\end{wrapfigure}

The natural question is whether we can keep going higher up in Layer 3 of the PCH. Could we construct higher order strategies that map from $\{X, Z, D_{x''}, D_{x'''}\}$ by drawing samples from more refined counterfactuals? Sadly no, because a distribution like $P(Y_x, X, D_{x''}, D_{x'''})$ is not realizable (Def. \ref{def:realizability}), and the machinery we developed in Sec. \ref{sec:realizability_main_sec} gives us the tools to reason about this.

\begin{theorem}[Optimality]\label{thm:optimality}
    Given a decision problem following the {MAB template} (Fig. \ref{fig:mab_template}), $\pi^{\text{opt}}$ is an optimal realizable strategy. I.e., $\mu_{\pi^{\text{opt}}} \geq \mu_\pi, \forall \pi \in \Pi,$ the space of realizable strategies. $\hfill \blacksquare$
\end{theorem}

The significance of this result is that it averts the need to apply Thm. \ref{thm:completeness} and Cor. \ref{cor:maximal_amwn} to search intractably over the space of all possible $\mathcal{L}_3$-distributions for which ones are realizable, as illustrated in Fig. \ref{fig:roadmap}. Of course, $\pi^{\text{opt}}$ need not be \textit{uniquely} optimal.


\begin{corollary}[$\mathcal{L}_3$-dominance]
\label{cor:l3_domination}
    Given an MAB decision problem with causal diagram described by the {MAB template} (Fig. \ref{fig:mab_template}), the optimal $\mathcal{L}_3$-strategy $\pi^{\text{opt}}$ dominates the $\mathcal{L}_1$-strategy $\pi^{\text{obs}}$ and the optimal $\mathcal{L}_2$-strategy $\pi^{\text{int}}$. I.e., $\mu_{\pi^{\text{opt}}} \geq \mu_{\pi^{\text{obs}}}$ and $\mu_{\pi^{\text{opt}}} \geq \mu_{\pi^{\text{int}}}$. $\hfill \blacksquare$
\end{corollary}

\xst{This is an important result. }For decades, the Fisherian RCT methodology \xst{(known as A/B testing in industry) }used to enact $\pi^{\text{int}}$ was deemed to be the "gold standard" for decision-making. We show that the $\mathcal{L}_3$-strategy $\pi^{\text{opt}}$ is at least as good (and often better) than $\mathcal{L}_2$-strategies. This means that if MAB solvers UCB, EXP3 etc. were deployed to enact an $\mathcal{L}_2$-strategy in an environment where $\pi^{\text{opt}}$ is better, the agent would incur \xst{infinite regret in the limit }\xadd{linear cumulative regret}, since the learning approach comes no closer to discovering the optimal strategy as the number of trials increases.



\begin{algorithm}[H]
        \caption{MAB-OPT} \label{alg:l3_opt}
        \begin{algorithmic}[1]
          \STATE {\bfseries Input:} MAB problem following Fig. \ref{fig:mab_template}; MAB solver (e.g. UCB, EXP3, TS); No. of rounds $T$; Obs. data $P(\*v)$

          \smallskip
          \FOR{each $z,x'$}
            \STATE Initialize $D$-arms $x''$
          \ENDFOR

          \smallskip
          \FOR{each $z,x',x'',d$}
            \STATE Initialize $Y$-arms $x$
            \STATE If $x = x' = x''$, hot-start using $P(\*v)$
          \ENDFOR

          \smallskip
          \FOR{$t \in [T]$}
            \STATE Observe $x', z$
            \STATE Draw D-arm $x''$ using MAB solver
            \STATE Perform $\textsc{ctf-Write}(x'' \rightarrow D)$ and get $D_{x''} = d$
            \STATE Draw Y-arm $x$ using MAB solver
            \STATE Perform $\textsc{ctf-Write}(x \rightarrow Y)$ and get $Y_{x} = y$
            \STATE Update D-arms and Y-arms according to MAB solver rules using $y$
          \ENDFOR
        \end{algorithmic}
      \end{algorithm}

\section{Conclusion}

In this paper, we tackle the open question of which counterfactual distributions are directly accessible by experimental methods - what we define as the \textit{realizability} of a distribution. Countering prevalent belief, we provide a complete algorithm and a graphical criterion for when a counterfactual can indeed be physically sampled from (Fig. \ref{fig:pch_realizability}). We demonstrate the practical relevance of this new framework with examples from causal fairness and causal RL, highlighting that ignoring this possibility could lead to poor outcomes. We believe that switching from an interventional to a counterfactual mindset could help researchers spot opportunities for \textit{counterfactual randomization} that permit exciting new types of experiments, and improved, more personalized decisions. 

\subsection*{Acknowledgements}
This research is supported in part by the NSF, ONR, AFOSR, DoE, Amazon, JP Morgan, and The
Alfred P. Sloan Foundation. We thank Juan D. Correa and the anonymous reviewers for their thoughtful comments.

\newpage
\bibliography{references}
\bibliographystyle{iclr2025_conference}


\newpage
\appendix
\section*{Appendices}

Appendix \ref{app:discussion}: Discussion

Appendix \ref{app:scm}: Graphical terminology

Appendix \ref{app:realize_algo}: Details on the \textbf{CTF-REALIZE} algorithm

Appendix \ref{app:assumptions_and_proofs}: Assumptions and realizability proofs

Appendix \ref{app:l3_conditions}: Details on counterfactual randomization

Appendix \ref{app:examples}: Details on examples

Appendix \ref{app:optimality_proofs}: Optimality proofs




\bigskip
\section{Discussion}
\label{app:discussion}

In this section, we discuss some important implications, future directions, and limitations of our work.

\textbf{Identification and bounding. } Much work has been done in the area of $\mathcal{L}_3$ identification and estimation {\citep{shpitser:08,correaetal:21,geneletti_dawid_ett}}. A natural extension to our work is to investigate the relationship between realizability and identification: which \textit{additional} $\mathcal{L}_3$-quantities now become identifiable if the environment permits even some counterfactual randomization? This warrants an update to existing identification algorithms to allow (some) $\mathcal{L}_3$-data as input. {Another fascinating research question involves "partial identification", where an input query is tightly bounded within a range that can be computed from available data \citep{zhang22ab}: how would the new $\mathcal{L}_3$-data further tighten the bounds for nonidentifiable $\mathcal{L}_3$-quantities?} 

\textbf{Experiment design. } One of the goals of this paper is to instigate new experiment design ideas that leverage {ctf-randomization} (Def. \ref{def:ctf-rand}) and go beyond the standard RCT methodology, as in Examples \hyperlink{example1}{1}-\hyperlink{example2}{2}. For instance, the increasingly automated HR pipeline in companies suggests opportunities for targeted interventions to randomize demographic details in virtual interviews, in standardized aptitude tests, or in performance-evaluation systems for remote workers, to track fairness metrics.

\textbf{Causal reinforcement learning (CRL). } While counterfactual strategies have been studied in CRL, 
the literature currently focuses on ETT-related strategies based on optimizing $\mathbb{E}[Y_x \mid x']$ \citep{bareinboim:etal15,forney:etal17,zhang22a}\citep[~\S 5.1]{richardson:robins13}. 
We presented an important extension by formalizing ctf-randomization (Def. \ref{def:ctf-rand}) via {counterfactual mediators} (Def. \ref{def:ctf_mediator_informal}), subsuming the previous approach. An ETT-based approach only allows one randomization of a variable $X$, affecting all downstream mechanisms. Our approach recognizes the possibility of isolating specific causal pathways and randomizing $X$ multiple times per unit, demonstrably surpassing the ETT baseline in Example \hyperlink{example3}{3}. We proved in Thm. \ref{thm:optimality} and Cor. \ref{cor:l3_domination} an optimality guarantee for our proposed strategy in bandit problems following a causal template. Generalizing this to sequential decision-making settings with arbitrary graphs is an important, non-trivial extension. 


\textbf{Limitations. } The first obvious limitation of our framework is that it requires causal knowledge in the form of a graph (or equivalent). This is a standard assumption, needed to make progress in several areas of causal machine learning. Subsequent work could accommodate partial knowledge or model misspecification. Second, it may not always be feasible to perform counterfactual randomization (Def. \ref{def:ctf-rand}) in a given setting. This is why Algo. \ref{alg:realize} and Thm. \ref{thm:completeness} are general and do not assume this capability a priori. 
But where it is possible, even in principle, our work pinpoints opportunities for novel experiment design, as discussed above.
\newpage
\section{Graphical terminology}
\label{app:scm}

Structural Causal Models (SCM) and causal diagrams are described in the preliminaries in Sec. \ref{sec:intro}. See \citep{Bareinboim2022OnPH} for full treatment. We use the following graphical kinship nomenclature w.r.t causal diagram $\mathcal{G}$:
\begin{itemize}
    \item Parent(s) of $V$, denoted $\*{Pa}_V$: the set of variables $\{V'\}$ s.t. there is a direct edge $V' \rightarrow V$ in $\mathcal{G}$. $\*{Pa}_V$ does not include $V$.
    \item Children of $V$, denoted ${Ch}(V)$: the set of variables $\{V'\}$ s.t. there is a direct edge $V \rightarrow V'$ in $\mathcal{G}$. $Ch(V)$ does not include $V$.
    \item Ancestors of $V$, denoted $An(V)$: the set of variables $\{V'\}$ s.t. there is a path (possibly length 0) from $V'$ to $V$ consisting only of edges pointing toward $V$, $V' \rightarrow ... \rightarrow V$. {$An(V)$ is defined to include $V$}.
    \item Descendants of $V$, denoted $Desc(V)$: the set of variables $\{V'\}$ s.t. there is a path (possibly length 0) from $V$ to $V'$ consisting only of edges pointing toward $V'$, $V \rightarrow ... \rightarrow V'$. {$Desc(V)$ is defined to include $V$}.
    \item Non-descendants of $V$, denoted $NDesc(V)$: the set $\*V \setminus Desc(V)$. $NDesc(V)$ does not include $V$.
\end{itemize}

Given a graph $\mathcal{G}$, $\mathcal{G}_{\overline{\*X}\underline{\*W}}$ is the result of removing edges coming
into variables in $\*X$, and edges coming out of $\*W$. 




\newpage
\section{Details on the \textbf{CTF-REALIZE} algorithm}
\label{app:realize_algo}


\subsection{Sub-routine of \textbf{CTF-REALIZE} algorithm (Algo. \ref{alg:realize})}
\label{app:subroutine}

\begin{algorithm}
\begin{multicols}{2}
\caption{COMPATIBLE (sub-routine)}
   \label{alg:subroutine}
\begin{algorithmic}[1]
   \STATE {\bfseries Input:} $V \in \*V$ of $\mathcal{G}$; $W_{\*t} \in \*W_\star$ of $Q$

   \smallskip
   \FOR{each $C \in Ch(V)$}

   \IF{$C \in An(W)$}

   \smallskip
   \IF{$V \in \*T$} 
   \STATE Let $v := $ value of $V$ in subscript $\*t$
   \STATE Find smallest $\*C \ni C$ s.t. $\textsc{ctf-Rand}(V \rightarrow \*C) \in \mathbb{A}$
   \IF{$\{\textsc{ctf-Rand}(V \rightarrow \*C): .\} \in \text{INT}_V$ and its label is not "$v$"}
        \STATE Return FAIL
        \ELSE
        \STATE Add $\{\textsc{ctf-Rand}(V \rightarrow \*C): v\}$ to $\text{INT}_V$, with the label "$v$"
    \ENDIF
   \IF{no such $\*C \ni C$ s.t. $\textsc{ctf-Rand}(V \rightarrow \*C) \in \mathbb{A}$}
        \IF{$\{\textsc{Rand}(V): .\} \in \text{INT}_V$ and its label is not "$v$"}
            \STATE Return FAIL \
            \ELSIF{$\textsc{Rand}(V) \not \in \mathbb{A}$}
                \STATE Return FAIL
            \ELSE
                \STATE Add $\{\textsc{Rand}(V): v\}$ to $\text{INT}_V$, with the label "$v$"
        \ENDIF

    \ENDIF
   \ENDIF

   \smallskip
   \IF{$V  \not\in \*T$} 
   
    \FOR{each $\*C \ni C$ s.t. $\textsc{ctf-Rand}(V \rightarrow \*C) \in \mathbb{A}$}
       \IF{$\{\textsc{ctf-Rand}(V \rightarrow \*C): .\} \in \text{INT}_V$ and its label is not "Natural"}
            \STATE Return FAIL
            \ELSE
            \STATE Add $\{\textsc{ctf-Rand}(V \rightarrow \*C): \text{Natural}\}$ to $\text{INT}_V$, with the label "Natural"
        \ENDIF
    \ENDFOR
    \IF{$\{\textsc{Rand}(V): .\} \in \text{INT}_V$ and its label is not "Natural"}
        \STATE Return FAIL \
        \ELSIF{$\textsc{Rand}(V) \in \mathbb{A}$}
            \STATE Add $\{\textsc{Rand}(V): \text{Natural}\}$ to $\text{INT}_V$, with the label "Natural"
    \ENDIF
    
   \ENDIF
   
   \smallskip
   \ENDIF
   
   \ENDFOR
\end{algorithmic}
\end{multicols}
\end{algorithm}


\subsection{Walk-through of the \textbf{CTF-REALIZE} algorithm (Algo. \ref{alg:realize})}
\label{app:algo_walkthrough}

\textbf{General strategy}. For each variable $V$ in the input graph, we check what are the necessary and sufficient interventions (or lack of interventions) we need to perform w.r.t each term $W_{\mathbf{t}}$ in the input query $\mathbf{W}_\star$. This is what the inner loops and subroutine \textbf{COMPATIBLE} are doing - accumulating correct and complete conditions in topological order. If there is no conflict across these conditions collectively, and if the feasible action set contains the necessary actions, the query is realizable. Otherwise not. {We use induction to prove that it is enough to check for conflicting conditions with regard to prior loops, in topological order. Proof of completeness results in Appendix \ref{app:assumptions_and_proofs}.}

Walk-through for Algo. \ref{alg:realize} \textbf{CTF-REALIZE}:
\begin{itemize}
    \item [i.] \textbf{Lines 5-7}: {we go over each node $V$ in the input graph, in topological order, and maintain a tracker of interventions (and lack of interventions) needed w.r.t $V$ to realize $\*W_\star$}
    ; we also check whether $V$ needs to be added to the final output vector.
    \item [ii.] \textbf{Line 10}: for each $W_\*t$ in the input query $\*W_\star$, we check if there is any conflict in necessary and sufficient conditions w.r.t $V$ for realizing $W_\*t$, by calling \textbf{COMPATIBLE}($V, W_\*t$). This only needs to be done if $V \in An(W)$; otherwise $V$ has no effect on $W_\*t$.
    \item [iii.] \textbf{Line 13}: if $V = W$, the measured value of $V$ needs to be added to the output vector $\*w$.
    \item [iv.] \textbf{Lines 17-18}: we perform all the interventions (if any) that are needed for $V$. {Step [ii] has already checked whether these actions can be performed, and if there are any conflicts.} 
    \begin{itemize}
        \item \textbf{Note on rejection sampling}: since we framed our actions as randomizations, in order to enact an intervention like $\doo{x}$, we draw a random value and reject if the draw is not $x$. This is for clarity of presentation, aligned with the rest of the paper. We could easily introduce deterministic actions, or add some concentration guarantees of drawing $x$ within finite samples etc. but this is well-understood and would be a distraction.
    \end{itemize}
    \item  [v.] \textbf{Lines 21-25}: if $V$ itself is part of the output, the set of necessary actions cannot involve a Fisherian $\textsc{Rand}$ of $V$ (because unlike $\textsc{ctf-Rand}$, Fisherian $\textsc{Rand}$ overrides the mechanism $f_V$ generating $V$).
    \item [vi.] \textbf{Line 29}: {if the above steps have been completed w.r.t $V$ for each $W_\*t$, there will be no further conflicts arising w.r.t $V$ and all nodes topologically prior to $V$}, regardless of conditions needed w.r.t subsequent nodes (by an induction argument). If this is can be completed for all $V$, then the query is realizable and we output the vector $\*w$.
\end{itemize}

Walk-through for sub-routine Algo. \ref{alg:subroutine} \textbf{COMPATIBLE}$(V,W_\*t)$ called in Step [ii]:
\begin{itemize}
    \item [ii.a.] \textbf{Lines 2-3}: the necessary and sufficient conditions w.r.t $V$ for $W_\*t$ involve how the children $Ch(V)$ receive the value of $V$ as an input. We only care about the children that belong in $An(W)$ for this sub-routine call; if a child is not in $An(W)$ it wouldn't affect $W_\*t$.
    \item [ii.b.] \textbf{Lines 4-19}: if $V \in \*T$, this means the potential response $W_\*t$ involves an intervention on $V$. We find the minimal interventions needed to achieve this ($\textsc{ctf-Rand}$ for the smallest subset of children possible, and failing this, a Fisherian $\textsc{Rand}$); and we update our tracker of necessary actions for $V$.
    {
    \begin{itemize}
        \item[ii.b.1.] The necessary and sufficient conditions to measure  $W_\*t$ are that $\*T$ should be fixed as $\*t$ (by intervention) as an input to all children $C \in Ch(\*T) \cap An(W)$; and that all other ancestors of $W$ are received naturally by their respective children. These two conditions ensure that $W$ is evaluated in the $\doo{\*t}$ regime, as defined in the SCM.
        \item[ii.b.2.] In particular, for $V \in \*T$ in this sub-routine call, this means that we need to fix $v \in \*t$ (\textbf{Line 5}) as input to each relevant child $C \in Ch(V) \cap An(W)$. So for each such child $C$, we find the smallest subset $\*C' \subseteq Ch(V)$ s.t. $C \in \*C'$ and $\textsc{ctf-Rand}(V \rightarrow \*C')$ is in the input action set $\mathbb{A}$ (\textbf{Line 6}), and we add this to the required actions, along with the tagged value "$v$" that needs to be fixed for $V$ by this action (\textbf{Line 10}). If $\textsc{ctf-Rand}$ is not available, we add Fisherian $\textsc{Rand}(V)$ to the required action list (\textbf{Line 18}). We call this chosen action the "minimal action".
        \item[ii.b.3.] If no such randomization action is available, return FAIL (\textbf{Line 16}). If the minimal action is already being used in a previous loop to enforce a different value $v' \neq v$ that is not compatible with $\*t$, return FAIL (\textbf{Lines 7-8, 13-14}). Or if the minimal action has already been tagged with the value "Natural" in a previous loop, return FAIL - this means a previous loop already recorded that this action must \textit{necessarily} \textit{not} be performed (\textbf{Lines 7-8, 13-14}). Such a conflict means that $W_\*t$ cannot be realised.
    \end{itemize}}
    \item [ii.c.] \textbf{Lines 22-34}: if $V \not \in \*T$, this means the potential response $W_\*t$ requires that $V$ be received without intervention (i.e. "naturally") by the relevant child nodes; we also add this necessary condition to our tracker.
    {
    \begin{itemize}
        \item[ii.c.1.] Again, the necessary and sufficient conditions to measure $W_\*t$ are that each ancestor $A \in An(W)_{\mathcal{G}_{\overline{\*T}}}$, with $A \not \in \*T\cup \{W\}$, needs to be received "naturally" (i.e., without intervention) by its children $C \in Ch(A) \cap An(W)$; and that $\*T$ needs to be fixed by intervention, as discussed earlier in [ii.b.1]. These two conditions ensure that $W$ is evaluated in the $\doo{\*t}$ regime, as defined in the SCM.  
        \item[ii.c.2.] In particular, for $V \not \in \*T$ in this sub-routine, this means that we should \textit{necessarily not} intervene on the value of $V$ that is received as input by each relevant child $C \in Ch(V)$ which is also an ancestor of $W$.
        \item[ii.c.3.] We track this requirement by adding that for every possible randomization $\textsc{ctf-Rand}(V \rightarrow \*C')$ and $\textsc{Rand}(V)$, where $\*C'$ contains an ancestor of $W$, that this action \textit{cannot} be performed, by using the tag "Natural" (\textbf{Lines 27, 33}).
        \item[ii.c.4.] If any such action identified in [ii.c.3] has already been tagged with some value $v'$, this means that a previous loop has recorded that this action \textit{necessarily} needs to be performed in order to fix the value as $v'$ - we return FAIL (\textbf{Lines 24-25, 30-31}). Such a conflict means that $W_\*t$ cannot be realized.
    \end{itemize}}
\end{itemize}


\subsection{Complexity analysis of \textbf{CTF-REALIZE} algorithm (Algo. \ref{alg:realize})}
\label{app:realize_complexity}

The \textit{time complexity} of Algorithm \ref{alg:realize} is $\mathcal{O}(kn^2)$, where $k$ is the number of terms in the input query $Q$, and $n$ is the number of variables in the input graph $\mathcal{G}$.

$k$ depends on the domain size of the variables. That is, $k \leq n. \prod_V \bigg( |Domain(V)| + 1 \bigg) = \mathcal{O}(m^n)$, where $m$ is the domain size of the variable with the most possible categorical values. 

The \textit{space complexity} is the same, as the algorithm needs to store up to all intermediate steps before terminating.


\subsection{Examples using the \textbf{CTF-REALIZE} algorithm}
\label{app:realize_examples}

\begin{example}(ETT realizability) \label{ex:realize_1}

    Query, $Q = P(Y_x, X)$ 
    
    Graph, $\mathcal{G}:$ Fig. \ref{fig:app_realize_examples_a}

    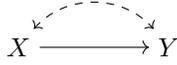
\begin{figure}[h!]
        \centering
        \begin{tikzpicture}
        \node (X) at (0,0.4) {$X$};
        \node (Y) at (2,0.4) {$Y$};
        \path [->] (X) edge (Y);
        \path [<->,dashed] (X) edge [bend left=50] (Y);
    \end{tikzpicture}
    \caption{Graph for Example \ref{ex:realize_1}}
    \label{fig:app_realize_examples_a}
\end{figure}
    
    Suppose action set $\mathbb{A} = \mathbb{A}^\dag(\mathcal{G}) := \{\textsc{ctf-Rand}(X \rightarrow Y)\}$
    
    \textbf{CTF-REALIZE}$(Q,\mathcal{G},  \mathbb{A}^\dag(\mathcal{G}))$ trace:
    \begin{itemize}
        \item Start with $X$ (first in topological order)
        \item For the first term in $\*W_\star$: $Y_x$
        \begin{itemize}
            \item Since $X \in An(Y)$, call Algo. \ref{alg:subroutine} \textbf{COMPATIBLE}($X,Y_x$)
            \begin{itemize}
                \item $Y \in Ch(X)$ and $Y \in An(Y)$
                \item $X \in $ subscript of $Y_x$
                \item $\textsc{ctf-Rand}(X \rightarrow Y) \in \mathbb{A}^\dag(\mathcal{G})$
                \item $\text{INT}_X \gets \{\textsc{ctf-Rand}(X \rightarrow Y): x\}$
            \end{itemize}
        \end{itemize}
        \item For the second term in $\*W_\star$: $X$
        \begin{itemize}
            \item $\text{OUTPUT}_X \gets \{X\}$
        \end{itemize}
        \item Moving to $Y$ (next in topological order)
        \item For the first term in $\*W_\star$: $Y_x$
        \begin{itemize}
            \item $\text{OUTPUT}_Y \gets \{Y_x\}$
        \end{itemize}
        \item Perform interventions in $\text{INT}_X$, followed by $\textsc{READ}$, and assign output vector based on $\text{OUTPUT}_X, \text{OUTPUT}_Y$
        \item \textcolor{ForestGreen}{Return i.i.d sample}
    \end{itemize}

For simplicity, we don't show the steps $\textsc{Select}^{(i)}$ and the rejection sampling involving in the randomization procedure (steps 17-18 of Algo. \ref{alg:realize}).
    
Thus, $Q$ is realizable given $\mathcal{G}, \mathbb{A}^\dag$. This is validated by the ancestor set $An(Y_x, X)_\mathcal{G} = \{Y_x, X\}$, which doesn't repeat any variables. This is also illustrated in Fig. \ref{fig:ett_realize_schematic}.

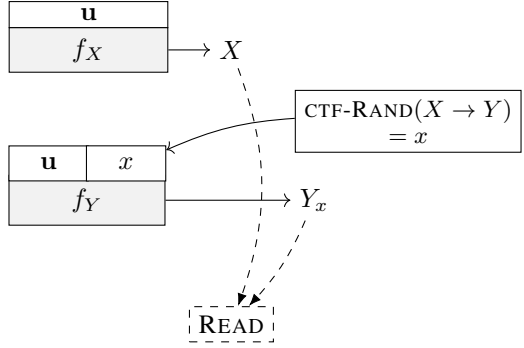
\begin{figure}[h!]
    \vspace{-0.in}
    \begin{center}
    \begin{tikzpicture}

        \node [fill=gray!10,draw,inner sep=0.4em, minimum width=6em] (fx) at (4,-2.85) {$f_X$};
        \node (x) at (5.9,-2.85) {$X$};
        \node [fill=gray!0,draw,inner sep=0.27em, minimum width=6em] (ux) at (4,-2.38) {$\*u$};
        \path [->] (fx) edge (x);

        \node [fill=gray!10,draw,inner sep=0.5em, minimum width=5.9em] (fy) at (3.98,-4.85) {$f_Y$};
        \node (y) at (7,-4.85) {{$Y_x$}};
        \node [fill=gray!0,draw,inner sep=0.44em, minimum width=3em] (uy) at (3.47,-4.36) {$\*u$};
        \node [fill=gray!0,draw,inner sep=0.45em, minimum width=3em] (xy) at (4.5,-4.36) {$x$};
        \node[align=center,draw, font=\small] (ctfrandxy) at (8.25,-3.8) {\textsc{ctf-Rand}$(X \rightarrow Y)$\\$=x$};
        \path [->] (fy) edge (y);
        \path[->] (ctfrandxy) edge [bend right=10] (xy);

        \node[align=left,draw,dashed] (expr) at (5.9,-6.5) {\textsc{Read}};
        \path [-Latex,dashed] (x) edge [bend left=20] (expr);
        \path [-Latex,dashed] (y) edge[bend left=10] (expr);

    \end{tikzpicture}
    \end{center}
    \vskip -0.in
    \caption{$P(Y_x,X)$ is realizable given the graph in Fig. \ref{fig:app_realize_examples_a} and $\mathbb{A}^\dag(\mathcal{G}).$}
    \label{fig:ett_realize_schematic}
    \vskip -0.in
    \end{figure}

However, suppose the agent's action set is

$\mathbb{A} = \{\textsc{Rand}(X)\}$, i.e., does not permit any counterfactual randomization procedures. 

In this case,

\textbf{CTF-REALIZE}$(Q,\mathcal{G},  \mathbb{A})$ trace:
    \begin{itemize}
        \item Start with $X$ (first in topological order)
        \item For the first term in $\*W_\star$: $Y_x$
        \begin{itemize}
            \item Since $X \in An(Y)$, call Algo. \ref{alg:subroutine} \textbf{COMPATIBLE}($X,Y_x$)
            \begin{itemize}
                \item $Y \in Ch(X)$ and $Y \in An(Y)$
                \item $X \in $ subscript of $Y_x$
                \item $\textsc{Rand}(X) \in \mathbb{A}$, and no other ctf-randomization procedure
                \item $\text{INT}_X \gets \{\textsc{Rand}(X): x\}$
            \end{itemize}
        \end{itemize}
        \item For the second term in $\*W_\star$: $X$
        \begin{itemize}
            \item $\text{OUTPUT}_X \gets \{X\}$
        \end{itemize}
        \item Moving to $Y$ (next in topological order)
        \item For the first term in $\*W_\star$: $Y_x$
        \begin{itemize}
            \item $\text{OUTPUT}_Y \gets \{Y_x\}$
        \end{itemize}
        \item $\text{OUTPUT}_X$ contains $X$, but the intervention set $\text{INT}_X$ contains $\textsc{Rand}(X)$
        \item \textcolor{red}{FAIL} (Line 22 of Algo. \ref{alg:realize})
    \end{itemize} $\hfill \blacksquare$
\end{example}


\medskip
\begin{example}(Probability of sufficiency (PS) realizability) \label{ex:realize_2}

    Query, $Q = P(Y_x, X,Y)$ 
    
    Graph, $\mathcal{G}:$ Fig. \ref{fig:app_realize_examples_b}

 \begin{figure}[h!]
        \centering
        \begin{tikzpicture}
        \node (X) at (0,0.4) {$X$};
        \node (Y) at (2,0.4) {$Y$};
        \path [->] (X) edge (Y);
    \end{tikzpicture}
    \caption{Graph for Example \ref{ex:realize_2}}
    \label{fig:app_realize_examples_b}
\end{figure}
    
    Suppose action set $\mathbb{A} = \mathbb{A}^\dag(\mathcal{G}) := \{\textsc{ctf-Rand}(X \rightarrow Y)\}$
    
    \textbf{CTF-REALIZE}$(Q,\mathcal{G},  \mathbb{A}^\dag(\mathcal{G}))$ trace:
    \begin{itemize}
        \item Start with $X$ (first in topological order)
        \item For the first term in $\*W_\star$: $Y_x$
        \begin{itemize}
            \item Since $X \in An(Y)$, call Algo. \ref{alg:subroutine} \textbf{COMPATIBLE}($X,Y_x$)
            \begin{itemize}
                \item $Y \in Ch(X)$ and $Y \in An(Y)$
                \item $X \in $ subscript of $Y_x$
                \item $\textsc{ctf-Rand}(X \rightarrow Y) \in \mathbb{A}^\dag(\mathcal{G})$
                \item $\text{INT}_X \gets \{\textsc{ctf-Rand}(X \rightarrow Y): x\}$
            \end{itemize}
        \end{itemize}
        \item For the second term in $\*W_\star$: $X$
        \begin{itemize}
            \item $\text{OUTPUT}_X \gets \{X\}$
        \end{itemize}
        \item For the third term in $\*W_\star$: $Y$
        \begin{itemize}
            \item Since $X \in An(Y)$, call Algo. \ref{alg:subroutine} \textbf{COMPATIBLE}($X,Y$)
            \begin{itemize}
                \item $Y \in Ch(X)$ and $Y \in An(Y)$
                \item $X \not \in $ subscript of $Y$ ; $X$ needs to be received naturally
                \item But $\text{INT}_X$ already contains $\{\textsc{ctf-Rand}(X \rightarrow Y) : x\}$ with label $x$ $\neq $ "Natural"
                \item \textcolor{red}{FAIL} (Line 25 of Algo. \ref{alg:subroutine})
            \end{itemize}
        \end{itemize}
    \end{itemize}
    
Thus, $Q$ is not realizable given $\mathcal{G}, \mathbb{A}^\dag$. This is validated by the ancestor set $An(Y_x, X,Y)_\mathcal{G} = \{Y_x, X,Y\}$, which contains both $Y_x,Y$. This is also illustrated in Fig. \ref{fig:ps_realize_schematic}.

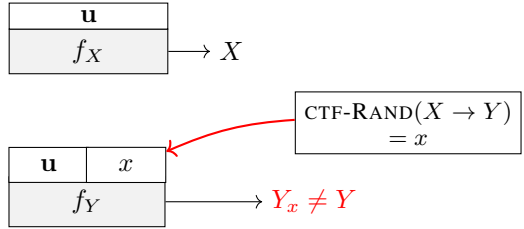
\begin{figure}[h!]
    \vspace{-0.in}
    \begin{center}
    \begin{tikzpicture}

        \node [fill=gray!10,draw,inner sep=0.4em, minimum width=6em] (fx) at (4,-2.85) {$f_X$};
        \node (x) at (5.9,-2.85) {$X$};
        \node [fill=gray!0,draw,inner sep=0.27em, minimum width=6em] (ux) at (4,-2.38) {$\*u$};
        \path [->] (fx) edge (x);

        \node [fill=gray!10,draw,inner sep=0.5em, minimum width=5.9em] (fy) at (3.98,-4.85) {$f_Y$};
        \node[red] (y) at (7,-4.85) {{$Y_x \neq Y$}};
        \node [fill=gray!0,draw,inner sep=0.44em, minimum width=3em] (uy) at (3.47,-4.36) {$\*u$};
        \node [fill=gray!0,draw,inner sep=0.45em, minimum width=3em] (xy) at (4.5,-4.36) {$x$};
        \node[align=center,draw, font=\small] (ctfrandxy) at (8.25,-3.8) {\textsc{ctf-Rand}$(X \rightarrow Y)$\\$=x$};
        \path [->] (fy) edge (y);
        \path[->,red,thick] (ctfrandxy) edge [bend right=10] (xy);


    \end{tikzpicture}
    \end{center}
    \vskip -0.in
    \caption{$P(Y_x,X,Y)$ is not realizable given the graph in Fig. \ref{fig:app_realize_examples_b} and $\mathbb{A}^\dag(\mathcal{G}).$}
    \label{fig:ps_realize_schematic}
    \vskip -0.in
    \end{figure} $\hfill \blacksquare$
\end{example}


\medskip
\begin{example}
\label{ex:realize_3}

    Query, $Q = P(W_{xt}, Z_{x'})$ 
    
    Graph, $\mathcal{G}:$ Fig. \ref{fig:app_realize_examples_c}

     \begin{figure}[h!]
        \centering
        \begin{tikzpicture}
        \node (T) at (0,0) {$T$};
        \node (X) at (2,0) {$X$};
        \node (A) at (1,-0.5) {$A$};
        \node (W) at (0,-1) {$W$};
        \node (Z) at (2,-1) {$Z$};
        \path [->] (T) edge (A);
        \path [->] (X) edge (A);
        \path [->] (A) edge (W);
        \path [->] (A) edge (Z);
    \end{tikzpicture}
    \caption{Graph for Example \ref{ex:realize_3}}
    \label{fig:app_realize_examples_c}
    \end{figure}
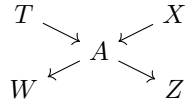
    
    Suppose action set $\mathbb{A} = \mathbb{A}^\dag(\mathcal{G}) := \{\textsc{ctf-Rand}(T \rightarrow A),\textsc{ctf-Rand}(X \rightarrow A),\textsc{ctf-Rand}(A \rightarrow W),\textsc{ctf-Rand}(A \rightarrow Z)\}$
    
    \textbf{CTF-REALIZE}$(Q,\mathcal{G},  \mathbb{A}^\dag(\mathcal{G}))$ trace:
    \begin{itemize}
        \item Start with $X$ (first in topological order)
        \item For the first term in $\*W_\star$: $W_{xt}$
        \begin{itemize}
            \item Since $X \in An(W)$, call Algo. \ref{alg:subroutine} \textbf{COMPATIBLE}($X,W_{xt}$)
            \begin{itemize}
                \item $A \in Ch(X)$ and $A \in An(W)$
                \item $X \in $ subscript of $W_{xt}$
                \item $\textsc{ctf-Rand}(X \rightarrow A) \in \mathbb{A}^\dag(\mathcal{G})$
                \item $\text{INT}_X \gets \{\textsc{ctf-Rand}(X \rightarrow A): x\}$
            \end{itemize}
        \end{itemize}
        \item For the second term in $\*W_\star$: $Z_{x'}$
        \begin{itemize}
            \item Since $X \in An(Z)$, call Algo. \ref{alg:subroutine} \textbf{COMPATIBLE}($X,Z_{x'}$)
            \begin{itemize}
                \item $A \in Ch(X)$ and $A \in An(Z)$
                \item $X \in $ subscript of $Z_{x'}$ ; $X$ needs to be fixed as $x'$
                \item But $\text{INT}_X$ already contains $\{\textsc{ctf-Rand}(X \rightarrow A) : x\}$ with label $x \neq  x'$
                \item \textcolor{red}{FAIL} (Line 8 of Algo. \ref{alg:subroutine})
            \end{itemize}
        \end{itemize}
    \end{itemize}
    
Thus, $Q$ is not realizable given $\mathcal{G}, \mathbb{A}^\dag$. This is validated by the ancestor set $An(W_{xt}, Z_{x'})_\mathcal{G} = \{W_{xt}, A_{xt}, Z_{x'}, A_{x'}\}$, which contains both $A_{xt},A_{x'}$. This is also illustrated in Fig. \ref{fig:ex3_realize_schematic}.

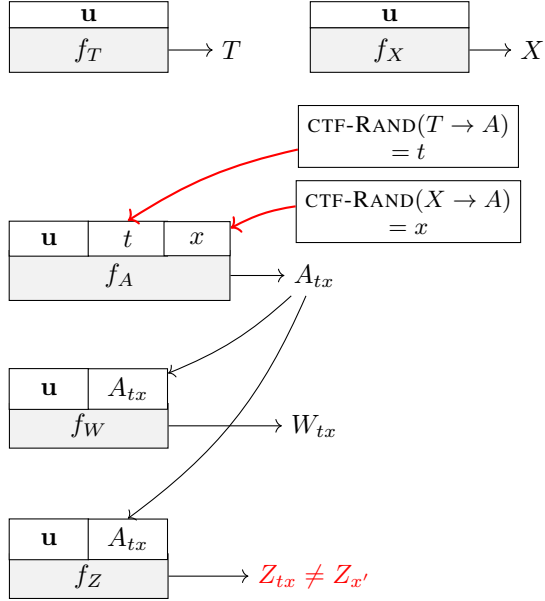
\begin{figure}[h!]
    \vspace{-0.in}
    \begin{center}
    \begin{tikzpicture}

        \node [fill=gray!10,draw,inner sep=0.4em, minimum width=6em] (ft) at (4,-0.85) {$f_T$};
        \node (t) at (5.9,-0.85) {$T$};
        \node [fill=gray!0,draw,inner sep=0.27em, minimum width=6em] (ut) at (4,-0.38) {$\*u$};
        \path [->] (ft) edge (t);
        
        \node [fill=gray!10,draw,inner sep=0.4em, minimum width=6em] (fx) at (8,-0.85) {$f_X$};
        \node (x) at (9.9,-0.85) {$X$};
        \node [fill=gray!0,draw,inner sep=0.27em, minimum width=6em] (ux) at (8,-0.38) {$\*u$};
        \path [->] (fx) edge (x);

        \node [fill=gray!10,draw,inner sep=0.5em, minimum width=8.35em] (fa) at (4.41,-3.85) {$f_A$};
        \node (a) at (7,-3.85) {$A_{tx}$};
        \node [fill=gray!0,draw,inner sep=0.44em, minimum width=3em] (ua) at (3.47,-3.36) {$\*u$};
        \node [fill=gray!0,draw,inner sep=0.35em, minimum width=3em] (ta) at (4.52,-3.36) {$t$};
        \node [fill=gray!0,draw,inner sep=0.43em, minimum width=2.5em] (xa) at (5.44,-3.36) {$x$};
        \node[align=center,draw, font=\small] (ctfrandta) at (8.25,-2) {\textsc{ctf-Rand}$(T\rightarrow A)$\\$=t$};
        \node[align=center,draw, font=\small] (ctfrandxa) at (8.25,-3) {\textsc{ctf-Rand}$(X\rightarrow A)$\\$=x$};
        \path [->] (fa) edge (a);
        \path[->,red,thick] (ctfrandta) edge [bend right=10] (ta.north);
        \path[->,red,thick] (ctfrandxa) edge [bend right=10] (xa);

        \node [fill=gray!10,draw,inner sep=0.4em, minimum width=6em] (fw) at (4,-5.85) {$f_W$};
        \node (w) at (7,-5.85) {{$W_{tx}$}};
        \node [fill=gray!0,draw,inner sep=0.55em, minimum width=3em] (uw) at (3.47,-5.36) {$\*u$};
        \node [fill=gray!0,draw,inner sep=0.35em, minimum width=3em] (aw) at (4.52,-5.36) {$A_{tx}$};
        \path [->] (fw) edge (w);
        \path[->] (a) edge [bend left=10] (aw);

        \node [fill=gray!10,draw,inner sep=0.4em, minimum width=6em] (fz) at (4,-7.85) {$f_Z$};
        \node (z) at (7,-7.85) {\textcolor{red}{$Z_{tx} \neq Z_{x'}$}};
        \node [fill=gray!0,draw,inner sep=0.55em, minimum width=3em] (uz) at (3.47,-7.36) {$\*u$};
        \node [fill=gray!0,draw,inner sep=0.35em, minimum width=3em] (az) at (4.52,-7.36) {$A_{tx}$};
        \path [->] (fz) edge (z);
        \path [->] (a) edge [bend left=15] (az.north);

    \end{tikzpicture}
    \end{center}
    \vskip -0.in
    \caption{$P(W_{xt},Z_{x'})$ is not realizable given the graph in Fig. \ref{fig:app_realize_examples_c} and $\mathbb{A}^\dag(\mathcal{G}).$}
    \label{fig:ex3_realize_schematic}
    \vskip -0.in
    \end{figure} $\hfill \blacksquare$
\end{example}


\medskip
\begin{example}
\label{ex:realize_4}

    Query, $Q = P(Y_x, Z_{x'}, W_{x''})$ 
    
    Graph, $\mathcal{G}:$ Fig. \ref{fig:app_realize_examples_d}

    \begin{figure}[h!]
        \centering
        \begin{tikzpicture}
        \node (X) at (0,0) {$X$};
        \node (Y) at (-1,-1) {$Y$};
        \node (Z) at (0,-1) {$Z$};
        \node (W) at (1,-1) {$W$};
        \path [->] (X) edge (Y);
        \path [->] (X) edge (Z);
        \path [->] (X) edge (W);
    \end{tikzpicture}
    \caption{Graph for Example \ref{ex:realize_4}}
    \label{fig:app_realize_examples_d}
\end{figure}
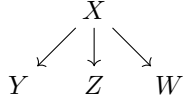
    
    Suppose action set $\mathbb{A} = \{\textsc{Rand}(X), \textsc{ctf-Rand}(X \rightarrow \{Z,W\})\}$
    
    \textbf{CTF-REALIZE}$(Q,\mathcal{G},  \mathbb{A})$ trace:
    \begin{itemize}
        \item Start with $X$ (first in topological order)
        \item For the first term in $\*W_\star$: $Y_x$
        \begin{itemize}
            \item Since $X \in An(Y)$, call Algo. \ref{alg:subroutine} \textbf{COMPATIBLE}($X,Y_x$)
            \begin{itemize}
                \item $Y \in Ch(X)$ and $Y \in An(Y)$
                \item $X \in $ subscript of $Y_{x}$
                \item $\textsc{Rand}(X) \in \mathbb{A}$ ; no other ctf-randomization procedures affecting $Y$
                \item $\text{INT}_X \gets \{\textsc{Rand}(X): x\}$
            \end{itemize}
        \end{itemize}
        \item For the second term in $\*W_\star$: $Z_{x'}$
        \begin{itemize}
            \item Since $X \in An(Z)$, call Algo. \ref{alg:subroutine} \textbf{COMPATIBLE}($X,Z_{x'}$)
            \begin{itemize}
                \item $Z \in Ch(X)$ and $Z \in An(Z)$
                \item $X \in $ subscript of $Z_{x'}$
                \item $\textsc{ctf-Rand}(X \rightarrow \{Z,W\}) \in \mathbb{A}$ ; no smaller ctf-randomization procedures affecting $Z$
                \item $\text{INT}_X \gets \text{INT}_X \cup  \{\textsc{ctf-Rand}(X \rightarrow \{Z,W\}): x'\}$
            \end{itemize}
        \end{itemize}
        \item For the third term in $\*W_\star$: $W_{x''}$
        \begin{itemize}
            \item Since $X \in An(W)$, call Algo. \ref{alg:subroutine} \textbf{COMPATIBLE}($X,W_{x''}$)
            \begin{itemize}
                \item $W \in Ch(X)$ and $W \in An(W)$
                \item $X \in $ subscript of $W_{x''}$ ; $X$ needs to be fixed as $x''$
                \item But $\text{INT}_X$ already contains $\{\textsc{ctf-Rand}(X \rightarrow \{Z,W\}) : x'\}$ with label $x' \neq  x''$
                \item No smaller ctf-randomization procedures affecting $W$
                \item \textcolor{red}{FAIL} (Line 8 of Algo. \ref{alg:subroutine})
            \end{itemize}
        \end{itemize}
    \end{itemize}
    
Thus, $Q$ is not realizable given $\mathcal{G}, \mathbb{A}$. 

However, suppose instead that action set $\mathbb{A}' = \{\textsc{Rand}(X), \textsc{ctf-Rand}(X \rightarrow \{Z,W\}),\textsc{ctf-Rand}(X \rightarrow Z)\}$

\textbf{CTF-REALIZE}$(Q,\mathcal{G},  \mathbb{A}')$ trace:
    \begin{itemize}
        \item Start with $X$ (first in topological order)
        \item For the first term in $\*W_\star$: $Y_x$
        \begin{itemize}
            \item Since $X \in An(Y)$, call Algo. \ref{alg:subroutine} \textbf{COMPATIBLE}($X,Y_x$)
            \begin{itemize}
                \item $Y \in Ch(X)$ and $Y \in An(Y)$
                \item $X \in $ subscript of $Y_{x}$
                \item $\textsc{Rand}(X) \in \mathbb{A}$ ; no other ctf-randomization procedures affecting $Y$
                \item $\text{INT}_X \gets \{\textsc{Rand}(X): x\}$
            \end{itemize}
        \end{itemize}
        \item For the second term in $\*W_\star$: $Z_{x'}$
        \begin{itemize}
            \item Since $X \in An(Z)$, call Algo. \ref{alg:subroutine} \textbf{COMPATIBLE}($X,Z_{x'}$)
            \begin{itemize}
                \item $Z \in Ch(X)$ and $Z \in An(Z)$
                \item $X \in $ subscript of $Z_{x'}$
                \item $\textsc{ctf-Rand}(X \rightarrow Z) \in \mathbb{A}$
                \item $\text{INT}_X \gets \text{INT}_X \cup  \{\textsc{ctf-Rand}(X \rightarrow Z): x'\}$
            \end{itemize}
        \end{itemize}
        \item For the third term in $\*W_\star$: $W_{x''}$
        \begin{itemize}
            \item Since $X \in An(W)$, call Algo. \ref{alg:subroutine} \textbf{COMPATIBLE}($X,W_{x''}$)
            \begin{itemize}
                \item $W \in Ch(X)$ and $W \in An(W)$
                \item $X \in $ subscript of $W_{x''}$
                \item $\textsc{ctf-Rand}(X \rightarrow \{Z,W\}) \in \mathbb{A}$
                \item $\text{INT}_X \gets \text{INT}_X \cup  \{\textsc{ctf-Rand}(X \rightarrow \{Z,W\}): x''\}$
            \end{itemize}
        \end{itemize}
    \item Moving to $Y$ (next in topological order)
        \begin{itemize}
            \item $\text{OUTPUT}_Y \gets \{Y_x\}$
        \end{itemize}
    \item Moving to $Z$ (next in topological order)
        \begin{itemize}
            \item $\text{OUTPUT}_Z \gets \{Z_{x'}\}$
        \end{itemize}
    \item Moving to $W$ (next in topological order)
        \begin{itemize}
            \item $\text{OUTPUT}_W \gets \{W_{x''}\}$
        \end{itemize}
        \item Perform interventions in $\text{INT}_X$, followed by $\textsc{READ}$, and assign output vector based on $\text{OUTPUT}_Y, \text{OUTPUT}_Z$,$\text{OUTPUT}_W$
        \item \textcolor{ForestGreen}{Return i.i.d sample}
    \end{itemize}

Thus, $Q$ is realizable given $\mathcal{G}, \mathbb{A}'$. 

Lastly, it is evident that $Q$ is realizable given $\mathcal{G}, \mathbb{A}^\dag$. This is validated by the ancestor set $An(Y_x, Z_{x'}, W_{x''})_\mathcal{G} = \{Y_x, Z_{x'}, W_{x''}\}$, which does not repeat any variables. This is also illustrated in Fig. \ref{fig:ex4_realize_schematic}.

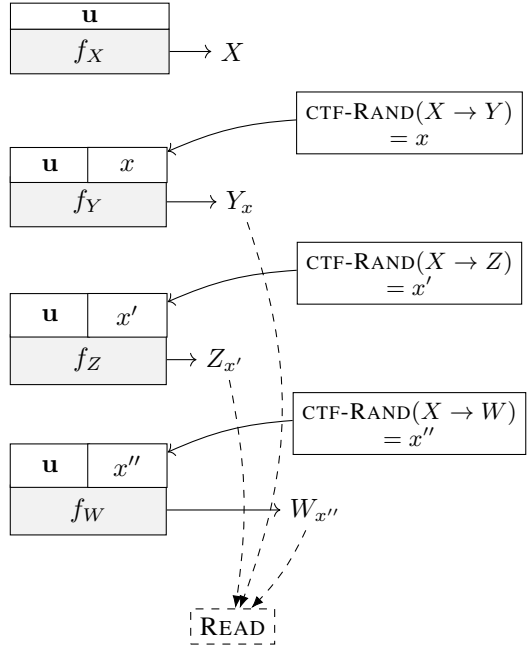
\begin{figure}[h!]
    \vspace{-0.in}
    \begin{center}
    \begin{tikzpicture}

        \node [fill=gray!10,draw,inner sep=0.4em, minimum width=6em] (fx) at (4,-2.85) {$f_X$};
        \node (x) at (5.9,-2.85) {$X$};
        \node [fill=gray!0,draw,inner sep=0.27em, minimum width=6em] (ux) at (4,-2.38) {$\*u$};
        \path [->] (fx) edge (x);

        \node [fill=gray!10,draw,inner sep=0.5em, minimum width=5.9em] (fy) at (3.98,-4.85) {$f_Y$};
        \node (y) at (6,-4.85) {{$Y_x$}};
        \node [fill=gray!0,draw,inner sep=0.44em, minimum width=3em] (uy) at (3.47,-4.36) {$\*u$};
        \node [fill=gray!0,draw,inner sep=0.45em, minimum width=3em] (xy) at (4.5,-4.36) {$x$};
        \node[align=center,draw, font=\small] (ctfrandxy) at (8.25,-3.8) {\textsc{ctf-Rand}$(X \rightarrow Y)$\\$=x$};
        \path [->] (fy) edge (y);
        \path[->] (ctfrandxy) edge [bend right=10] (xy);

        \node [fill=gray!10,draw,inner sep=0.5em, minimum width=5.9em] (fz) at (3.98,-4.85-2.1) {$f_Z$};
        \node (z) at (5.8,-4.85-2.1) {{$Z_{x'}$}};
        \node [fill=gray!0,draw,inner sep=0.6em, minimum width=3em] (uz) at (3.47,-4.36-2) {$\*u$};
        \node [fill=gray!0,draw,inner sep=0.45em, minimum width=3em] (xz) at (4.5,-4.36-2) {$x'$};
        \node[align=center,draw, font=\small] (ctfrandxz) at (8.25,-3.8-2) {\textsc{ctf-Rand}$(X \rightarrow Z)$\\$=x'$};
        \path [->] (fz) edge (z);
        \path[->] (ctfrandxz) edge [bend right=10] (xz);

        \node [fill=gray!10,draw,inner sep=0.5em, minimum width=5.9em] (fw) at (3.98,-4.85-4.1) {$f_W$};
        \node (w) at (7,-4.85-4.1) {{$W_{x''}$}};
        \node [fill=gray!0,draw,inner sep=0.6em, minimum width=3em] (uw) at (3.47,-4.36-4) {$\*u$};
        \node [fill=gray!0,draw,inner sep=0.45em, minimum width=3em] (xw) at (4.5,-4.36-4) {$x''$};
        \node[align=center,draw, font=\small] (ctfrandxw) at (8.25,-3.8-4) {\textsc{ctf-Rand}$(X \rightarrow W)$\\$=x''$};
        \path [->] (fw) edge (w);
        \path[->] (ctfrandxw) edge [bend right=10] (xw);

        \node[align=left,draw,dashed] (expr) at (5.9,-6.5-4) {\textsc{Read}};
        \path [-Latex,dashed] (y) edge [bend left=20] (expr);
        \path [-Latex,dashed] (z) edge[bend left=10] (expr);
        \path [-Latex,dashed] (w) edge[bend left=10] (expr);

    \end{tikzpicture}
    \end{center}
    \vskip -0.in
    \caption{$P(Y_x, Z_{x'}, W_{x''})$ is realizable given the graph in Fig. \ref{fig:app_realize_examples_d} and $\mathbb{A}^\dag(\mathcal{G}).$}
    \label{fig:ex4_realize_schematic}
    \vskip -0.in
    \end{figure} $\hfill \blacksquare$
\end{example}
\newpage
\section{Assumptions and realizability proofs}
\label{app:assumptions_and_proofs}


\subsection{Assumptions}
\label{app:assumptions}

In this sub-section, we gather together all the structural assumptions we make in this paper, for ease of reference. We also include related remarks.

\medskip
\begin{assumption}[Unobservability] \label{assn:unobservability}
    An agent deployed in the environment does not know the underlying SCM $\mathcal{M}$ of the environment, and does not know the latent features $\*U^{(i)}$ of any unit $i$ in the target population. $\hfill \blacksquare$
\end{assumption}

\medskip
\begin{assumption}[Feasible actions] \label{assn:feasible_actions}
    Given causal diagram $\mathcal{G}$, the physical actions that an agent can perform on any unit $i$ in the target population are limited to: $\textsc{Select}^{(i)}$, $\textsc{Read}(V)^{(i)}$, $\textsc{Rand}(X)^{(i)}$, and $\textsc{ctf-Rand}(X \rightarrow \*C)^{(i)}$, for some $V, X \in \*V$ and $\*C \subseteq Ch(X)_{\mathcal{G}}$, per Defs. \ref{def:physical_actions},\ref{def:ctf-rand}. $\hfill \blacksquare$
\end{assumption}

\medskip
\textbf{Assumption \ref{assn:fce}} (Fundamental constraint of experimentation (FCE))\textbf{.} A unit $i$ in the target population can physically undergo a causal mechanism $f_V \in \mathcal{F}$ at most once. $\hfill \blacksquare$

\medskip
{
    We define the probability measure $P^{\mathbb{C}}(.)$ from the perspective of an exogenous agent (i.e., an agent external to the system) $\mathbb{C}$'s beliefs about the environment, distinguished by superscript from $P^{\mathcal{M}}(.)$, the true unknown distribution.  }

\medskip
{
\begin{remark}\label{rem:exogeneity}
    Let $\mathcal{A}^{(i)}$ be a sequence of actions taken by agent $\mathbb{C}$ on unit $i$ that is not conditional on any data gathered regarding $i$. The assumption of $\mathbb{C}$ behaving exogenously means that $P^{\mathbb{C}}(\*U^{(i)} = \*u \mid \mathcal{A}^{(i)}) = P^{\mathcal{M}}(\*u)$.
    $\hfill \blacksquare$
\end{remark}}

\medskip
Regarding the structural conditions involving counterfactual randomization (Def. \ref{def:ctf-rand}), we make the following assumption, mainly as a simplifying step for use in the proofs.

\textbf{Assumption \ref{assn:proxy_tree}} (Tree structure)\textbf{.} Given a variable $X$, causal diagram $\mathcal{G}$, and an "expanded" diagram $\mathcal{G}^+$ (Def. \ref{def:expanded_SCM}) including the set of all the counterfactual mediators $\*W$ (Def. \ref{def:l2_proxy}) of $X$ in the environment, each $W \in \*W$ has only one parent in $\mathcal{G}^+$, and each $C \in Ch(X)_{\mathcal{G}}$ has at most one $W \in \*W$ as a parent in $\mathcal{G}^+$. $\hfill \blacksquare$

\medskip
From this assumption, and from the definition of a counterfactual mediator (Def. \ref{def:l2_proxy}), we can derive the following observations:

\smallskip
\begin{remark}[No bypassing children] \label{rem:nobypassing}
    Given causal diagram $\mathcal{G}$, the procedure $\textsc{ctf-Rand}(X \rightarrow \*C)$, either by eliciting a unit's natural decision or via a counterfactual mediator, can only be performed w.r.t $\*C \subseteq Ch(X)_{\mathcal{G}}$. It cannot by-pass child mechanisms and directly affect a descendant. This is elaborated in App. \ref{app:ctf_rand_constraints}, and specifically in Lemma \ref{lem:no_forks}. $\hfill \blacksquare$ 
\end{remark}

\smallskip
\textit{Remark} \ref{rem:containment} (Procedure containment). Assumption \ref{assn:proxy_tree} implies that if an agent is capable of performing both $\textsc{ctf-Rand}(X \rightarrow \*C)^{(i)}$ and $\textsc{ctf-Rand}(X \rightarrow \*C')^{(i)}$ s.t. $\*C \neq \*C'$ and $\*C \cap \*C' \neq \emptyset$, then either $\*C \subseteq \*C'$ or $\*C' \subseteq \*C$. $\hfill \blacksquare$ 

\smallskip
\textit{Remark} \ref{remark:supersede} (Superseding action). Given a decision variable $X$, the action $\textsc{ctf-Rand}(X \rightarrow \*C')^{(i)}$ can \textit{supersede} the action $\textsc{ctf-Rand}(X \rightarrow \*C)^{(i)}$ if $\*C' \subsetneq \*C$, where \textit{supersede} means that the former action $\textsc{ctf-Rand}(X \rightarrow \*C')^{(i)}$ blocks any effect that the latter action has on the variables $\*C'$. Additionally, the action $\textsc{ctf-Rand}(X \rightarrow \*C)^{(i)}$ \textit{supersedes} the action $\textsc{Rand}(X)^{(i)}$. $\hfill \blacksquare$

\smallskip
Counterfactual randomization permits multiple randomizations for the same variable $X$ for a single unit $i$. But some randomizations block the effects of others. See App. \ref{app:multiple_rands}.


\bigskip
\bigskip
\subsection{Proofs for Section \ref{sec:realizability_main_sec}}
\label{app:realizability_proofs}

Recall, $P^{\mathbb{C}}(.)$ is the probability measure from the perspective of an exogenous agent (i.e., an agent external to the system) $\mathbb{C}$'s beliefs about the environment, distinguished by superscript from $P^{\mathcal{M}}(.)$, the true unknown distribution. 

Since unit selection is randomized, $\textsc{Select}^{(i)}$ yields an unbiased sample of a unit with latent features distributed according to the target population frequency $P(\*u)$. I.e., $P^{\mathbb{C}}(\*U^{(i)} = \*u \mid \textsc{Select}^{(i)}) = P^{\mathcal{M}}(\*u)$.

\begin{lemma}[I.i.d requirement] \label{lem:iid_requirement}
    Consider a sequence of actions $\mathcal{A}^{(i)}$ performed on unit $i$ in the target population, that yields a vector of realized values $\*W^{(i)}_\star$. $\*W^{(i)}_\star$ is an i.i.d sample from $P^\mathcal{M}(\*W_\star)$, for arbitrary $\mathcal{M}$ iff
    \begin{itemize}
        \item [i.] $P^\mathbb{C}(\*U^{(i)}=\*u \mid \mathcal{A}^{(i)}) = P^\mathcal{M}(\*U = \*u)$; and
        \item [ii.] $\mathbbm{1}[\*W^{(i)}_\star= \*w \mid \mathcal{A}^{(i)}, \*U^{(i)}=\*u] = \mathbbm{1}[\*W_\star(\*u) = \*w]$, $\forall \*w, \*u$.
    \end{itemize}
\end{lemma}
\begin{proof}
    Recall from Def. \ref{def:iid} that $\*W^{(i)}_\star$ being an i.i.d sample from $P^\mathcal{M}(\*W_\star)$ means that 
    \begin{align}
        P^{\mathbb{C}}(\*W_\star^{(i)} = \*w \mid \mathcal{A}^{(i)}) = P^{\mathcal{M}}(\*W_\star = \*w), \forall \*w \label{eq:iid_proof_1}
    \end{align}
    
    \underline{Reverse direction:}

    We simply multiply respective l.h.s and r.h.s of conditions [i] and [ii] and sum over all $\*u$ to get
    \begin{align}
        \sum_\*u P^\mathbb{C}(\*U^{(i)}=\*u \mid \mathcal{A}^{(i)}).\mathbbm{1}[\*W^{(i)}_\star= \*w \mid \mathcal{A}^{(i)}, \*U^{(i)}=\*u] &= \sum_\*u P^\mathcal{M}(\*U = \*u).\mathbbm{1}[\*W_\star(\*u) = \*w]\\
        & = P^\mathcal{M}(\*W_\star = \*w),
    \end{align}
    which we get from the Layer 3 valuation formula (see preliminaries in Sec. \ref{sec:intro}). On the l.h.s, we apply the chain rule to get the result we need.
    \begin{align}
        P^{\mathbb{C}}(\*W_\star^{(i)} = \*w \mid \mathcal{A}^{(i)}) = P^{\mathcal{M}}(\*W_\star = \*w)
    \end{align}

    \underline{Forward direction:}

    Assume Eq. \ref{eq:iid_proof_1}.
    


    From Remark \ref{rem:exogeneity}, we conclude that condition [i] automatically holds. Since the agent acts exogenously to the system, $P^\mathbb{C}(\*U^{(i)}=\*u \mid \mathcal{A}^{(i)}) = P^\mathcal{M}(\*U = \*u), \forall \*u$.

    Applying the chain rule on both sides of Eq. \ref{eq:iid_proof_1},
    \begin{align}
        \sum_\*u P^\mathbb{C}(\*U^{(i)}=\*u \mid \mathcal{A}^{(i)}).\mathbbm{1}[\*W^{(i)}_\star= \*w \mid \mathcal{A}^{(i)}, \*U^{(i)}=\*u] &= \sum_\*u P^\mathcal{M}(\*U = \*u).\mathbbm{1}[\*W_\star(\*u) = \*w]
    \end{align}
    The probability terms are equal, for each $\*u$.
    
    Since the probability terms are free parameters, and we need this equation to hold for any arbitrary probability simplex, it must be the case that the indicator terms are also equal. Thus begetting condition [ii].


    
\end{proof}

\bigskip
\begin{lemma} \label{lem:realizable_proofs_1}
    Given a causal diagram $\mathcal{G}$, for any SCM $\mathcal{M}$ compatible with $\mathcal{G}$, the jointly necessary and sufficient conditions to measure a potential response $W_\*t(\*u)$ are {[i]} $\*T$ is fixed as $\*t$ (by intervention) as an input to all children $C \in Ch(\*T) \cap An(W)$; {[ii]} each $A \in An(W)_{\mathcal{G}_{\overline{\*T}}}$, $A \not \in \{\*T, W\}$ is received "naturally" (i.e., without intervention) by its children $C \in Ch(A) \cap An(W)$; and [iii] the mechanism $f_W$ is not erased and overwritten (by a Fisherian intervention). 
\end{lemma}
\begin{proof}
    $W_\*t$ is the variable $W$ evaluated in the sub-model $\mathcal{M}_\*t$, where the equations for $\*T$ are replaced by constant values in $\*t$.

    For any changes to the function for $T \in \*T$, the function $f_W$ is only affected by any effect on the children of $T$ which are also ancestors of $W$. Any effect of $T$ on some $C' \in Ch(T)$ s.t. $C' \not \in An(W)$ has no effect on $W$.

    Further, in the submodel $\mathcal{M}_\*t$ there are no interventions on any other ancestors of $W$ in $\mathcal{G}_{\overline{\*T}}$, besides $\*T$. Even if there were tnterventions involving some $X \not \in An(W)_{\mathcal{G}_{\overline{\*T}}}$, this would have no effect on $W$ in the sub-model $\mathcal{M}_\*t$, by Rule 3 of do-calculus.

    It is evident that $f_W$ evaluated according to the sub-model $\mathcal{M}_\*t$, and evaluated according to a sub-model satisfying conditions [i] and [ii] are identical for each $\*u$, since the sequence of structural equations that eventually generate $W$ are the same.

    Finally, in order to measure $W_\*t$, we need to measure the output of the mechanism $f_W$ in the real world. The mechanism cannot not be erased and overwritten, as per condition [iii].
\end{proof}

\bigskip
\begin{lemma} \label{lem:realizable_proofs_2}
    Given a set $\*W_\star$ and graph $\mathcal{G}$, where each member $W_\*t \in \*W_\star$ has its respective conditions [i-iii] (per Lemma \ref{lem:realizable_proofs_1}), suppose these conditions introduce conflicts when combined across $\*W_\star$. Removing $X$ from $\*W_\star$ and from all subscripts in $\*W_\star$ to get a new set $\*W'_\star$ does not introduce new conflicts between the terms, if $X$ is first in a topological ordering of $\mathcal{G}$.
\end{lemma}
\begin{proof}
    Let us consider the conditions in the necessary-and-sufficient set given in Lemma \ref{lem:realizable_proofs_1} for each $W_\*t \in \*W_\star$.

    Condition [i] add requirements for each $t \in \*t$ of some $W_\*t \in \*W_\star$. Since $X$ is removed from every subscript, this no longer applies to any term in $\*W'_\star$.

    Condition [ii] requires that if $X$ is an ancestor of some $W_\*t \in \*W_\star$, and it doesn't appear in $\*T$, then $X$ must be received without intervention by mediating children. Since $X$ is removed from every subscript, $X$ not being intervened upon at all meets this condition [ii] for every term $\*W'_\star$, without conflicting with condition [i], which no longer applies.

    Importantly, even though $X$ is no longer being intervened upon, for each $W_\*t \in \*W_\star$, removing $X$ from $\*T$ (if it appears) does not add any additional ancestors in $\mathcal{G}_{\overline{T}}$ that need to be tracked for condition [ii], since $X$ is first in topological order.

    Condition [iii] would only apply to $X$ itself, which is not present as a potential response in $\*W'_\star$. 
\end{proof}

\bigskip
\paragraph{Theorem \ref{thm:completeness}}

:

Let $\mathcal{A}^{(i)}$ be a sequence of actions conducted by an exogenous agent to beget a vector of values $\*W_\star^{(i)}$ for a unit $i$.

By Lemma \ref{lem:iid_requirement}, if an agent wants $\*W_\star^{(i)}$  to be an i.i.d sample from $P(\*W_\star)$, then for each possible $\*U^{(i)}=\*u$, the vector $\*W_\star^{(i)}$ needs to be identical to the $\*W_\star(\*u)$ as evaluated according to the SCM. Essentially, this says that the agent's actions need to output the same vector $\*W_\star(\*u)$ as if it has been evaluated according to the SCM.

By Lemma \ref{lem:realizable_proofs_1}, $\*W_\star(\*u)$ can be evaluated if and only if the following three conditions are met for each $W_\*t \in \*W_\star$ simultaneously:
\begin{itemize}
    \item [i] $\*T$ is fixed as $\*t$ (by intervention) as an input to all children $C \in Ch(\*T) \cap An(W)$;  
    \item [ii] Each $A \in An(W)_{\mathcal{G}_{\overline{\*T}}}$, $A \not \in \{\*T, W\}$ is received "naturally" (i.e., without intervention) by its children $C \in Ch(A) \cap An(W)$; and
    \item [iii] The mechanism $f_W$ is not erased and overwritten (by a Fisherian intervention).
\end{itemize}

\newpage
\underline{\textbf{Inductive hypothesis (IH)}}:

\textbf{CTF-REALIZE}($P(\*W_\star),\mathcal{G},\mathbb{A})$ returns FAIL if and only if conditions [i-iii] are not met simultaneously, when combined across all $W_\*t \in \*W_\star$ w.r.t a causal diagram $\mathcal{G}$ having $\leq n$ nodes

\underline{\textbf{Base case}}: 

Consider an SCM with only one variable $V \in \*V$. IH is trivially true, since the conditions are always met, and since \textbf{CTF-REALIZE} will just return the value $\textsc{READ}(V)$.

Assume IH is true for any SCM with causal diagram having $\leq n$ nodes.

\underline{\textbf{n+1 case}}: 

Consider an SCM whose causal diagram $\mathcal{G}$ has $n+1$ nodes. Let $X$ be the first in some topological ordering of $\mathcal{G}$. Consider an action set $\mathbb{A}$ that the agent can perform in the environment, and an arbitrary distribution $P(\*W_\star)$.

WLOG, we can begin the outer loop of \textbf{CTF-REALIZE}($P(\*W_\star),\mathcal{G},\mathbb{A})$) with $X$ (first in topological order).

\begin{itemize}
    \item The inner loop calls \textbf{COMPATIBLE}($X,W_\*t$) for each $W \in Desc(X), X \neq W$.
    \item It maintains a tracker $\text{INT}_X$ of the smallest counterfactual interventions needed to satisfy condition [i] for each $W_\*t$, resorting to Fisherian intervention if needed. Note (per Remark \ref{rem:containment}), interventions follow a tree-like structure, so conflicts can be tracked by tagging the smallest available intervention that is needed for each $W_\*t$ w.r.t each child of $X$.
    \item Note also (per Remark \ref{remark:supersede}) that if there are two simultaneous interventions added to $\text{INT}_X$, $\textsc{ctf-rand}(X \rightarrow \*C)$, $\textsc{ctf-rand}(X \rightarrow \*C')$, where $\*C' \subseteq \*C$, then the set $\*C'$ is unaffected by the first procedure.
    \item This inner loop exactly checks if there are any conflicts in conditions [i-ii] among $\*W_\star$ w.r.t $X$, by "tagging" each procedure with the fixed value $x$ needed for that intervention (including the requirement of no intervention). 
    \item Finally the outer loop in Line 20 of Algo. \ref{alg:realize} checks if $X$ appears as a potential response anywhere in $\*W_\star$. If so, $\text{INT}_X$ cannot contain the requirement of Fisherian $\textsc{Rand}(X)$, since this violates condition [iii] w.r.t $X$.
    \item $X$ does not appear anywhere else in subsequent algorithm iterations. 
\end{itemize}

Thus, we conclude that \textbf{CTF-REALIZE}($P(\*W_\star),\mathcal{G},\mathbb{A})$) does not return FAIL on the outer loops evaluated for $X$, if and only if there are no conflicts in the conditions [i-iii] for $\*W_\star$ w.r.t $X$. \textit{In other words, all conflicts w.r.t $X$, in the conditions [i-iii] combined across the terms in $\*W_\star$,  are identified in the algorithm steps that involve $X$}.

Next, define the new set $\*W'_\star$ by dropping $x$ from the subscript (if it appears) for each $W_\*t \in \*W_\star$, and dropping $X$ from $\*W_\star$ (if it appears). Since $X$ is first in topological order of $\mathcal{G}$, this does not add any \textit{new} conflicts across conditions [i-iii] induced by each term in $\*W'_\star$ (by Lemma \ref{lem:realizable_proofs_2}).

It is also clear that if there are conflicts \textit{not} involving $X$, that are induced by conditions [i-iii] across the terms in $\*W_\star$, then these conflicts are also induced by $\*W'_\star$. Suppose there are two terms $W_\*t, Y_\*h \in \*W_\star$ s.t. $\*T\setminus X$ needs to be received as $\*t \setminus X$ by mediating children (condition [i]) for $W_\*t$, and this conflicts with the requirement that $\*T\setminus X$ needs to be received as $\*t' \setminus X$ (or naturally) by the same mediating children, for $Y_\*h$. Removing $X$ does not affect this conflict, since $X$ is topologically prior.

Next, define the graph $\mathcal{G}'$ as the projection of $\mathcal{G}$ that marginalizes out $X$ (and adds bidirected edges if needed). $\mathcal{G}'$ has $\leq n$ nodes. Therefore, from the IH, we conclude that \textbf{CTF-REALIZE}($P(\*W'_\star),\mathcal{G}',\mathbb{A})$ does not return FAIL if and only if there are no conflicts induced by conditions [i-iii], combined across terms in $\*W'_\star$.

Now, we note that \textbf{CTF-REALIZE}($P(\*W_\star),\mathcal{G},\mathbb{A})$ is merely \textbf{CTF-REALIZE}($P(\*W'_\star),\mathcal{G}',\mathbb{A})$, plus all the steps involving $X$ that we discussed earlier (can be verified from inspecting the algorithm - the former has an outer loop involving $X$ and then contains the same steps as the latter). \textit{Therefore, all conflicts induced by conditions [i-iii] that involve $X$ and do not involve $X$ are identified in the algorithm steps when run on $\mathcal{G}$ and $\*W_\star$}.

Thus, we show that \textbf{CTF-REALIZE}($P(\*W_\star),\mathcal{G},\mathbb{A})$ returns FAIL if and only if conditions [i-iii] are not met simultaneously, when combined across all $W_\*t \in \*W_\star$ w.r.t a causal diagram having $\leq n+1$ nodes. The IH stands proved.

By Lemma \ref{lem:realizable_proofs_1}, we know that conditions [i-iii] are necessary and sufficient to evaluate each term in $\*W_\star(\*u)$ simultaneously, for any SCM compatible with $\mathcal{G}$. By Lemma \ref{lem:iid_requirement}, we know that this is equivalent to drawing an i.i.d sample from $P(\*W_\star)$. This gives us the proof of the theorem.

Note: we don't discuss the rejection sampling steps involved steps 17-18 of Algo. \ref{alg:realize} as this is trivially equivalent to intervening using a fixed value. $\hfill \blacksquare$

\bigskip
\paragraph{Corollary \ref{cor:maximal_amwn}}

:

The proof intuition is as follows: given a graph $\mathcal{G}$ and a potential response $Y_\*x$, the set of (counterfactual) ancestors of $Y_\*x$ \citep{correaetal:21} lists each ancestor of $Y$ and \textit{what regime it must be realized in}, in order for $Y_\*x$ to be evaluated. In other words $An(Y_\*x)$ tracks the regimes necessary and sufficient for its ancestors to be evaluated under to beget $Y_\*x$. 

For instance, in graph $\mathcal{G}_1$ in Fig. \ref{fig:realize_examples}, in order to evaluate $W_t$, we need $A_t$ to be evaluated in the regime $\mathcal{M}_t$. In order to evaluate $Z_x$, we need $A, T$ to both be evaluated naturally. This reveals a conflict at the bottleneck $f_A$, which renders the distribution non-realizable.

Thus, Corollary \ref{cor:maximal_amwn} provides a \textit{sufficient} condition to conclude that a distribution is non-realizable, if $An(\*W_\star)$ contains two potential responses of the same variable under different regimes. It also becomes a \textit{necessary} condition for non-realizability, if the agent can perform $\textsc{ctf-Rand}(X \rightarrow C)$, separately for each $C \in Ch(X)$, for all $X$. I.e., if the action set is $\mathbb{A}^\dag(\mathcal{G})$.

The proof steps are similar to Theorem \ref{thm:completeness}.

\underline{\textbf{Inductive Hypothesis (IH)}:}

Given a graph $\mathcal{G}$ with $\leq n$ nodes, and an arbitrary distribution $\*W_\star$, \textbf{CTF-REALIZE}($P(\*W_\star),\mathcal{G},\mathbb{A}^\dag(\mathcal{G}))$ if and only if $An(\*W_\star)$ does not contain a pair of potential responses $W_\*t, W_\*s$ of the same variable $W$ under different regimes.

\underline{\textbf{Base case}:}

For a graph containing only one variable $Y$, this is trivially true. $An(Y) = Y$, and the distribution $P(Y)$ is realizable.

Assume IH is true for a graph of $\leq n$ nodes.

\underline{\textbf{n+1 case}:}

Consider an SCM whose causal diagram $\mathcal{G}$ has $n+1$ nodes. Let $X$ be the first in some topological ordering of $\mathcal{G}$. The agent can perform $\mathbb{A}^\dag$ in the environment, and the distribution is some arbitrary $P(\*W_\star)$. WLOG, we can begin the outer loop of \textbf{CTF-REALIZE}($P(\*W_\star),\mathcal{G},\mathbb{A})$) with $X$ (first in topological order).

From Lemmas \ref{lem:realizable_proofs_1} and \ref{lem:iid_requirement}, we know that conditions [i-iii] for each $W_t \in \*W_\star$, combined across $\*W_\star$ form a necessary and sufficient set to realize $P(\*W_\star)$.

Note that condition [iii] is always satisfied because the agent need never perform a Fisherian $\textsc{Rand}(V)$ for any $V$. It can get the same effect by performing $\textsc{ctf-Rand}(V \rightarrow C)$ for each $C \in Ch(V)$. Step 12 of the sub-routine, Algo. \ref{alg:subroutine} would never be invoked.

From Theorem \ref{thm:completeness}, we know that \textbf{CTF-REALIZE}($P(\*W_\star),\mathcal{G},\mathbb{A}^\dag(\mathcal{G}))$ returns FAIL if and only if there are conflicts in conditions [i-ii] when combined across all the terms $\*W_\star$.

Define the new set $\*W'_\star$ by dropping $x$ from the subscript (if it appears) for each $W_\*t \in \*W_\star$, and dropping $X$ from $\*W_\star$ (if it appears). Since $X$ is first in topological order of $\mathcal{G}$, this does not add any \textit{new} conflicts across conditions [i-ii] induced by each term in $\*W'_\star$ (by Lemma \ref{lem:realizable_proofs_2}). It also doesn't \textit{remove} any conflicts that are not related to $X$, as argued in the proof of Theorem \ref{thm:completeness}, since $X$ comes topologically first.

Define the graph $\mathcal{G}'$ as the projection of $\mathcal{G}$ that marginalizes out $X$ (and adds bidirected edges if needed). $\mathcal{G}'$ has $\leq n$ nodes. From the IH, we conclude that \textbf{CTF-REALIZE}($P(\*W'_\star),\mathcal{G}',\mathbb{A}^\dag(\mathcal{G}'))$ does not return FAIL if and only if $An(\*W'_\star)$ does not contain two potential responses $W_\*t, W_\*s$ of the same variable under different regimes.

However, note that (as discussed in the proof of Theorem \ref{thm:completeness}, and from inspecting the algorithm), the only difference between \textbf{CTF-REALIZE}($P(\*W_\star),\mathcal{G},\mathbb{A}^\dag(\mathcal{G}))$ and \textbf{CTF-REALIZE}($P(\*W'_\star),\mathcal{G}',\mathbb{A}^\dag(\mathcal{G}'))$ is that in the former, the outer loop of \textbf{CTF-REALIZE} first checks for conflicts in the conditions [i-ii] across $\*W_\star$ w.r.t $X$. After that, the steps for both algorithms are the identical.

Therefore, any conflicts detected by \textbf{CTF-REALIZE}($P(\*W_\star),\mathcal{G},\mathbb{A}^\dag(\mathcal{G}))$ that are \textit{not} detected by \textbf{CTF-REALIZE}($P(\*W'_\star),\mathcal{G}',\mathbb{A}^\dag(\mathcal{G}'))$ must be conflicts w.r.t $X$. By the IH, these additional conflicts (unrelated to $X$) cannot be because of a pair of conflicting potential responses in $An(\*W'_\star)$.

We have already established that removing $X$ to make $\*W'_\star$ does not remove or add any conflicting potential response pairs that don't involve $X$. Therefore, our task is to now show that each of these additional conflicts (involving $X$) must correspond to at least one conflicting pair of potential responses in $An(\*W_\star)$, that are not present in $An(\*W'_\star)$. And conversely, we need to show that each pair of conflicting potential responses in $An(\*W_\star)$ involving $X$ (i.e., that is not present in $An(\*W'_\star)$) corresponds to at least one conflict detected by \textbf{CTF-REALIZE}($P(\*W_\star),\mathcal{G},\mathbb{A}^\dag(\mathcal{G}))$ in the outer loop involving $X$.

\underline{Forward direction}:

As discussed in the proof of Theorem \ref{thm:completeness}, \textbf{CTF-REALIZE}($P(\*W_\star),\mathcal{G},\mathbb{A}^\dag(\mathcal{G}))$ returns FAIL in the outer loop involving $X$ if and only if the input of $X$ to some $C \in Ch(X)$ is required to be some $x$ to satisfy condition [i] w.r.t some $W_\*t \in \*W_\star$, but also required to be $x'$ or "natural" to satisfy condition [i/ii] w.r.t some $Y_\*h \in \*W_\star$. 

Note that the action set is $\mathbb{A}^\dag(\mathcal{G})$. Therefore step 6 of sub-routine Algo. \ref{alg:subroutine} would always pick only the procedure $\textsc{ctf-Rand}(X \rightarrow C)$ whenever $C$ needs to receive a fixed value. The interventions affecting other $C' \in Ch(X)$ would not affect $C$. 

In this case, it is easy to see that the set $An(W_\*t)$ must contain $C_{x...}$ per Def. \ref{def:ctf_ancestor}, and $An(Y_\*h)$ must contain $C_{x'...}$ or a potential response of $C$ without $X$ in the subscript. Thus, if \textbf{CTF-REALIZE}($P(\*W_\star),\mathcal{G},\mathbb{A}^\dag(\mathcal{G}))$ returns FAIL in the outer loop involving $X$, there must be a pair of conflicting potential responses in $An(\*W_\star)$.

\underline{Reverse direction}:

Assume there exists a conflicting pair of potential responses $A_{\*t}$, $A_{\*s} \in An(\*W_\star)$ where $x \in \*t$ and $\*s$ contains some $x'$ or does not contain $X$ at all. 

This means there is some $W_\*h \in \*W_\star$ s.t. $A \in An(W)_{\mathcal{G}_{\underline{X}}}$ and some $Y_\*j \in \*W_\star$ s.t. $A \in An(Y)_{\mathcal{G}_{\underline{X}}}$. I.e., $A$ mediates the effect of $X$ on $W,Y$. Further, from Def. \ref{def:ctf_ancestor}, it means that $A$ needs to be realized in conflicting regimes w.r.t $X$. 

From Lemma \ref{lem:realizable_proofs_1}, such conflict happens because for $A_{\*t}$, condition [i] requires that $X$ is fixed by intervention to be $x$ for all children $C \in Ch(X) \cap An(A)$. Whereas, for $A_{\*s}$, condition [i] or [ii] requires that each child $C \in Ch(X) \cap An(A)$ receives $X$ either fixed as $x'$, or naturally (as the case may be, for $\*s$). For any such $C \in Ch(X) \cap An(A)$, it is clear from the proof of Theorem $\ref{thm:completeness}$ that this conflict will trigger a FAIL from \textbf{CTF-REALIZE}($P(\*W_\star),\mathcal{G},\mathbb{A}^\dag(\mathcal{G}))$ in the first outer loop involving $X$.

\medskip
Thus, we have shown that the IH holds for any $\*W_\star$ involving a graph with $n+1$ nodes.

Since Theorem \ref{thm:completeness} shows \textbf{CTF-REALIZE} is complete, we have thus proved that $\*W_\star$ is realizable given $\mathcal{G}$ and $\mathbb{A}^\dag(\mathcal{G})$ if and only if $An(\*W_\star)$ does not contain a pair of conflicting potential responses for the same variable under different regimes.

$\hfill \blacksquare$

\bigskip
\newpage
\paragraph{Corollary \ref{cor:fpci}}

:

This follows from Corollary \ref{cor:maximal_amwn}. For any causal diagram, the ancestral set of $\{Y_x,Y_{x'}\}$ would include both these potential responses.

Thus, the query is not realizable.

$\hfill \blacksquare$

\bigskip
\bigskip
\subsection{Realizability of $\mathcal{L}_1$- and $\mathcal{L}_2$-distributions}
\label{app:l1l2_realizability}

It is widely known and acknowledged that it is possible to draw samples from $\mathcal{L}_1$- and $\mathcal{L}_2$-distributions: the former by simply observing a system's natural behaviour, and the latter by intervening in the system through interventions like Fisherian randomization.

Still, we find it educational to derive these proofs from first principles. This sub-section is not strictly needed to follow the main contributions in Secs. \ref{sec:data_collection_procedures} and \ref{sec:realizability_main_sec}.

We define the probability measure $P^{\mathbb{C}}(.)$ from the perspective of an exogenous agent (i.e., an agent external to the system) $\mathbb{C}$'s beliefs about the environment, distinguished by superscript from $P^{\mathcal{M}}(.)$, the true unknown distribution. 

Since unit selection is randomized, $\textsc{Select}^{(i)}$ yields an unbiased sample of a unit with latent features distributed according to the target population frequency $P(\*u)$. I.e., $P^{\mathbb{C}}(\*U^{(i)} = \*u \mid \textsc{Select}^{(i)}) = P^{\mathcal{M}}(\*u)$. We also assume that target population size is large enough that $\textsc{Select}^{(i)}$ does not significantly change the distribution of the remaining population.

Further, we assume that the actions $\textsc{Read}(V)^{(i)}$ and $\textsc{Rand}(V)^{(i)}$ do not disrupt any other mechanism $f_{V'}$ for unit $i$.

\bigskip
\begin{lemma}[Observational sample]
\label{lem:l1_pv}
An agent $\mathbb{C}$ can draw an i.i.d sample distributed according to the $\mathcal{L}_1$ query $P(\*V)$ associated with an SCM $\mathcal{M}$, by the following actions:

i. $\textsc{Select}^{(i)}$

ii. $\textsc{Read}(\*V)^{(i)} = \*v \sim P(\*V)$

Given $N$ i.i.d samples, the consistent unbiased estimate of $P(\*v)$ is
\begin{align}
    \label{eq:obs_pv} 
    \hat{P}(\*v) 
    := \frac{1}{N}\sum\limits_{i} \prod\limits_{v \in \*v}\mathbbm{1}[\textsc{Read}(V)^{(i)}=v]
\end{align}
\end{lemma}

\begin{proof}
    This follows directly from the definitions of the actions. $\textsc{Select}^{(i)}$ chooses a unit at random from the population. By Remark \ref{rem:exogeneity}, $P^{\mathbb{C}}(\*U^{(i)} = \*u \mid \textsc{Select}^{(i)}) = P^{\mathcal{M}}(\*u)$. For randomly selected unit $i$,
\begin{align}
    &P^{\mathbb{C}}(\textsc{Read}(\*V)^{(i)} = \*v \mid \textsc{Select}^{(i)})\\
    &= \sum_{\*u} P^{\mathbb{C}}(\*U^{(i)} = \*u \mid \textsc{Select}^{(i)}).\\ \nonumber
    &P^{\mathbb{C}}(\textsc{Read}(\*V)^{(i)} = \*v \mid \*U^{(i)} = \*u, \textsc{Select}^{(i)}) &\text{Chain rule}\\ 
    &= \sum_{\*u} P^{\mathbb{C}}(\*U^{(i)} = \*u \mid \textsc{Select}^{(i)}).\mathbbm{1}^{\mathcal{M}}[\*V(\*u) = \*v] &\text{Def. \ref{def:physical_actions}(ii)}\\ 
    &= \sum_{\*u} P^{\mathcal{M}}(\*u).\mathbbm{1}^{\mathcal{M}}[\*V(\*u) = \*v] &\text{Rem. \ref{rem:exogeneity}}\\
    &= P^{\mathcal{M}}(\*v) & \text{Definition}
\end{align}
I.e., this record is an i.i.d. sample from $P^{\mathcal{M}}(\*V)$. Now consider the estimator below.
\begin{align}
    \hat{P}(\*v) &:= \frac{1}{N}\sum\limits_{n} \prod\limits_{v \in \*v}\mathbbm{1}^{\mathbb{C}}[\textsc{Read}(V)^{(i)}=v]\\
    &= \frac{1}{N}\sum\limits_{n} \sum_{\*u} \prod\limits_{v \in \*v}\mathbbm{1}^{\mathcal{M}}[\*U^{(i)} = \*u].\mathbbm{1}^{\mathcal{M}}[V(\*u) = v]
\end{align}
Un-biasedness is established by taking expectation on either side, w.r.t the agent $\mathbb{C}$'s actions (choice of units to observe):
\begin{align}
    \mathbb{E}_{\mathbb{C}}[\hat{P}(\*v)] &= \mathbb{E}_{\mathbb{C}}\bigg[\frac{1}{N}\sum\limits_{n} \sum_{\*u} \prod\limits_{v \in \*v}\mathbbm{1}^{\mathcal{M}}[\*U^{(i)} = \*u].\mathbbm{1}^{\mathcal{M}}[V(\*u) = v] \bigg]\\
    &=  \sum_{\*u} \frac{1}{N}\mathbb{E}_{\mathbb{C}}\bigg[\sum\limits_{n} \mathbbm{1}^{\mathcal{M}}[\*U^{(i)} = \*u] \prod\limits_{v \in \*v}.\mathbbm{1}^{\mathcal{M}}[V(\*u) = v] \bigg] &\text{Linearity of expectation}\\
    &=  \sum_{\*u} \frac{1}{N} \mathbb{E}_{\mathbb{C}}\bigg[\sum\limits_{n} \mathbbm{1}^{\mathcal{M}}[\*U^{(i)} = \*u]\bigg] \prod\limits_{v \in \*v}\mathbbm{1}^{\mathcal{M}}[V(\*u) = v] &\text{$V(\*u)$ constant wrt $\mathbb{C}$}\\
    &=  \sum_{\*u} \frac{1}{N} \bigg[N.P^{\mathcal{M}}(\*u)\bigg] \prod\limits_{v \in \*v}\mathbb{I}^{\mathcal{M}}[V(\*u) = v] &\text{Def. \ref{def:physical_actions}(i), Rem. \ref{rem:exogeneity}}\\
    &=  P^{\mathcal{M}}(\*v) &\text{Definition}
\end{align}

Consistency is established by the fact that as $\mathcal{N}$ (target population size) $\rightarrow \infty$, and $N$ (sample size) $\rightarrow \infty$,
\begin{align}
    \frac{1}{N}\sum\limits_{n}  \mathbb{I}^{\mathcal{M}}[\*U^{(i)} = \*u] \rightarrow P^{\mathcal{M}}(\*u)
\end{align}
\end{proof}

\bigskip
\begin{lemma}\label{lemma:do-sigma}
The $\mathcal{L}_2$ distribution of an atomic intervention is equivalent to the $\mathcal{L}_2$ distribution of the corresponding conditional stochastic intervention.
\begin{align}
    &P^{\mathcal{M}}(\*v;do(\*x)) = P^{\mathcal{M}}(\*v|\*x;\sigma_{\*X}) \label{eq:do-sigma}\\
    &= \sum\limits_{\*u} \underbrace{\mathbbm{1}[\*V_{\sigma_{\*X}} (\*u) = \*v  \mid X_{\sigma_{\*X}}=\*x]}_{\text{\textcircled{1}}}.\underbrace{P(\*u)}_{\text{\textcircled{2}}} \label{eq:l2_pv}
\end{align}
\end{lemma}
\begin{proof}
    The step from the r.h.s of Eq. \ref{eq:do-sigma} to Eq. \ref{eq:l2_pv} is derived  as follows: in the submodel $\mathcal{M}_{\sigma_{\*X}}$, if we are given that $\*X$ has been randomly assigned $\*x$, then the remaining variables are deterministically generated as a function of $\*u$ and $\*x$ via their respective equations. The probability mass is collected over all the $\*u$ which produce the output $\*v$ over all these equations.
\begin{align}
    P^{\mathcal{M}}(\*v|\*x;\sigma_{\*X}) = \sum\limits_{\*u} \mathbb{I}[\*V_{\sigma_{\*X}} (\*u) = \*v  \mid X_{\sigma_{\*X}}=\*x].P^{\mathcal{M}}(\*u)
\end{align}
Notice: if $\*v$ is incompatible with $\*x$, the indicator in the r.h.s evaluates to 0. Next, we prove. Eq. \ref{eq:do-sigma}.

In $\mathcal{M}_{\sigma{\*X}}$, as defined, $\*X$ is assigned according to an independent random vector. Notate this vector as $\*X_{\sigma_{\*X}}$ and let the distribution of this vector be $P_{\sigma_{\*X}}(\*X)$, defined by the assignment frequency over the target population.

$\mathcal{M}_{\sigma{\*X}}$ is defined such that the target population is split into groups, each assigned $(X_{\sigma_{\*X}} = \*x)$ for some $\*x$. Note, the assignment vector $\*X_{\sigma_{\*X}}$ is independent of the latent features $\*U$ across the target population iff each finite group assigned $(X_{\sigma_{\*X}} = \*x)$ has the same distribution of latent features $P(\*U)$ as in the overall target population.

The above discussion handles the finite size of the target population. Starting with the r.h.s of Eq. \ref{eq:do-sigma},
\begin{align}
    P^{\mathcal{M}}(\*v|\*x;\sigma_{\*X}) = \frac{P(\*v,\*x;\sigma_{\*X})}{P(\*x;\sigma_{\*X})} = \begin{cases} P(\*v;\sigma_{\*X})/P(\*x;\sigma_{\*X}) & \mbox{if $\*v$ compatible with $\*x$} \\ 0 & \mbox{otherwise} \end{cases} \label{eq:do-sigma-cases}
\end{align}
Evaluating for when $\*v$ is compatible with $\*x$:
\begin{align}
    \frac{P(\*v;\sigma_{\*X})}{P(\*x;\sigma_{\*X})} &= \frac{P(\*v;\sigma_{\*X})}{P_{\sigma_{\*X}}(\*x)}\\
    &= \frac{\sum_{\*u}  \bigg( P(\*u) \prod_{V_i \in \*V \setminus \*X }  P(v_i \mid \*{pa}_i,\*{u}_i).P_{\sigma_{\*X}}(\*x)\bigg)}{P_{\sigma_{\*X}}(\*x)} & \text{Truncated factorization product} \label{eq:trunc1}\\
    &= \sum_{\*u} P(\*u) \prod_{V_i \in \*V \setminus \*X }  P(v_i \mid \*{pa}_i,\*{u}_i)\\
    &= P^{\mathcal{M}}(\*v ; \doo{\*x}) & \text{Truncated factorization product}
\end{align}

Eq. \ref{eq:trunc1} uses the fact that each sub-group assigned $(X_{\sigma_{\*X}} = \*x)$, by independence, has the same frequency of latent features $P(\*u)$.
\end{proof}

\bigskip
\begin{lemma}[Interventional sample]
\label{lem:l2_pv}
An agent $\mathbb{C}$ can draw an i.i.d sample distributed according to the $\mathcal{L}_2$ query $P(\*V; do(\*x))$ associated with an SCM $\mathcal{M}$, by the following actions:

i. $\textsc{Select}^{(i)}$

ii. $\textsc{Rand}(\*X)^{(i)}$

iii. If $\textsc{Rand}(\*X)^{(i)} = \*x$, then $\textsc{Read}(\*V)^{(i)} = \*v \sim P(\*V;\doo{\*x})$, else repeat i-iii.

Given $N_{\*x}$ i.i.d samples, the consistent unbiased estimate of Eq. \ref{eq:l2_pv} is given by
\begin{align}
    \label{eq:exp_pv} 
    &\hat{P}(\*v;do(\*x)) =\nonumber\\
    &\underbrace{\frac{1}{N_{\*x}}\sum_{i}}_{\text{\textcircled{2}}} \underbrace{\mathbbm{1}[\textsc{Read}(\*V)^{(i)}=\*v, \textsc{Rand}(\*X)^{(i)}=\*x]}_{\text{\textcircled{1}}},
\end{align}
\end{lemma}
\begin{proof}
    The proof steps are similar to the ones used for the Observational i.i.d sample case. Note that Remark \ref{rem:exogeneity} still hold since even though the agent is conditioning on the value randomly assigned to a particular unit $i$, this value is independent of the unit's latent features $\*U^{(i)}$. 
\end{proof}
\newpage
\section{Details on counterfactual randomization}
\label{app:l3_conditions}

In this appendix, we provide a formal account for the procedure we define in Sec. \ref{sec:data_collection_procedures}, called $\textsc{ctf-Rand}(X \rightarrow \*C)$ (Def. \ref{def:ctf-rand}). This appendix is intended for the interested reader, and details some useful remarks to be used in the proofs of our results in Appendix \ref{app:assumptions_and_proofs}. These details are not strictly needed to follow the results presented in the main body of the paper.

\begin{itemize}
    \item In Sec. \ref{app:ctf_rand_conditions}, we lay out the structural conditions under which it is possible to perform this procedure, and we provide an algorithm (Algorithm \ref{alg:ctf_procedures}) by which an agent can translate the structural conditions in the environment into a list of $\textsc{ctf-Rand}$ procedures that it is able to perform in the given setting.
    \item In Sec. \ref{app:multiple_rands}, we emphasize that it is possible for an agent to enact multiple randomization procedures involving the same variable $X$ for a single unit $i$, and illustrate this with an example.
    \item In Sec. \ref{app:ctf_rand_constraints}, we discuss the constraints implied by the assumptions we make. In particular, we discuss why $\textsc{ctf-Rand}(X \rightarrow \*C)$ can only be performed on some $\*C \subseteq Ch(X)$, and not by-pass children to directly affect some distant descendants of $X$.
\end{itemize}

\subsection{Structural conditions required for counterfactual randomization}
\label{app:ctf_rand_conditions}

Counterfactual randomization (Def. \ref{def:ctf-rand}) can be performed under two circumstances:
\begin{itemize}
    \item [i.] $\textsc{ctf-Rand}(X \rightarrow Ch(X))$ can be performed by eliciting a unit's natural decision $X$, while simultaneously randomizing its actual enforced decision. Thus, the agent can affect the value of the decision $X$ as received by all the children of $X$, whilst also recording the natural realization of $X$. As discussed in Sec. \ref{sec:data_collection_procedures}, this was established in \citep{bareinboim:etal15, forney:etal17, zhang22a}.
    \item [ii.] $\textsc{ctf-Rand}(X \rightarrow \*C)$ can also be performed for some $\*C \subseteq Ch(X)$ if there is a special \textit{counterfactual mediator} (defined below) by which the mechanisms generating $\*C$ perceive the value of $X$. This counterfactual mediator then allows the agent to intervene on the value of $X$ as \textit{perceived} by $\*C$, thus mimicking an actual intervention on $X$.
\end{itemize}

\medskip
\begin{definition}[Expanded SCM] \label{def:expanded_SCM}
    Given an SCM $\mathcal{M}$ containing observable variables $\*V$, we define an \textit{expanded SCM} $\mathcal{M}^+$ of the same environment to be a model containing a bigger set of observable variables $\*V^+ \supset \*V$, and which relaxes the positivity requirement. I.e., it is possible that $P^{\mathcal{M}^+}(\*v^+)=0,$ for some $\*v^+$ in $\mathcal{L}_1$. We call the causal diagram of $\mathcal{M}^+$ an \textit{expanded} causal diagram $\mathcal{G}^+$.  $\hfill \blacksquare$    
\end{definition}

\medskip
\begin{definition}[Counterfactual mediator (formal)] \label{def:l2_proxy}
    Given a variable $X$ in a causal diagram $\mathcal{G}$, we call any variable $W \not \in \*V$ a \textit{counterfactual mediator} of $X$ w.r.t $Y \in Ch(X)_{\mathcal{G}}$ if

    i. In an "expanded" SCM of the environment $\mathcal{M}^+$ (Def. \ref{def:expanded_SCM}), $W$ is generated according to an invertible mechanism $W \gets f_W(X, \*U_W)$ with $\*U_W$ possibly empty, s.t. $f_W^{-1}(W) = X$; 
    
    ii. It is physically possible to perform $\textsc{Rand}(W)^{(i)}$ (Def. \ref{def:physical_actions}); and
    
    iii. In $\mathcal{M}^+$, $Y$ is generated by the mechanism $Y \gets f_Y(f_W^{-1}(W), \*A, \*U_Y)$, where $\*A$ is the set $\*{Pa}_Y \setminus X$ in $\mathcal{G}$. $\hfill \blacksquare$
\end{definition}

The intuition behind Def. \ref{def:l2_proxy} is that a \textit{counterfactual mediator} is a real variable in the environment which fully encodes information about the variable $X$, and which mediates how $Y$ perceives the value of $X$ via the "direct" causal path. For instance, in Example 1 (\textit{Mediation analysis}), the RGB values of the video frames $W$ are a counterfactual mediator for the mechanism $f_Y$ (decision to issue a speeding ticket) to perceive the car's color $X$ via the "direct" path, not via the actual speeding of the car). 

\textbf{Condition [i]} of Def. \ref{def:l2_proxy} divides the domain of $W$ into \textit{equivalence classes} s.t. each value $w$ belongs to an equivalence class $\{w': f_W^{-1}(w)= x\}$ for some value $x$. 

\textbf{Condition [iii]} of Def. \ref{def:l2_proxy} essentially says that the mechanism $f_Y$ only cares about which equivalence class $W$ belongs to. I.e., $Y$ only cares about what $W$ reveals about $X$. 

Note: these conditions does not require an agent to have full knowledge of the SCM. They are rather structural assumptions about the underlying mechanisms which can be verified in a given setting. In Example 1, treating $W$ as counterfactual mediator means the assumption that (1) the video features $W$ uniquely map back to the actual color of the car in the footage; and (2) the computer vision system only cares $W$ reveals about $X$, and is indifferent to any stochasticity \textit{within} some equivalence class $\{w': f_W^{-1}(w)= x\}$.

\medskip
\begin{assumption}(Tree structure)\label{assn:proxy_tree}
    Given a variable $X$, causal diagram $\mathcal{G}$, and an "expanded" diagram $\mathcal{G}^+$ (Def. \ref{def:expanded_SCM}) including the set of all the counterfactual mediators $\*W$ (Def. \ref{def:l2_proxy}) of $X$ in the environment, each $W \in \*W$ has only one parent in $\mathcal{G}^+$, and each $C \in Ch(X)_{\mathcal{G}}$ has at most one $W \in \*W$ as a parent in $\mathcal{G}^+$. $\hfill \blacksquare$
\end{assumption}

\begin{figure}[t]
    \centering
    \begin{subfigure}[b]{0.24\textwidth}
        \centering
        \begin{tikzpicture}
        \node (X) at (0,0) {$X$};
        \node[orange] (W1) at (-0.5,-0.75) {$W_1$};
        \node[orange] (W2) at (0.5,-0.75) {$W_2$};
        \node (Y) at (-0.75,-1.5) {$Y$};
        \node (Z) at (0,-1.5) {$Z$};
        \node (T) at (0.75,-1.5) {$T$};
        \path [->] (X) edge (W1);
        \path [->] (X) edge (W2);
        \path [->] (W1) edge (Y);
        \path [->] (W1) edge (Z);
        \path [->] (W2) edge (Z);
        \path [->] (W2) edge (T);
    \end{tikzpicture}
        \caption{}
    \end{subfigure}
    \begin{subfigure}[b]{0.24\textwidth}
        \centering
        \begin{tikzpicture}
        \node (T) at (0,0.75) {$T$};
        \node (X) at (0,0) {$X$};
        \node (Y) at (-0.75,-0.75) {$Y$};
        \node (Z) at (0.75,-0.75) {$Z$};
        \path [->] (T) edge (X);
        \path [->] (X) edge (Y);
        \path [->] (X) edge (Z);
        \path [<->,dashed] (T) edge [bend right=30] (Y);
        \path [<->,dashed] (T) edge [bend left=30] (Z);
        \end{tikzpicture}
        \caption{}
    \end{subfigure}
    \begin{subfigure}[b]{0.24\textwidth}
        \centering
        \begin{tikzpicture}
        \node (X) at (0,0) {$X$};
        \node[orange] (W1) at (-0.5,-0.75) {$W_1$};
        \node[orange] (W2) at (0.5,-0.75) {$W_2$};
        \node (Y) at (-0.75,-1.5) {$Y$};
        \node (Z) at (0,-1.5) {$Z$};
        \node (T) at (0.75,-1.5) {$T$};
        \path [->] (X) edge (W1);
        \path [->] (X) edge (W2);
        \path [->] (W1) edge (Y);
        \path [->] (W2) edge (Z);
        \path [->] (W2) edge (T);
        \path [->] (Z) edge (Y);
    \end{tikzpicture}
        \caption{}
    \end{subfigure}
    \begin{subfigure}[b]{0.24\textwidth}
        \centering
        \begin{tikzpicture}
        \node (X) at (0,0.75) {$X$};
        \node[orange] (W1) at (0,0) {$W_1$};
        \node[orange] (W2) at (0.5,-0.75) {$W_2$};
        \node (Y) at (-0.75,-1.5) {$Y$};
        \node (Z) at (0,-1.5) {$Z$};
        \node (T) at (0.75,-1.5) {$T$};
        \path [->] (X) edge (W1);
        \path [->] (W1) edge (Y);
        \path [->] (W1) edge (W2);
        \path [->] (W2) edge (Z);
        \path [->] (W2) edge (T);
        \path [->] (Y) edge (Z);
        \path [->] (T) edge (Z);
    \end{tikzpicture}
        \caption{}
    \end{subfigure}
    \caption{"Expanded" causal diagrams, where counterfactual mediators (labeled $W_i$) of $X$ have been marked in red. (a) is not permitted by Assumption \ref{assn:proxy_tree}. Assume the environment in diagram (b) permits the agent to elicit the unit's natural decision $X$ even when randomizing the actual decision.}
    \label{fig:proxies_to_ctfrand}
    \vskip -0.1in
\end{figure}
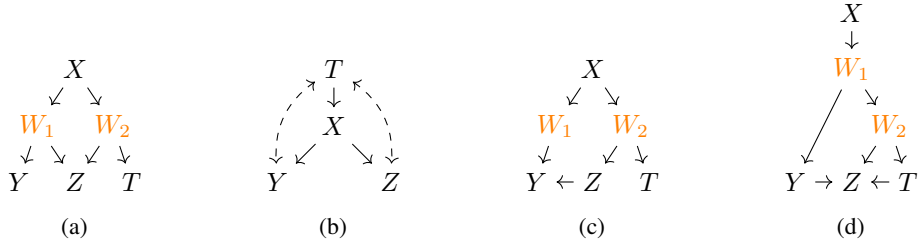

Assumption \ref{assn:proxy_tree} enforces that each child of $X$ perceives $X$ through at most one proxy pathway. This assumption rules out possible structures like Fig. \ref{fig:proxies_to_ctfrand}(a) where a child perceives $X$ through multiple proxy pathways. 

This assumption is general enough to allow most cases of interest. If $X$ is a construct like gender identity, then it is possible that a child perceives $X$ via a cluster of personal attributes $\*W$ which indicate $X$. In this case, no single attribute solely satisfies Def. \ref{def:l2_proxy} of a counterfactual mediator. However, the cluster of attributes $\*W$ could be collapsed into a single variable having domain equal to the cartesian product of the sub-domains \citep{anandetal23,Xia_Bareinboim_2024}. This single node $\*W$ would indeed satisfy the definition of a counterfactual mediator and would comply with the tree structure in Assumption \ref{assn:proxy_tree}. For a comprehensive discussion of the semantics of interventions on the perception of a compound attribute such as race or gender identity, see \citep[~App. D.1]{plecko:bareinboim24}.
The following Lemma is the key property that enables path-specific randomization.

\medskip
\begin{lemma} \label{lem:l2_proxy}
    Given a causal diagram $\mathcal{G}$ containing variables $X$ and $Y \in Ch(X)_{\mathcal{G}}$. Let $W$ be a counterfactual mediator of $X$ w.r.t $Y$ (Def. \ref{def:l2_proxy}). For any value $x$, we have
    \begin{align}
        Y_{w\*a}(\*u) &= Y_{x\*a}(\*u), &\forall \*u, \forall w \in \{w': f_W^{-1}(w)= x\},\label{eq:l2_proxy}
    \end{align}
    where $\*A := \*{Pa}_Y \setminus X$ in  $\mathcal{G}$. 
\end{lemma}
\begin{proof}
    This follows from Def. \ref{def:l2_proxy}. Suppose we are given values $(w,x)$ where $f^{-1}_W(w) = x$. Let $\*A:= \*{Pa}_Y \setminus X$ in $\mathcal{G}$.

The variable $W_x(\*u) = W_{x\*a}(\*u) = f_W(x, \*u)$, in the enhanced submodel $\mathcal{M}_{x\*a}^+$. Adding $\*a$ to the subscript does not matter - by Assumption  \ref{assn:proxy_tree} and Lemma \ref{lem:no_forks}, $\*A$ cannot be an ancestor of $W$ in $\mathcal{M}^+$.

Since $f_W$ is invertible by condition [i] in $\mathcal{M}^+$, it is also invertible in submodel submodel $\mathcal{M}_{x\*a}^+$. Therefore, we have $f^{-1}_W(W_{x\*a}(\*u)) = x$.
\begin{align*}
    Y_{w\*a}(\*u) &= f_Y(f^{-1}_W(w), a, \*u) \\
    &= f_Y(x, a, \*u)\\
    Y_{W_{x\*a} \*a}(\*u) &= f_Y(f^{-1}_W(W_{x\*a}(\*u)), a, \*u)\\
    &= f_Y(x, a, \*u)
\end{align*}
The r.h.s is identical, giving us $Y_{w\*a} = Y_{W_{x\*a} \*a}$. Finally, we argue that $Y_{W_{x\*a} \*a} = Y_{x\*a}$.

The counterfactual $Y_{x\*a}$ is evaluated in a submodel of $\mathcal{M}^+$, where $f_W$ receives input $x$ and this value of $W_x$ is an input to $f_Y$, while $\*A$ is fixed to be $\*a$. Structurally, this is identical to how the counterfactual $Y_{W_{x\*a} \*a} = Y_{W_{x} \*a}$ is evaluated. Therefore, it is evident that $Y_{W_x \*a} = Y_{x\*a}$. 
\end{proof}

Given a variable $X$, the way an agent actually performs the action $\textsc{ctf-Rand}$ is as follows:
\begin{itemize}
    \item[i.] \textbf{Performing $\textsc{ctf-Rand}$ by eliciting natural decision:} The agent can perform $\textsc{ctf-Rand}(X \rightarrow Ch(X))^{(i)}$ by randomizing the unit's decision. The agent can further perform $\textsc{Read}(X)^{(i)}$ to elicit the unit's natural decision, which has not been erased. This is described in in \citep{bareinboim:etal15,forney:etal17,zhang22a}.
    \item[ii.] \textbf{Performing $\textsc{ctf-Rand}$ using counterfactual mediators:} If $X$ has a counterfactual mediator $W$ in the environment, and $\*C \subseteq Ch(X)$ are the children which perceive $X$ via $W$, then the agent can perform $\textsc{ctf-Rand}(X \rightarrow \*C)^{(i)}$ by randomizing $W$. Each value $w$ mimics randomizing $X$ as perceived by $\*C$, per Lemma \ref{lem:l2_proxy}. The agent can still perform $\textsc{Read}(X)^{(i)}$ by measuring $X^{(i)}$ to get the unit's natural decision, which has not been erased.
\end{itemize}

Having described the structural conditions that permit counterfactual randomization, we want to abstract away the mediators and succinctly describe the agent's physical actions via the definition of $\textsc{ctf-Rand}$. Given a variable $X$, and assumptions/knowledge about $X$ in the environment stated in points \textbf{[i]} and \textbf{[ii]} above, we translate this knowledge into a set of counterfactual randomizations that the agent is physically able to perform in the environment, using Algorithm \ref{alg:ctf_procedures}.

\begin{algorithm}[h]
   \caption{CTF-PROCEDURES}
   \label{alg:ctf_procedures}
\begin{algorithmic}[1]
   \STATE {\bfseries Input:} Causal diagram $\mathcal{G}$ with decision variable $X$; "expanded" diagram $\mathcal{G}^+$ (Def. \ref{def:expanded_SCM}) including the counterfactual mediators of $X$ in the environment
   \STATE {\bfseries Output:} $\mathbb{A}_X$ - the set of $\textsc{ctf-Rand}$ actions that can be performed involving $X$

   \smallskip
   \STATE $\mathbb{A}_X \gets \emptyset$

   \smallskip
   \IF{environment allows eliciting natural decision $X$ even when randomizing actual decision}
        \IF{$X$ can be randomized}
        \STATE $\mathbb{A}_X \gets \mathbb{A}_X \cup \{\textsc{ctf-Rand}(X \rightarrow Ch(X)_{\mathcal{G}})\}$
        \ENDIF
   \ENDIF
   
   \smallskip
   \FOR{each counterfactual mediator $W$ of $X$}
    \STATE Let $\*C := \{C \mid C \in Ch(X)_{\mathcal{G}} \text{ and perceives $X$ via $W$}\}$ 
    \STATE $\mathbb{A}_X \gets \mathbb{A}_X \cup \{\textsc{ctf-Rand}(X \rightarrow \*C)\}$
   \ENDFOR

   \STATE Return $\mathbb{A}_X$
\end{algorithmic}
\end{algorithm}

Consider Fig. \ref{fig:proxies_to_ctfrand}(a-d). (a) is not permitted by Assumption \ref{assn:proxy_tree}.  We assume that in the environment represented by (b) $X$ can be randomized for a unit in the target population without erasing the unit's natural decision, satisfying condition \textbf{[i]} mentioned earlier. Thus, when applying Algorithm \ref{alg:ctf_procedures} to diagrams (b-d), we get the following resulting set of counterfactual randomization procedures which are permitted by the structural assumptions made (unit superscript $i$ is omitted for legibility):
\begin{itemize}
    \item [(b)] $\{ \textsc{ctf-Rand}(X \rightarrow \{Y,Z\}) \}$
    \item [(c)] $\{ \textsc{ctf-Rand}(X \rightarrow Y),  \textsc{ctf-Rand}(X \rightarrow \{Z,T\})\}$
    \item [(d)] $\{ \textsc{ctf-Rand}(X \rightarrow \{Y,Z,T\}, \textsc{ctf-Rand}(X \rightarrow \{Z,T\}) \}$
\end{itemize}


\subsection{Multiple simultaneous randomizations are possible, for a single unit}
\label{app:multiple_rands}

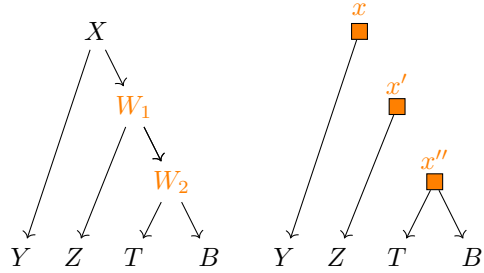
\begin{wrapfigure}{r}{0.5\textwidth}
    \vspace{-0.3in}
    \begin{center}
    \begin{tikzpicture}

        \node (X) at (0,0) {$X$};
        \node[orange] (W1) at (0.5,-1) {$W_1$};
        \node[orange] (W2) at (1,-2) {$W_2$};
        \node (Y) at (-1,-3) {$Y$};
        \node (Z) at (-0.3,-3) {$Z$};
        \node (T) at (0.5,-3) {$T$};
        \node (B) at (1.5,-3) {$B$};
        \path [->] (X) edge (Y);
        \path [->] (X) edge (W1);
        \path [->] (W1) edge (Z);
        \path [->] (W1) edge (W2);
        \path [->] (W2) edge (T);
        \path [->] (W2) edge (B);

        \node[fill=orange,draw,inner sep=0.3em, minimum width=0.3em] (intervention1) at (3.5,0) {\ };
        \node[orange] (x2) at (3.5,0.3) {$x$};
        \node[fill=orange,draw,inner sep=0.3em, minimum width=0.3em] (intervention2) at (4,-1) {\ };
        \node[orange] (x2) at (4,-0.7) {$x'$};
        \node[fill=orange,draw,inner sep=0.3em, minimum width=0.3em] (intervention3) at (4.5,-2) {\ };
        \node[orange] (x2) at (4.5,-1.7) {$x''$};
        \node (Y) at (2.5,-3) {$Y$};
        \node (Z) at (3.2,-3) {$Z$};
        \node (T) at (4,-3) {$T$};
        \node (B) at (5,-3) {$B$};
        \path [->] (intervention1) edge (Y);
        \path [->] (intervention2) edge (Z);
        \path [->] (W1) edge (W2);
        \path [->] (intervention3) edge (T);
        \path [->] (intervention3) edge (B);
        
    \end{tikzpicture}
    \end{center}
    \vskip -0.in
    \caption{(Left) "Expanded" causal diagram showing counterfactual mediators $W_1, W_2$ of $X$; (Right) Agent performing actions $\textsc{Rand}(X)^{(i)}, \textsc{ctf-Rand}(X \rightarrow \{Z,T,B\})^{(i)}$ and $\textsc{ctf-Rand}(X \rightarrow \{T,B\})^{(i)}$ all together on the single unit $i$.}
    \label{fig:supersede_rand}
    \vskip -0.4in
    \end{wrapfigure}

For a particular decision variable $X$, there could be multiple randomization procedures which an agent can perform. Consider the example in Fig. \ref{fig:supersede_rand}. The "expanded" diagram on the left shows two counterfactual mediators, $W_1, W_2$ in a causal structure which permit an agent to perform all of the following randomization procedures: $\textsc{Rand}(X)^{(i)}, \textsc{ctf-Rand}(X \rightarrow \{Z,T,B\})^{(i)}$ and $\textsc{ctf-Rand}(X \rightarrow \{T,B\})^{(i)}$ for the same unit $i$. 

However, if all three actions are performed in parallel, randomizing $W_1$ to enact $\textsc{ctf-Rand}(X \rightarrow \{Z,T,B\})^{(i)}$ will only affect variable $Z$. This is since the action of randomizing $W_2$ to further enact $\textsc{ctf-Rand}(X \rightarrow \{T,B\})^{(i)}$ blocks any effect on $T,B$ from the previous action. Similarly, $\textsc{Rand}(X)^{(i)}$ ends up affecting only variable $Y$, because $\textsc{ctf-Rand}(X \rightarrow \{Z,T,B\})^{(i)}$ blocks any effect from the previous action on $Z,T,B$. We formalize this observation in Remark \ref{remark:supersede}.

\medskip
\begin{remark}[Superseding action]
 \label{remark:supersede}
    Given a decision variable $X$, the action $\textsc{ctf-Rand}(X \rightarrow \*C')^{(i)}$ can \textit{supersede} the action $\textsc{ctf-Rand}(X \rightarrow \*C)^{(i)}$ if $\*C' \subsetneq \*C$, where \textit{supersede} means that the former action $\textsc{ctf-Rand}(X \rightarrow \*C')^{(i)}$ blocks any effect that the latter action has on the variables $\*C'$. Additionally, the action $\textsc{ctf-Rand}(X \rightarrow \*C)^{(i)}$ \textit{supersedes} the action $\textsc{Rand}(X)^{(i)}$. $\hfill \blacksquare$
\end{remark}

Further, Assumption \ref{assn:proxy_tree} ensures that all such procedures follow a "nested" structure. I.e., given any two randomization procedures involving the same variable, the sets of children affected by one will be a subset of the set affected by the other, as shown in Fig. \ref{fig:supersede_rand}. 
\medskip
\begin{remark}(Procedure containment) \label{rem:containment}
    Assumption \ref{assn:proxy_tree} implies that if an agent is capable of performing both $\textsc{ctf-Rand}(X \rightarrow \*C)^{(i)}$ and $\textsc{ctf-Rand}(X \rightarrow \*C')^{(i)}$ s.t. $\*C \neq \*C'$ and $\*C \cap \*C' \neq \emptyset$, then either $\*C \subseteq \*C'$ or $\*C' \subseteq \*C$. $\hfill \blacksquare$ 
\end{remark}


\subsection{Counterfactual randomization is only possible for direct children of $X$}
\label{app:ctf_rand_constraints}

Our definition of $\textsc{ctf-Rand}(X \rightarrow \*C)^{(i)}$, is only valid for some $\*C \subseteq Ch(X)$ in the causal diagram (Def. \ref{def:ctf-rand}). This action essentially randomizes the value of decision variable $X$ \textit{as an input} to the mechanisms generating its causal children $\*C$, while leaving open the possibility of measuring the unit $i$'s natural decision (what it would have normally decided in the $\mathcal{L}_1$ regime), and also the possibility of separately and in parallel randomizing the value of $X$ as an input to other causal children $\*C' = Ch({X}) \setminus \*C$.

\begin{figure}[t]
    \vspace{-0.in}
    \begin{center}
    \begin{tikzpicture}
    
        \node (X) at (1,1) {$X$};
        \node (A) at (1,-1) {$A$};
        \node (Y) at (0,-2) {$Y$};
        \node (Z) at (2,-2) {$Z$};
        \path [->] (X) edge (A);
        \path [->] (A) edge (Z);
        \path [->] (A) edge (Y);
        \path [<->,dashed] (Y) edge[bend right=20] (Z);

        \node (X) at (5,1) {$X$};
        \node (A) at (5,-1) {$A$};
        \node[orange] (W1) at (5,0) {$W_1$};
        \node (Y) at (4,-2) {$Y$};
        \node (Z) at (6,-2) {$Z$};
        \path [->] (X) edge (W1);
        \path [->] (A) edge (Z);
        \path [->] (A) edge (Y);
        \path [->] (W1) edge (A);
        \path [<->,dashed] (Y) edge[bend right=20] (Z);

        \node (X) at (9,1) {$X$};
        \node (A) at (9,-1) {$A$};
        \node[fill=orange,draw,inner sep=0.3em, minimum width=0.3em] (intervention) at (9,0) {\ };
        \node[orange] (x1) at (9.5,0) {$x$};
        \node (Y) at (8,-2) {$Y$};
        \node (Z) at (10,-2) {$Z$};
        \path [->] (A) edge (Z);
        \path [->] (A) edge (Y);
        \path [->] (intervention) edge (A);
        \path [<->,dashed] (Y) edge[bend right=20] (Z);
        
        \node[align=left] (t1) at (5,-3) {(a) : (Left) Graph $\mathcal{G}_1$; (Middle) enhanced diagram showing counterfactual mediator of $X$; \\ (Right) agent performing $\textsc{ctf-Rand}(X\rightarrow A)^{(i)}$ using $W_1$};

        \node (X) at (1,-4) {$X$};
        \node (Y) at (0,-7) {$Y$};
        \node (Z) at (2,-7) {$Z$};
        \path [->] (X) edge (Z);
        \path [->] (X) edge (Y);
        \path [<->,dashed] (Y) edge[bend right=20] (Z);

        \node (X) at (5,-4) {$X$};
        \node[orange] (W2) at (5.6,-5.75) {$W_2$};
        \node (Y) at (4,-7) {$Y$};
        \node (Z) at (6,-7) {$Z$};
        \path [->] (X) edge (W2);
        \path [->] (X) edge (Y);
        \path [->] (W2) edge (Z);
        \path [<->,dashed] (Y) edge[bend right=20] (Z);

        \node (X) at (9,-4) {$X$};
        \node[fill=orange,draw,inner sep=0.3em, minimum width=0.3em] (intervention) at (9.6,-5.75) {\ };
        \node[orange] (x2) at (9.5,-5.4) {$x'$};
        \node (Y) at (8,-7) {$Y$};
        \node (Z) at (10,-7) {$Z$};
        \path [->] (intervention) edge (Z);
        \path [->] (X) edge (Y);
        \path [<->,dashed] (Y) edge[bend right=20] (Z);
        \node [correct forbidden sign,line width=1ex,draw=red,fill=white, align=center, text=white] at (11,-5.4) {Not};
        
        \node[align=left] (t1) at (5,-8) {(b) : (Left) Graph $\mathcal{G}_2$; (Middle) enhanced diagram showing counterfactual mediator of $X$; \\ (Right) agent performing $\textsc{ctf-Rand}(X\rightarrow Z)^{(i)}$ using $W_2$};
            
        \end{tikzpicture}
    \end{center}
    \vskip -0.in
\caption{Given a causal diagram $\mathcal{G}_1$ of the environment, $\mathcal{G}_2$ is a valid projection of $\mathcal{G}_1$ (marginalizing $A$). If $A$ satisfies positivity w.r.t $X$, then there \textit{cannot} be a counterfactual mediator $W_2$ as shown in (b-Middle). Which means an agent \textit{cannot} perform $\textsc{ctf-Rand}(X\rightarrow Z)^{(i)}$ as shown in (b-Right).}
\label{fig:projection_graphs}
\vskip -0.125in
\end{figure}

However, the notion of "child" is an abstraction w.r.t a specific diagram of the environment under study. Consider Fig. \ref{fig:projection_graphs}(a-Left), where $\mathcal{G}_1$ is the diagram of some environment. Assume there exists a counterfactual mediator $W_1$ of $X$ (Def. \ref{def:l2_proxy}) as shown in Fig. \ref{fig:projection_graphs}(a-Middle), which means an agent is able to perform the physical action $\textsc{ctf-Rand}(X\rightarrow A)^{(i)}$, while still being able to measure the natural value of $X$ for unit $i$.

Now consider the diagram $\mathcal{G}_2$ shown in Fig. \ref{fig:projection_graphs}(b-Left). $\mathcal{G}_2$ is a valid projection of $\mathcal{G}_1$ obtained by marginalizing out variable $A$, and is thus also a valid causal diagram of the environment.

Suppose that there exists a counterfactual mediator $W_2$ as shown in \ref{fig:projection_graphs}(b-Middle). This means that the agent can also perform $\textsc{ctf-Rand}(X\rightarrow Z)^{(i)}$ in the same environment. However, since we are referring to the same environment, this means that the agent is able to perform $\textsc{ctf-Rand}(X\rightarrow Z)^{(i)}$ w.r.t the diagram $\mathcal{G}_1$, where $Z$ is not a child node of $X$! This would translate to even greater experimental power w.r.t graph $\mathcal{G}_1$, where an agent is able to perform counterfactual randomization of $X$ w.r.t further descendants like $Z$ and draw i.i.d samples from queries like $P(A_x, Z_{x'})$ by simultaneously performing both counterfactual randomizations (i.e. by randomizing $W_1, W_2$ simultaneously).

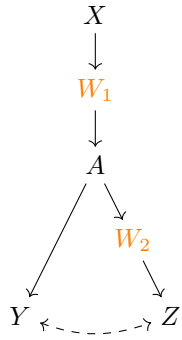
\begin{figure}
    \vspace{-0.in}
    \begin{center}
    \begin{tikzpicture}

        \node (X) at (5,1) {$X$};
        \node (A) at (5,-1) {$A$};
        \node[orange] (W1) at (5,0) {$W_1$};
        \node[orange] (W2) at (5.5,-2) {$W_2$};
        \node (Y) at (4,-3) {$Y$};
        \node (Z) at (6,-3) {$Z$};
        \path [->] (X) edge (W1);
        \path [->] (A) edge (W2);
        \path [->] (W2) edge (Z);
        \path [->] (A) edge (Y);
        \path [->] (W1) edge (A);
        \path [<->,dashed] (Y) edge[bend right=20] (Z);
        
    \end{tikzpicture}
    \end{center}
    \vskip -0.in
    \caption{"Expanded" diagram that is needed to sample directly from $P(A_x, Z_{x'})$. This is not possible, per Lemma \ref{lem:no_forks}.}
    \label{fig:projection_graphs2}
    \vskip -0in
\end{figure}

However, this scenario is not possible. Essentially, this would require an "expanded" causal diagram (Def. \ref{def:expanded_SCM}) like shown in Fig. \ref{fig:projection_graphs2}, where $W_2$ is a counterfactual mediator of $X$ w.r.t $Z$ that comes after another variable $A$. If $A$ satisfies positivity w.r.t $X$, i.e., if $P^{\mathcal{M}}(x,a)>0, \forall x,a$ in $\mathcal{L}_1$, then $W_2$ cannot be a counterfactual mediator since it cannot be uniquely mapped back to $X$.

\medskip
\begin{lemma}\label{lem:no_forks}
    Given a causal diagram $\mathcal{G}$ of a true $SCM$ $\mathcal{M}$ with a variable $X$ and $A \in Desc(X)_{\mathcal{G}}$ where $P(x,a)>0, \forall x,a$. There cannot be a variable $W$ in an "expanded" SCM $\mathcal{M}^+$ of the environment (Def. \ref{def:expanded_SCM}) s.t.
    \begin{itemize}
        \item $W \in Desc(A)_{\mathcal{G}^+}$, where $\mathcal{G}^+$ is the "expanded" causal diagram of $\mathcal{M}^+$; and
        \item $W$ is invertible to $X$, i.e. exists $f_W^{-1}$ s.t. $f_W^{-1}(W)=X$. $\hfill \square$
    \end{itemize}
\end{lemma}
\begin{proof}
    If $A$ satisfies positivity w.r.t $X$, then a given value $w$ cannot be mapped back to a unique $x$, even if we marginalize out $A$ from the SCM.

    Note that, by Assumption \ref{assn:proxy_tree}, a counterfactual mediator has only one parent in the "expanded" causal diagram (Def. \ref{def:expanded_SCM}). I.e., if it were a descendant of $A$, its perception of $X$ is fully mediated by $A$.
    
    If $f'_W(X,\*U)$ is invertible from $W$ to $X$, then so is $f_W \circ f_A(X,\*U)$. It is evident that $f_W^{'-1}$ is well defined iff $f_A^{-1}\circ f_W^{-1}$ is well defined.

    $f_A^{-1}$ is not defined. The positivity condition entails that a given value $a$ could have been generated by any value $x$ (when unit is unknown).

    Since $f'_W$ is not invertible, $W$ cannot be a counterfactual mediator.
\end{proof}

Lemma \ref{lem:no_forks} leads to some important conclusions.

\medskip
\begin{remark} \label{remark:projection1}
    There cannot be an "expanded" causal diagram (such as in Fig. \ref{fig:projection_graphs2}), with a counterfactual mediator that bypasses a child-node and directly fixes a descendent-node's perception of $X$. I.e., an agent cannot perform $\textsc{ctf-Rand}(X \rightarrow D)^{(i)}$ for some $D \in Desc(X) \setminus Ch(X)\cup \{X\}$. $\hfill \blacksquare$
\end{remark}

\medskip
\begin{remark} \label{remark:projection2}
    Conversely, given a graph like $\mathcal{G}_2$ in Fig. \ref{fig:projection_graphs}(b), if we are told that the agent can perform the action $\textsc{ctf-Rand}(X \rightarrow Z)^{(i)}$, then $\mathcal{G}_2$ cannot be a projection of $\mathcal{G}_1$ (Fig. \ref{fig:projection_graphs}(a)) for the same environment. $\hfill \blacksquare$
\end{remark}

The upshot of this discussion is that, in general (i.e. without making further assumptions), counterfactual randomization can only be done via counterfactual mediators (Def. \ref{def:l2_proxy}) of a decision variable $X$, and it can only be performed on the children-nodes of $X$ in the general case.


\newpage
\section{Details on Examples}
\label{app:examples}

\subsection{Example 1 (Mediation analysis) - further discussion}
\label{app:mediation_analysis}

For the interested reader who wants to track the formal definition of a counterfactual mediator (Def. \ref{def:l2_proxy}), we provide below a discussion of how these structural assumptions may be verified in a field experiment setting.

This discussion is not strictly needed for following the main contributions in Secs. \ref{sec:data_collection_procedures} and \ref{sec:realizability_main_sec}.

\textbf{Verification of structural assumptions:} In order for $W$ to be a counterfactual mediator of $X$ w.r.t $Y$, it needs to satisfy 3 conditions in Def. \ref{def:l2_proxy}.

\textbf{Condition [i]} says that each $w$ (say, a specific RGB range of pixels) belongs to an equivalence class that maps back uniquely to a car color $x$. This can be verified using the RCT data. \textbf{Condition [ii]} is satisfies since they can do a targeted randomization of $W$. \textbf{Condition [iii]} stipulates that $f_Y$ is not affected by any artefacts introduced by the color-editing tool: this can be verified, for instance, by swapping a car's color from $x$ to $x'$ and then swap it back from $x'$ to $x$, to ensure that the model's decision $Y$ does not change, thus verifying that the mechanism $f_Y$ only cares about what the color features $W$ reveal about $X$, and not about any image artefacts that may be introduced by editing.

\subsection{Example 2 (Causal fairness)}
\label{app:college_admissions}

\subsubsection{Simulations}

The details for the simulations shown in Fig. \ref{fig:example2}(c) are as follows. We first parameterize the space of SCMs that are compatible with the causal diagram in Fig. \ref{fig:example2}(a) using \textit{canonical parameters} \citep[~Def. 1, Thm. 1]{zhang22ab}. 

These parameters essentially discretize the domain of the latent confounder between $Y$ and $Z$ where each value of the confounder represents a joint mapping in $(X \rightarrow Y) \times (X \rightarrow Z)$. I.e., $2^2 \times 2^2 = 16$ values. We then set up a constrained optimization program to draw samples from the space of all SCMs (i.e., from the space of all trained models in Example 2) which satisfy either
\begin{itemize}
    \item $\mathcal{L}_2$ fairness measure $\mu_{int1} + \mu_{int2}$ being penalized; or
    \item $\mathcal{L}_3$ fairness measure $\mu_{ctf}$ being penalized
\end{itemize}
Drawing 1000 samples from each gives us the chart in Fig. \ref{fig:example2}(c).

\subsubsection{SCM specification}

Just to give intuition for how this disparity can arise, we present below an instantiation of an SCM using canonical parameters, showing how $\mathcal{L}_2$ measures misleadingly suggest no discrimination, whereas the $\mathcal{L}_3$ measure actually detects unfairness.


For simplicity, we assume $X$ is binary ($0$ indicates Race A, $1$ indicates Race B). $Y, Z$ are binary outcomes indicating, respectively, the CV passing through the 1st stage of screening for college admission, and for receiving a financial scholarship.

$U_X$ is the random assignment of applicant race in the CV, in the absence of intervention.

The mechanisms $f_Y$ and $f_Z$ represent the decisions of the trained classifiers used by the college data science team, which have been trained using data from previous years' decisions of committees for CV screening and financial aid.

\begin{align*}
    &\underline{\text{SCM } \mathcal{M}^\star}\\\\
    &X \gets U_X \sim Bernoulli(0.5)\\\\
    &Y \gets \begin{cases} 1, & \mbox{if Y-type = always-approve}\\ X, & \mbox{if Y-type = approve iff }x_1\\ 1-X, & \mbox{if Y-type = approve iff }x_0\\ 0, & \mbox{if Y-type = always-reject}  \end{cases}\\\\
    &Z \gets \begin{cases} 1, & \mbox{if Z-type = always-approve}\\ X, & \mbox{if Z-type = approve iff }x_1\\ 1-X, & \mbox{if Z-type = approve iff }x_0\\ 0, & \mbox{if Z-type = always-reject}  \end{cases}
\end{align*}

\begin{table}[h!]
\centering
\vspace{0.2in}
        \begin{tabular}{c c c}
            \toprule
            \textbf{Y-type}  & \textbf{Z-type} & $P(U_{YZ})$\\
            \midrule
            \midrule
            Always-approve & Always-approve & 0.040\\
            \midrule
            Always-approve & Approve iff $x_1$ & 0.175\\
            \midrule
            Always-approve & Approve iff $x_0$ & 0.160\\
            \midrule
            Always-approve & Always-reject & 0.010\\
            \midrule
            Approve iff $x_1$ & Always-approve & 0.040\\
            \midrule
            Approve iff $x_1$ & Approve iff $x_1$ & 0.055\\
            \midrule
            Approve iff $x_1$ & Approve iff $x_0$ & 0.170\\
            \midrule
            Approve iff $x_1$ & Always-reject & 0.010\\
            \midrule
            Approve iff $x_0$ & Always-approve & 0.040\\
            \midrule
            Approve iff $x_0$ & Approve iff $x_1$ & 0.140\\
            \midrule
            Approve iff $x_0$ & Approve iff $x_0$ & 0.025\\
            \midrule
            Approve iff $x_0$ & Always-reject & 0.025\\
            \midrule
            Always-reject & Always-approve & 0.050\\
            \midrule
            Always-reject & Approve iff $x_1$ & 0.010\\
            \midrule
            Always-reject & Approve iff $x_0$ & 0.025\\
            \midrule
            Always-reject & Always-reject & 0.025\\
            \bottomrule
          \end{tabular}
\vspace{-0in}
\end{table}

The CV bodies of fake applicants used in the holdout set are divided into "canonical types" such that each type elicits an approval/reject response from the models $f_Y$ and $f_Z$. Factors influencing this decision could be the prejudice of the committees, the accomplishments listed on the CV etc. (since those preferences went into building the model). $U_{YZ}$ represents of the distribution of these CV types. There are 16 such types, based on the 4 types each per model, as shown above.

For instance, row 1 of the probability table indicates a CV body such that both $f_Y$ and $f_Z$ would approve such a candidate regardless of the perceived race. Row 2 indicates a CV body such that $f_Y$ would always approve such a candidate, but $f_Z$ is biased to only approve such a candidate if they belonged to Race B.

We want to track whether, given a candidate of Race B who passed the CV screening but was rejected for financial aid, this candidate would still be denied financial aid had they been of Race A. In particular, they care about the fairness metric $\mu_{ctf}$, defined as follows.
\begin{align}
    \mu_{ctf} := P(Y_{x_1} = 1, Z_{x_1} = 0) - P(Y_{x_1} = 1, Z_{x_0} = 0) = P(y_x, z'_{x}) - P(y_x, z'_{x'}) = 10\%
\end{align}

The actual value of this measure can be computed from the probabilities in the SCM above. In practice, $\mu_{ctf}$ can be directly estimated using the counterfactual randomization procedure illustrated above in Fig. \ref{fig:example2}(b).

If the college data scientists instead follow the standard procedure of using only $\mathcal{L}_2$-data from a Fisherian RCT, they can only estimate $\mathcal{L}_2$ fairness metrics, such as $\mu_{int1}, \mu_{int2}$ defined below.
\begin{align}
    \mu_{int1} &:= P(y_x).P(z'_x) - P(y_x).P(z'_{x'}) = 0\\
    \mu_{int2} &:= P(y,z'; \doo{x}) - P(y,z'; \doo{x'}) = 0
\end{align}
The $\mathcal{L}_2$-metrics $\mu_{int1}, \mu_{int2}$ show no issues with fairness, and thus fail to capture the insight obtained from the $\mathcal{L}_3$-metric $\mu_{ctf} = 10\%$. This counterfactual insight helps the college to quantitatively characterize the financial hurdles faced by different racial groups in accessing college education, and to prevent unfair disparities.



\subsection{Example 3 (Counterfactual bandit policies)}
\label{app:bigtechsurveillance}

The SCM used in this hypothetical scenario to generate data is as follows:
\begin{align*}
&U_1 \sim Bernoulli(0.5)\\
&U_2 \sim Bernoulli(0.5)\\
&U_3 \sim Bernoulli(0.5)\\\\
&X \gets U_1 \oplus U_2 & \text{($\oplus$ is the XOR function)}\\
&D \gets X \oplus U_3
\end{align*}
{Since $Y$ is a function of $X$, the average outcome is shown below for different realizations of the latents}

\underline{i. Avg. $Y$, when $U_3=0$}
\begin{table}[h]
\vspace{-0.in}
\begin{tabular}{|c|c|c|c|c|}
\hline
  & \multicolumn{4}{|c|}{$U_3=0$}\\
  \hline
  & \multicolumn{2}{c|}{$U_1=0$} & \multicolumn{2}{c|}{$U_1=1$}\\
  \hline
  & $U_2=0$ & $U_2=1$ & $U_2=0$ & $U_2=1$\\
 \hline
 $do(x_0)$  & {\textbf{0.6}}  & 0.9 & 0.8 & \textbf{0.5}\\
 \hline
 $do(x_1)$  & {0.9}  & \textbf{0.6} & \textbf{0.5} & 0.8\\
\hline
\multicolumn{5}{c}{ } \\
\multicolumn{5}{l}{Natural choice of $X$ marked \textbf{bold}}
\end{tabular}
\vspace{-0.1in}
\end{table}

\medskip
\underline{ii. Avg. $Y$, when $U_3=1$}
\begin{table}[h]
\vspace{-0.in}
\begin{tabular}{|c|c|c|c|c|}
\hline
  & \multicolumn{4}{|c|}{$U_3=1$}\\
  \hline
  & \multicolumn{2}{c|}{$U_1=0$} & \multicolumn{2}{c|}{$U_1=1$}\\
  \hline
  & $U_2=0$ & $U_2=1$ & $U_2=0$ & $U_2=1$\\
 \hline
 $do(x_0)$  & {\textbf{0.8}}  & 0.7 & 0.6 & \textbf{0.7}\\
 \hline
 $do(x_1)$  & {0.7}  & \textbf{0.8} & \textbf{0.7} & 0.6\\
\hline
\end{tabular}
\vspace{-0.1in}
\end{table}

\medskip
\underline{iii. Avg. $Y$, with $U_3$ marginalized (consolidating i. and ii.)}
\begin{table}[h!]
\vspace{-0.in}
\begin{tabular}{|c|c|c|c|c|}
\hline
  & \multicolumn{2}{c|}{$U_1=0$} & \multicolumn{2}{c|}{$U_1=1$}\\
  \hline
  & $U_2=0$ & $U_2=1$ & $U_2=0$ & $U_2=1$\\
 \hline
 $do(x_0)$  & {\textbf{0.7}}  & 0.8 & 0.7 & \textbf{0.6}\\
 \hline
 $do(x_1)$  & {0.8}  & \textbf{0.7} & \textbf{0.6} & 0.7\\
\hline
\end{tabular}
\vspace{-0.in}
\end{table}

Let us call the social media user Alice, for ease of reference. $U_1, U_2, U_3$ are latent attributes affecting Alice's decisions each evening. In particular, $U_1$ indicates whether she is tired, $U_2$ indicates whether she had a busy day and is distracted, $U_3$ indicates whether she is hungry, on any given evening.

If Alice is either tired but mentally relaxed ($X = 1 \oplus 0$), or if she is physically energetic but distracted ($X = 0 \oplus 1$), Alice decides to take a walk and use social media via mobile app. If Alice is neither tired nor distracted, she prefers to continue working on her desktop and uses social media via desktop app during breaks ($X = 0 \oplus 0$). If she is both tired and distracted, she also decides to use the social media app on her desktop because she has no energy to take a walk ($X = 1 \oplus 1$).

There are so many possible factors affecting her decisions, Alice is unaware that these are the specific unconscious causes of her natural choices. However, the social media company's unscrupulous data scientists surveil $U_1, U_2, U_3$ (perhaps by tracking Alice's wearable health monitor and calendar) and predict her natural choice. The company then uses behavioural insights to ping Alice with the precise notifications and content to maximize her time spent on the platform for each realization of $U_1, U_2, U_3$.

$D$ is the type of ads Alice sees when she logs in to the social media app for the day.

\begin{figure}[h]
    \centering
    \begin{tikzpicture}
        \node (X) at (0,0) {$X$};
        \node (D) at (2,1.5) {$D$};
        \node (Y) at (4,0) {$Y$};
        \node (U3) at (4,1.5) {$U_3$};
        \node (U12) at (2,-1.5) {$U_1, U_2$};
        \path [->,dashed] (U3) edge (D);
        \path [->,dashed] (U3) edge (Y);
        \path [->,dashed] (U12) edge (X);
        \path [->,dashed] (U12) edge (Y);
        \path [->] (X) edge (Y);
        \path [->] (X) edge (D);
    \end{tikzpicture}
    \vspace{0.1in}
    \caption{Causal diagram for Example 3. $X:$ app usage type; $D:$ advertisement-type received; $Y:$ compliance with app usage time-limit; $U_1:$ agent tiredness; $U_2:$ agent busyness earlier in the day; $U_3:$ indicator of whether the agent is hungry.}
    \label{fig:app_example3_DAG}
\end{figure}
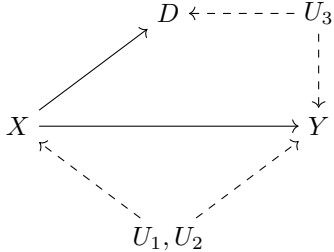

The detailed causal diagram is shown in Fig. \ref{fig:app_example3_DAG}.

\paragraph{$\mathcal{L}_1$-regime:}The observational data is contained in Table (iii) in the SCM above, where the bold values correspond to Alice's natural choices. Given that all combinations of latents happen with equal probability, it is easy to see that the expected reward in the observational regime is $E[Y] = (0.25)(0.7 + 0.7 + 0.6 + 0.6) = 0.65$.

\begin{wraptable}{r}{6 cm}
\centering
\vspace{-0.in}
\begin{tabular}{|c|c|}
\hline
  & $E[Y ; do(x)]$\\
 \hline
 $do(x_0)$  & {0.7}  \\
 \hline
 $do(x_1)$  & {0.7}   \\
\hline
\end{tabular}
\caption{Expected outcome $E[Y_x]$ computed under the interventional regime.}
\label{tab:alice_l2}
\vspace{-0.1in}
\end{wraptable}

\paragraph{$\mathcal{L}_2$-regime:} Applying the interventions $do(x_0), do(x_1)$, we can compute the expected outcome from the SCM as shown in Table \ref{tab:alice_l2}. This is simply the average of all the values in Table (iii) of the SCM above.

An interventional strategy of randomizing ones actions (or fixing a constant action) outperforms the observational $\mathcal{L}_1$ regime of allowing one's actions to be determined by natural inclination.

\paragraph{$\mathcal{L}_3$-regime - ETT:} By counterfactual randomization Alice can sample from the $\mathcal{L}_3$ distribution $P(Y_x, X)$. She records her natural choice $X=x'$ on a particular evening (what she would have normally done) and randomizes the choice of $X$ that she actually undertakes, during the explore phase. Using this distribution, she then performs the following action, for the natural $X = x'$ that she observes in the exploit phase:
$$do(X = \arg \max_x E[Y_x \mid x'])$$

We can compute this from Table (iii) of the SCM. Alice simply chooses to do the opposite of what she naturally feels like doing (corresponding to the the non-bold cells of the Table). This "ETT" $\mathcal{L}_3$ strategy yields an expected outcome of 
$$\sum_{x'} P(x').\max_x E[Y_x \mid x'] = (0.5)[0.7 + 0.8] = 0.75,$$
outperforming both $\mathcal{L}_1$ and $\mathcal{L}_2$ strategies.

Of course, the explore-exploit phases are combined adaptively in a bandit algorithm like Thompson Sampling.

\paragraph{$\mathcal{L}_3$-regime - Optimal:} Finally, Alice leverages her ability to perform \textit{path-specific} randomization to sample from the distribution $P(Y_x, X, D_{x'})$. She then adapts a bandit algorithm to  performs the following actions in the exploit phase:
$$\textsc{Read}(X) = x'$$
$$\textsc{ctf-Write}(x'' \rightarrow D), \text{ where } x'' = \arg \max_{x''} \bigg(\max_x E[Y_x \mid x', D_{x''}]\bigg) ; \textsc{Read}(D) = d $$
$$\textsc{ctf-Write}(x \rightarrow Y), \text{ where } x = \arg \max_{x} E[Y_x \mid x', d_{x''}]$$,

where $\textsc{ctf-Write}$ is simply the deterministic equivalent of $\textsc{ctf-Rand}$.

In words, during the explore phase, Alice gathers data on which arm $x$ optimizes $\mathbb{E}[Y_x \mid x', d_{x''}]$, for all $x',, x'', d$. Then, during the exploit phase, Alice first observes $D_{x''}=d$ and $X=x'$, and then performs the action $x$ which maximizes her outcome $Y_x$. Performing another optimization over the $x''$ gives her the best global optimum of $E[Y_x \mid x', d_{x''}]$.

Again, the explore-exploit phases are not separated in a bandit algorithm like Thompson Sampling, but incorporated adaptively.

Computing this from the SCM, suppose Alice chooses to record $D_{x_0} = 0 \oplus U_3 = U_3$.
\begin{itemize}
    \item When $D_{x_0} = U_3 = 0$, Alice sees according to Table (i) of the SCM that the optimal strategy is to choose the opposite of what she naturally feels like doing (the values not in bold), giving the expected outcome $E[Y_x \mid x', D_{x_0} = 0],$ where $x \neq x'$, as $(0.5)[0.9 + 0.8] = 0.85$
    \item When $D_{x_0} = U_3 = 1$, Alice sees according to Table (ii) of the SCM that the optimal strategy is to go with her natural inclination (the values in bold), giving the expected outcome $E[Y_{x'} \mid x', D_{x_0} = 0] = (0.5)[0.8 + 0.7] = 0.75$
    \item Overall, since both values of $D_{x_0}$ are equally likely, this strategy yields an expected outcome of $0.5[0.85 + 0.75] = 0.8$, which outperforms $\mathcal{L}_1$, $\mathcal{L}_2$ and $\mathcal{L}_3$-ETT strategies.
\end{itemize}
This walk-through can be repeated identically from the SCM had Alice chosen to measure $D{x_1}$ instead.

\subsubsection{Simulations} 
\label{app:example3_simulations}

\begin{figure}[h!]
    \centering
        \includegraphics[width=0.7\textwidth]{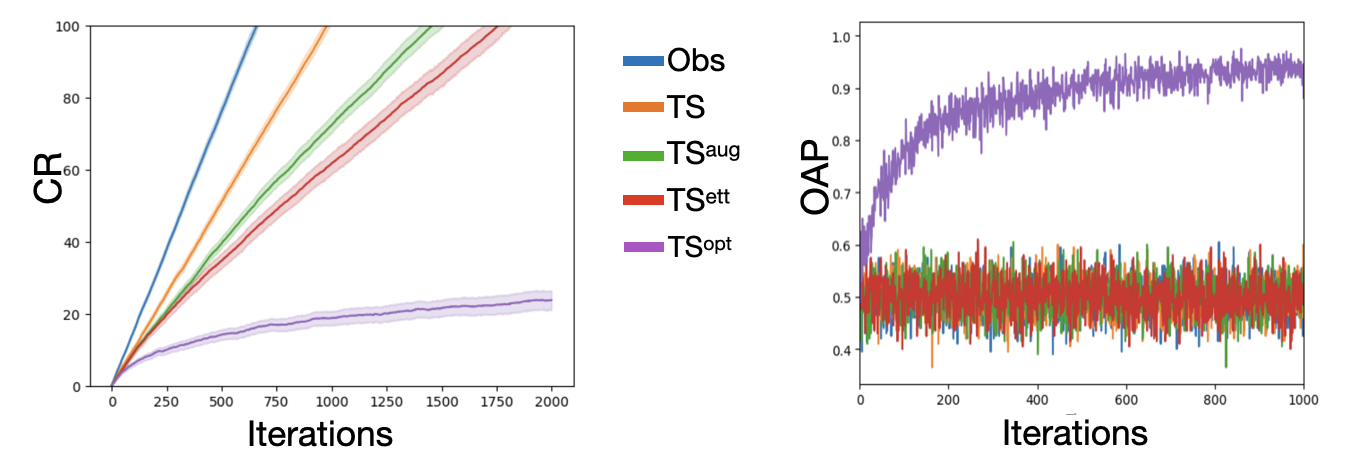}
    \vspace{-0.in}
    \caption{Example 3: Cumulative Regret (CR) and Optimal Arm Probability (OAP) for all strategies tested via Thompson Sampling.}
    \vspace{-0.in}
    \label{fig:example3_app}
\end{figure}

The simulation compares the performance of four algorithms
\begin{itemize}
    \item $\text{TS}$ is the conventional Thompson Sampling algorithm that optimizes the $\mathcal{L}_2$ learning objective $\mathbb{E}[Y; \doo{x}]$;
    \item $\text{TS}^{aug}$ is a contextual Thompson Sampling algorithm that treats $\{X=x',D_{x''}=d\}$ as merely some extra context variables in each round, and ignoring the $\mathcal{L}_3$ significance of these variables;
    \item $\text{TS}^{ett}$ is given in Algorithm \ref{alg:ts_ett}, implementing the ETT baseline strategy described earlier;
    \item $\text{TS}^{opt}$ is given in Algorithm \ref{alg:ts_opt}, implementing the $\mathcal{L}_3$-optimal strategy described earlier. 
\end{itemize}

Importantly, $\text{TS}^{opt}$ doesn't treat $\{X=x', D_{x''} = d \}$ merely as extra context variables. Rather, the counterfactual significance of these variables is leveraged via the consistency property
\begin{align}
    E[Y_{x} \mid x,d_{x}] = E[Y\mid x,d]
\end{align}

This means that for several arms being explored, the r.h.s allows us to hot-start the Thompson Sampling using offline ($\mathcal{L}_1)$ data, as implemented in Line 18 of Algorithm \ref{alg:ts_opt}. This allows for a dramatically faster convergence of the purple vs. green plot in Fig. \ref{fig:example3_app}.

Simulations were run for 2000 iterations, 200 epochs (Confidence Interval = 95\%).

\begin{algorithm}[h]
   \caption{$TS^{opt}$: Thompson Sampling OPTIMAL (Bernoulli-Beta case)}
   \label{alg:ts_opt}
\begin{algorithmic}[1]
   \STATE {\bfseries Input:} No. of timesteps, $T$; Observational data, $P(\*v)$ 

   \smallskip
   \FOR{$x'' \in$ Domain($X$)}
   \FOR{$x' \in$ Domain($X$)}
        \STATE $\alpha_D[x''][x'] \gets 1$
        \STATE $\beta_D[x''][x'] \gets 1$ \COMMENT{Initializing $D$-priors}
   \ENDFOR
   \ENDFOR

   \smallskip
   \FOR{$i \in$ Domain($X$)}
   \FOR{$j \in$ Domain($X$)}
   \FOR{$d \in$ Domain($D$)}
   \FOR{$k \in$ Domain($X$)}
        \STATE $\alpha_Y[x_i][x_j][d][x_k] \gets 1$
        \STATE $\beta_Y[x_i][x_j][d][x_k] \gets 1$ \COMMENT{Initializing $Y$-priors}
   \ENDFOR
   \ENDFOR
   \ENDFOR
   \ENDFOR

   \smallskip
   \STATE $t = 1$
   \WHILE{$t<=T$}
   \smallskip
   \STATE Perform $\textsc{Read}(X) = x'$, for unit

   \smallskip
   \FOR{$j \in$ Domain($X$)}
    \STATE $\mu_i^D \sim Beta(\alpha_D[x''][x_j], \beta_D[x''][x_j])$
   \ENDFOR

   \smallskip
   \STATE Perform $\textsc{ctf-Write}(x' \rightarrow D)$ for $x' = x_j; j := \arg \max_{j'} \mu_{j'}^D$
   \STATE Perform $\textsc{Read}(D) = d$, for unit \COMMENT{Get value of $D_{x''}$}

   \smallskip
   \FOR{$k \in$ Domain($X$)}
   \IF{$x_k = x' = x''$}
        \STATE $\mu_k^Y \gets E[Y \mid x'',d]$ \COMMENT{Hot-start using obs. data}
   \ELSE
        \STATE $\mu_k^Y \sim Beta(\alpha_Y[x''][x'][d][x_k], \beta_Y[x''][x'][d][x_k])$
   \ENDIF
   \ENDFOR

   \smallskip
   \STATE Perform $\textsc{ctf-Write}(x \rightarrow Y)$ for $x = x_k; k := \arg \max_{k'} \mu_{k'}^Y$
   \STATE Perform $\textsc{Read}(Y) = y$, for unit \COMMENT{Get value of $Y_{x}$}

   \smallskip
   \STATE $\alpha_D[x''][x'] \gets \alpha_D[x''][x'] + y$
   \STATE $\beta_D[x''][x'] \gets \beta_D[x''][x'] + 1-y$ \COMMENT{Update D-priors}

   \IF{$\neg(x = x' = x'')$}
        \STATE $\alpha_Y[x''][x'][d][x] \gets \alpha_Y[x''][x'][d][x] + y$
        \STATE $\beta_Y[x''][x'][d][x] \gets \beta_Y[x''][x'][d][x] + 1-y$ \COMMENT{Update Y-priors}
   \ENDIF
   
   \smallskip
   \STATE $t \gets t+1$

   \ENDWHILE
\end{algorithmic}
\end{algorithm}

\begin{algorithm}[h!]
   \caption{$TS^{ett}$: Thompson Sampling ETT (Bernoulli-Beta case)}
   \label{alg:ts_ett}
\begin{algorithmic}[1]
   \STATE {\bfseries Input:} No. of timesteps, $T$; Observational data, $P(\*v)$ 

   \smallskip
   \FOR{$z \in$ Domain($Z$)}
   \FOR{$x' \in$ Domain($X$)}
        \STATE $\alpha[z][x'] \gets 1$
        \STATE $\beta[z][x'] \gets 1$ \COMMENT{Initializing priors}
   \ENDFOR
   \ENDFOR

   \smallskip
   \STATE $t = 1$
   \WHILE{$t<=T$}
   \smallskip
   \STATE Perform $\textsc{Read}(Z) = z$, for unit
   \STATE Perform $\textsc{Read}(X) = x'$, for unit

   \smallskip
   \FOR{$i \in$ Domain($X$)}
   \IF{$x_i = x'$}
        \STATE $\mu_i \gets E[Y \mid x',z]$ \COMMENT{Hot-start using obs. data}
   \ELSE
        \STATE $\mu_i \sim Beta(\alpha[z][x_i], \beta[z][x_i])$
   \ENDIF
   \ENDFOR

   \smallskip
   \STATE Perform $\textsc{ctf-Write}(x \rightarrow Y)$ where $x = x_i$ s.t. $i := \arg \max_{i'} \mu_{i'}$
   \STATE Perform $\textsc{Read}(Y) = y$, for unit \COMMENT{Get value of $Y_{x}$}

   \smallskip
   \IF{$x \neq x'$}
       \STATE $\alpha[z][x'] \gets \alpha[z][x'] + y$
       \STATE $\beta[z][x'] \gets \beta[z][x'] + 1-y$ \COMMENT{Update priors}
   \ENDIF
   
   \smallskip
   \STATE $t \gets t+1$

   \ENDWHILE
\end{algorithmic}
\end{algorithm}

\newpage
{   }


\clearpage
\section{Optimality proof}
\label{app:optimality_proofs}

In this appendix, we provide proofs for Theorem \ref{thm:optimality} and Corollary \ref{cor:l3_domination}.

\bigskip
\begin{remark}\label{rem:optimal_strategy}
    In order for an agent to enact a non-trivial decision strategy $\pi: \{\*W_\star=\*w\} \mapsto \mathcal{A}$, we observe that (1) the distribution $P(Y_{\mathcal{A}},\*W_\star)$ must be realizable (Def. \ref{def:realizability}); and (2) the agent must be able to observe $\*W_\star$ before performing actions $\mathcal{A}$. We call this a \textit{realizable decision strategy}, and notate the space of all realizable strategies in a MAB problem as $\Pi$. $\hfill \blacksquare$
\end{remark}

\bigskip
\paragraph{Corollary \ref{cor:l3_domination}}

:
This result follows immediately from Theorem \ref{thm:optimality}, by simply recognizing that $\pi^{int}, \pi^{obs} \in \Pi$, the space of realizable strategies (the agent is presumed capable of performing the actions $\textsc{Rand}(X), \textsc{Write}(X:x)$).

Therefore, $\mu_{\pi^{ctf}}$ cannot be less than $\mu_{\pi^{int}}, \mu_{\pi^{obs}}$, by Theorem \ref{thm:optimality}. $\hfill \blacksquare$

\bigskip
\bigskip

\paragraph{Theorem \ref{thm:optimality}}

:

From Lemma \ref{lem:opt_lem_1}, all strategies involve mappings, where each mapping maps to one of the following 5 possible action sets: (1) $\{\}$ (no action); (2) $\textsc{Write}(X:x)$, for some $x$; (3) only $\textsc{ctf-Write}(x \rightarrow Y)$ for some $x$; (4) only $\textsc{ctf-Write}(x'' \rightarrow D)$ for some $x''$; or (5) both $\textsc{ctf-Write}(x \rightarrow Y)$, $\textsc{ctf-Write}(x'' \rightarrow D)$ for some $x,x''$.

Define $\Pi_5$ to be the space of strategies where every mapping of each strategy in $\Pi_5$ is mapping to a pair of actions $\textsc{ctf-Write}(x \rightarrow Y)$, $\textsc{ctf-Write}(x'' \rightarrow D)$ for some $x,x''$. I.e., all mappings only involve possibility (5) under these strategies, for all any unit encountered in the decision problem. 

Let $\pi_5$ be an optimal strategy in this space. I.e., $\pi_5 \in \arg \max_{\pi \in \Pi_5} \mu_{\pi}$.

By Lemma \ref{lem:opt_lem_3}, $\pi_5$ is also an optimal strategy in the space of all possible strategies. This means we only need to consider strategies whose mappings are mappings to a pair of $\textsc{ctf-Write}$ procedures.

Let $\*W_\star$ be the context used by $\pi_5$. If $\*W_\star$ does not already contain the natural variables $X,Z$, we can always define $\pi'_5$ that use context $\*W'_\star = \*W_\star \cup \{X,Z\}$ s.t. $\mu_{\pi'_5} = \mu_{\pi_5}$, where $\pi'_5$ simply ignores the extra context variable in the mapping.

Such a move would not affect the realizability of $\pi'_5$ because $\textsc{ctf-Write}$ does not override the natural value of $X$, and both $X,Z$ can be observed before decision-making.

Combinatorially, there are only 3 possibilities for picking each mapping in $\pi'_5$.
\begin{enumerate}
    \item Mapping from $\{x',z\}$ to a pair of actions $\{\textsc{ctf-Write}(x \rightarrow Y),\textsc{ctf-Write}(x'' \rightarrow D)\}$
    \item Mapping from $\{x',z\}$ to some $\textsc{ctf-Write}(x \rightarrow Y)$, observing $Y_x = y$, and mapping from $\{x',z,y_x\}$ to $\textsc{ctf-Write}(x'' \rightarrow D)$; or
    \item Mapping from $\{x',z\}$ to some $\textsc{ctf-Write}(x'' \rightarrow D)$, observing $D_{x''} = d$, and mapping from $\{x',z,d_{x''}\}$ to $\textsc{ctf-Write}(x \rightarrow Y)$
\end{enumerate}

We can use similar arguments to Lemma \ref{lem:opt_lem_3}, where we restricted our attention to the space of strategies $\Pi_5$ which could mimic all other optimal strategies.

Possibility 2 can be mimicked by some mapping following possibility 1 which maps to a joint pair of actions. The two are equivalent in terms of outcome, because conditioning on $y_x$ to choose $x''$ does not affect the outcome $Y$. So we can restrict our attention to possibilities 1 and 2.

Each mapping of possibility 1 can be mimicked by possibility 3, where the extra step of conditioning on $d_{x''}$ just ignores the extra information about $d_{x''}$. Thus, we can replace all mappings in the optimal strategy $\pi'_5$ with mappings of possibility 3, to get a strategy $\pi''_5$ that also performs optimally.

Since there are two mappings in $\pi''_5$, they must be the mappings which maximize the outcome.

This is precisely the definition of the strategy $\pi^{opt}$ given in Equation \ref{eq:pi_opt} and in the description immediately following it.

$\hfill \blacksquare$

\bigskip
\begin{lemma}\label{lem:opt_lem_1}
    Any decision strategy $\pi$ for a decision problem having causal structure same as the MAB template is s.t. each mapping of the strategy maps from domain of the context to one of the five following possible sets of actions: (1) $\{\}$ (no action); (2) $\textsc{Write}(X:x)$, for some $x$; (3) only $\textsc{ctf-Write}(x \rightarrow Y)$ for some $x$; (4) only $\textsc{ctf-Write}(x'' \rightarrow D)$ for some $x''$; or (5) both $\textsc{ctf-Write}(x \rightarrow Y)$, $\textsc{ctf-Write}(x'' \rightarrow D)$ for some $x,x''$. 
\end{lemma}
\begin{proof}
    Since the physical action space only involves doing nothing, $\textsc{Write}$ or $\textsc{ctf-Write}$. Any other combination would be equivalent to one of the 5 above. E.g., $\textsc{Write}(X:x)$ and $\textsc{ctf-Write}(x'' \rightarrow D)$ is the equivalent to the pair $\textsc{ctf-Write}(x\rightarrow Y)$ and $\textsc{ctf-Write}(x'' \rightarrow D)$ (see Remark \ref{remark:supersede}).

    We ignore randomized actions for simplicity. From standard results in learning theory, there is an optimum to be found at a simplex corner so we need only search over the space of hard interventions.
\end{proof}

\bigskip
\begin{lemma}\label{lem:opt_lem_2}
    The context $\*W_\star$ used in the strategy $\pi: \{\*W_\star=\*w\} \mapsto \mathcal{A}$ can only possibly contain a subset of $X, Z, D, D_{x''}$ for some $x''$, and at most one potential response of $D$.
\end{lemma}
\begin{proof}
    There are only 4 variables to consider: $X,Y,Z,D$.
    
    By the definition of a realizable strategy (Remark \ref{rem:optimal_strategy}), we need $P(Y_\mathcal{A},\*W_\star)$ to be realizable. By Cor. \ref{cor:maximal_amwn} there cannot be two potential responses of the same variable in a realizable distribution. This rules out any other potential response of $Y$, and ensures only one each of $X,Z,D$. 

    Since the only possible actions are interventions involving $X$, which do not affect $Z$ and $X$ (natural variable), these are the only potential responses that could appear involving these variables.

    Likewise, with $D$, $D$ (natural value) and $D_{x''}$ are the only possible potential responses that could appear, and at most one of them can. 
\end{proof}

\bigskip
\begin{lemma}\label{lem:opt_lem_3}
    If $\pi_5$ is an optimal strategy in $\Pi_5$, the set of all strategies which map to a pair of $\textsc{ctf-Write}$ procedures, then $\pi_5$ is also an optimal strategy in the set of all strategies possible in the MAB decision problem. 
\end{lemma}
\begin{proof}
    Let $\Pi_1$ be the space of all strategies possible in the problem. Note that $\Pi_5 \subseteq \Pi_1$. Let $\pi_1 \not \in \Pi_5$ be an optimal strategy. I.e. $\pi_1 \in \arg \max_{\pi \in \Pi_1} \mu_\pi$. If no such $\pi_1$ the Lemma stands proved.

    Let $\*W_\star$ be the context used by $\pi_1$. If $\*W_\star$ does not already contain the natural variable $X$, we can always define $\pi'_1$ that uses context $\*W'_\star = \*W_\star \cup \{X\}$ s.t. $\mu_{\pi'_1} = \mu_{\pi_1}$, where $\pi'_1$ simply ignores the extra context variable in the mapping. For now, it doesn't matter whether such $\pi'_1$ is realizable or not. Just that it is also an optimal strategy.

    Each mapping in the strategy $\pi'_1$ maps from the domain of $\*W'_\star$ to one of the five possible action sets mentioned in Lemma \ref{lem:opt_lem_1}. E.g., for some $\*W'_\star=\*w$, the strategy $\pi'_1$ maps this to $\*w \mapsto \{\}$ or $\*w \mapsto \textsc{Write(x)}$.

    Consider a mapping in $\pi'_1$ from the domain of $\*W'_\star = \{x',...\}$ to possibility (1), empty set of actions (recall, the context includes natural $X$). Such a mapping can be mimicked by an equivalent mapping $\*W'_\star = \{x',...\} \mapsto \{\textsc{ctf-Write}(x' \rightarrow Y),\textsc{ctf-Write}(x' \rightarrow D)\}$. By the consistency property if $X(\*u)=x'$, then $Y_{x'}(\*u) = Y(\*u)$ and $D_{x'}(\*u)=D(\*u)$. 
    
    Thus, we can replace all the mappings in $\pi'_1$ that involve a mapping to the empty set of actions, with an equivalent pair of $\textsc{ctf-Write}$ using the natural value of $X$ observed in the context. Call this new strategy $\pi_2$. $\pi_2$ is as good as $\pi'_1$ because the mappings are all equivalent. Thus, $\pi_2$ is also optimal in $\Pi_1$. Again, it doesn't matter that $\pi_2$ may not be realizable, just that it is optimal.

    Next, consider a mapping in $\pi_2$ from the domain of $\*W'_\star = \{x',...\}$ to possibility (2), some action $\textsc{Write}(X:x)$. Such a mapping can be mimicked by an equivalent mapping $\*W'_\star = \{x',...\} \mapsto \{\textsc{ctf-Write}(x \rightarrow Y),\textsc{ctf-Write}(x \rightarrow D)\}$. The evaluation of $f_Y(x,Z,\*u)$ in both scenarios is the same, with the only difference being that $f_X$ is overwritten, which doesn't affect the outcome $Y$ for each $\*u$. I.e., the outcome $Y$ would be the same for every unit under both strategies.
    
    Thus, we can replace all the mappings in $\pi_2$ that involve a mapping to some action $\textsc{Write}(X:x)$, with an equivalent pair of $\textsc{ctf-Write}$. Call this new strategy $\pi_3$. $\pi_3$ is as good as $\pi_2$ because the mappings are all equivalent in terms of outcome. Thus, $\pi_3$ is also optimal in $\Pi_1$.

    Next, consider a mapping in $\pi_3$ from the domain of $\*W'_\star = \{x',...\}$ to possibility (3), some action $\textsc{ctf-Write}(x \rightarrow Y)$. Such a mapping can be mimicked by an equivalent mapping $\*W'_\star = \{x',...\} \mapsto \{\textsc{ctf-Write}(x \rightarrow Y),\textsc{ctf-Write}(x' \rightarrow D)\}$ for natural value $x'$. By the consistency property, if $X(\*u)=x'$ then $D_{x'}(\*u) = D(\*u)$.
    
    Thus, we can replace all the mappings in $\pi_3$ that involve a mapping to some action $\textsc{ctf-Write}(x \rightarrow Y)$, with an equivalent pair of $\textsc{ctf-Write}$. Call this new strategy $\pi_4$. $\pi_4$ is as good as $\pi_3$ because the mappings are all equivalent in terms of outcome. Thus, $\pi_4$ is also optimal in $\Pi_1$.

    Next, consider a mapping in $\pi_4$ from the domain of $\*W'_\star = \{x',...\}$ to possibility (4), some action $\textsc{ctf-Write}(x \rightarrow D)$. Such a mapping can be mimicked by an equivalent mapping $\*W'_\star = \{x',...\} \mapsto \{\textsc{ctf-Write}(x' \rightarrow Y),\textsc{ctf-Write}(x \rightarrow D)\}$ for natural value $x'$. By the consistency property, if $X(\*u)=x'$ then $Y_{x'}(\*u) = Y(\*u)$.
    
    Thus, we can replace all the mappings in $\pi_4$ that involve a mapping to some action $\textsc{ctf-Write}(x \rightarrow D)$, with an equivalent pair of $\textsc{ctf-Write}$. Call this new strategy $\pi'_5$. $\pi'_5$ is as good as $\pi_4$ because the mappings are all equivalent in terms of outcome. Thus, $\pi'_5$ is also optimal in $\Pi_1$.

    However, note that the only possible mappings in $\pi'_5$ are possibility (5) involving a pair of $\textsc{ctf-Write}$ actions. Which means $\pi'_5 \in \Pi$.

    Thus, we show that all optimal strategies in $\Pi_5$ are also optimal in the overall space of strategies.
\end{proof}

\end{document}